\documentclass{article}

\usepackage{arxiv}

\usepackage[utf8]{inputenc} 
\usepackage[T1]{fontenc}    
\usepackage{hyperref}       
\usepackage{url}            
\usepackage{booktabs}       
\usepackage{amsfonts}       
\usepackage{nicefrac}       
\usepackage{microtype}      
\usepackage{lipsum}
\usepackage{graphicx}
\usepackage{multirow}
\usepackage{tabularx}
\usepackage{caption}
\usepackage{subcaption}
\usepackage{algcompatible}
\usepackage{algorithm}
\usepackage{amsmath}
\usepackage{color}

\newcolumntype{Y}{>{\arraybackslash}X}

\newcommand{\rev}[1]{\textcolor{black}{#1}}

\title{RB-CCR: Radial-Based Combined Cleaning and Resampling algorithm for imbalanced data classification}

\author{
  Michał Koziarski \\
  Department of Electronics\\
  AGH University of Science and Technology\\
  Al. Mickiewicza 30, 30-059 Kraków, Poland\\
  \texttt{michal.koziarski@agh.edu.pl} \\
  \And
  Colin Bellinger \\
  Digital Technologies\\
  National Research Council of Canada \\
  Ottawa, Canada \\
  \texttt{colin.bellinger@nrc-cnrc.gc.ca} \\
  \And
  Michał Woźniak \\
  Department of Systems and Computer Networks \\
  Wroc\l{}aw University of Science and Technology \\
  Wybrze\.ze Wyspia\'nskiego 27, 50-370 Wroc\l{}aw, Poland \\
  \texttt{michal.wozniak@pwr.edu.pl} \\
}

\begin{document}
\maketitle

\begin{abstract}
Real-world classification domains, such as medicine, health and safety, and finance, often exhibit imbalanced class priors and have asynchronous misclassification costs. In such cases, the classification model must achieve a high recall without significantly impacting precision. Resampling the training data is the standard approach to improving classification performance on imbalanced binary data. However, the state-of-the-art methods ignore the local joint distribution of the data or correct it as a post-processing step. This can causes sub-optimal shifts in the training distribution, particularly when the target data distribution is complex. In this paper, we propose Radial-Based Combined Cleaning and Resampling (RB-CCR). RB-CCR utilizes the concept of class potential to refine the energy-based resampling approach of CCR. In particular, RB-CCR exploits the class potential to accurately locate sub-regions of the data-space for synthetic oversampling. The category sub-region for oversampling can be specified as an input parameter to meet domain-specific needs or be automatically selected via cross-validation. Our $5\times2$ cross-validated results on 57 benchmark binary datasets with 9 classifiers show that RB-CCR achieves a better precision-recall trade-off than CCR and generally out-performs the state-of-the-art resampling methods in terms of AUC and G-mean.
\end{abstract}


\section{Introduction}

Machine learning classifiers are quickly becoming a tool of choice in application areas ranging from finance to robotics and medicine. This is largely owing to the growth in the availability of labeled training data and declining computing costs. When applied correctly, machine learning classifiers have the potential to improve safety and efficiency and reduce costs. However, many of the most important domains, such as those related to health and safety, are limited by the problem of class imbalance. In binary classification, the class imbalance is defined as occurring when the prior probability of one class (referred to as the minority class) is significantly lower than the prior probability of the other class (majority class). 

The induction of binary classifiers on imbalanced training data results in a predictive bias toward the majority class and has been associated with poor performance during application \cite{branco2016survey}. Detailed empirical studies have demonstrated that class imbalance exacerbates the difficulty of learning accurate predictive models from complex data involving class overlap, sub-concepts, non-parametric distributions, \textit{etc.} \cite{he2009learning,stefanowski2016dealing}. 

Traditional methods of improving the predictive performance of classification models trained on imbalanced data involve resampling (random undersampling the majority class, random oversampling the minority class and generating additional synthetic minority samples) or cost-adjustment \cite{branco2016survey}. Synthetic minority sampling methods, such as SMOTE and its derivatives \cite{chawla2002smote,han2005borderline,he2008adasyn,barua2012mwmote,bellinger2016beyond}, generate synthetic minority samples to balance the training set. Generation-based methods of this nature are widely applied because they are classifier independent and can reduce the risk of overfitting.

In addition to elevating the learning challenge, in many cases, imbalanced training data results from sensitive application domains that exhibit asymmetric misclassification cost \cite{wallace2012class}. For example, in medicine, misclassifying benign cases as cancerous (false positive) can have negative consequences in terms of mental anguish and additional tests. Whilst false positives should be kept to a minimum, misclassifying a cancerous case as benign (false negative) can significantly increase cost in terms of delayed treatment and premature death. In domains of this nature, additional effort must be made to induce a classifier with good predictive performance on the minority class. 

To achieve satisfactory performance on sensitive imbalanced domains with asymmetric misclassification costs, the resampling strategy ought to prioritizing high recall whilst having minimal impact on precision. In this work, we propose a refinement to the CCR algorithm \cite{koziarski2017ccr} that utilizes the radial-based (RB) approach to calculate the class potential to satisfy this objective. Specifically, CCR is a resampling algorithm that cleanses majority class training samples and randomly generates synthetic minority samples in the regions around the minority class. Whilst this technique has been shown to improve the recall of the induced classifier, the specific resampling strategy employed may limit the improvement in recall and risks harming the precision. To improve upon this, we propose the RB-CCR resampling algorithm. \rev{It focuses the generation processes in sub-regions of the data-space that satisfy the user-specified class potential targets. The ability to do this gives the user better control over the precision-recall trade-off. This, for example, enables higher recall on domain for which this is critical.} 

We empirically compare RB-CCR to CCR and the state-of-the-art resampling methods on 57 benchmark datasets with 9 classifiers. Our empirical results show that resampling with RB-CCR can be exploited to control the precision-recall trade-off in a domain-appropriate way. On average, RB-CCR outperforms the state-of-the-art alternatives in terms of AUC and G-mean.

The main contributions of this paper can be summarized as follows:

\begin{itemize}

\item Proposition of the RB-CCR resampling algorithm, which employs the radial-based approach to calculate the class potential, so that a classifier trained on modified data improves recall and has less impact on precision.

\item Analysis of the impact of sampling region on algorithms behavior and performance.

\item Showing that the proposed method can outperform the quality of the CCR algorithm.

\item Experimental evaluation of the proposed approach based on diverse benchmark datasets and a detailed comparison with the \emph{state-of-the-art} approaches.

\end{itemize}

The paper is organized as follows. The next section discusses the related work and situates RB-CCR concerning the state-of-the-art in imbalanced binary classification. Section \ref{sec:algorithm}, provides the details of CCR and RB-CCR, demonstrates resampling with RB-CCR and contrasts its run-time complexity with that of CCR. In Section \ref{sec:experiments}, we describe the experimental setup, report the results along with our analysis, and finally, Section \ref{sec:conclusion} includes our concluding remarks and a discussion of future work.

\section{Related work}


\emph{Imbalance ratio} ($IR$) \cite{GARCIA201213} is defined as the \rev{ratio between the number of majority and minority class observations. A moderate to high $IR$ (typically greater than $10:1$) can pose a significant challenge to learning a sufficiently accurate classifier across all classes. This is particularly the case when it is combined with other adverse data properties, such as class overlap, sparsity, complex clustering, and noise \cite{he2009learning,Napierala:2012}. In such cases, the classifier is at great risk of becoming biased towards the majority class \cite{he2009learning}, and / or \emph{overfitting} the training data \cite{Chen:2008}}. Problems of this nature are a focus of intense research \cite{chawla2002smote,bunkhumpornpat2009safe,Kubat:1997}.

\rev{Measuring the quality of a model on imbalanced data requires some attention. It is well-known that using classic metrics, such as accuracy and error rate, on imbalanced datasets can cause misleading interpretations of the efficacy of the model \cite{jeni2013facing}. As a result, the imbalanced learning community has shifted to use metrics, such as $precision$, $recall$ ($sensitivity$), $specificity$, \emph{G-mean}, $F_{\beta} score$, and $AUC$  \cite{kubat1997learning,Krawczyk:2016}. More recently, however, it has been noted that the widely used metrics $F_{\beta} score$, and $AUC$ can be sub-optimal for evaluating performance on imbalanced data. Brzeziński et al. \cite{brzezinski2019dynamics} demonstrated that $F_{\beta} score$ is usually more biased towards the majority class than AUC and G-mean. The flaws of $F_{\beta} score$ are also discussed in a study by Hand and Christen \cite{hand2018note}, in which authors suggest that to make a fair comparison, precision and recall have to be weighed separately for each problem, depending on the imbalance ratio. Alternatively, the authors in \cite{davis2006relationship} argue that ROC curves, and AUC by extension, can present an overly optimistic view of an algorithm’s performance if there is a large skew.} 

Classification strategies to deal with imbalanced data can be divided into three main groups \cite{Lopez:2012}: inbuilt mechanism, data-level methods, and hybrid methods.

\textbf{Inbuilt mechanisms.} In this approach, existing classification algorithms are adapted to imbalanced problems by ensuring balanced accuracy for instances from both classes. \rev{Two of the most popular areas of research of these methods are:} using one-class classification \cite{Japkowicz:1995}, where the goal is to learn the minority class decision boundaries, and because of the frequently assumed regular, closed shape of the decision borders it is adequate for the clusters created by minority classes \cite{Krawczyk:2014ins}. Secondly, algorithms employing kernel functions \cite{Mathew:2018}, splitting criteria in decision trees \cite{Li:2018}, to make them cost-sensitive methods employing different forms of the loss function \cite{Khan:2018}, where the algorithm assigns a higher misclassification cost for instances from the minority class \cite{Krawczyk:2014,Lopez:2012,he2009learning,Zhou:2006}. Unfortunately, such methods can cause a reverse bias towards the minority class. 
Worth noting are methods based on ensemble classification \cite{Wozniak:2014}, like \emph{\textsc{smote}Boost} \cite{Chawla:2003} and \emph{AdaBoost.NC} \cite{Wang:2010}, or \emph{Multi-objective Genetic Programming-Based Ensemble} \cite{Bhowan:2013}.

\textbf{Data-level methods}. This work focuses on data preprocessing to reduce imbalance ratio by decreasing the number of majority observations (undersampling) or increasing minority observations (oversampling). \rev{After applying such preprocessing, the data can be classified using traditional learning algorithms.}
The most straightforward approaches to dealing with the imbalanced data are \emph{Random Oversampling} (ROS) and \emph{Random Undersampling} (RUS). When applying ROS, new minority class instances are generated by duplicating randomly chosen minority instances. This procedure can create small, dense clusters of replicated minority objects leading to overfitting. The most recognized data-level method is the SMOTE \cite{chawla2002smote} algorithm. It reduces the risk of overfitting by generating synthetic minority instances via random interpolation in-between existing minority objects. 

The well-studied limitations of SMOTE have inspired many new synthetic oversampling techniques, such as \cite{Perez-Ortiz:2016,Bellinger:2018}. The most significant shortcomings of SMOTE are that it assumes a homogeneous minority class cluster, and it does not consider the majority objects in the neighborhood when generating synthetic objects. In cases where the minority class forms many small disjointed clusters, SMOTE may cause an increase the class overlapping, and thus, the complexity of the classification problem \cite{Krawczyk:2019}. Numerous methods have been proposed to address these weaknesses by considering both classes during generation, or as a \textit{post-hoc} cleaning step.

Safe-level SMOTE~\cite{bunkhumpornpat2009safe} and LN-SMOTE~\cite{Maciejewski:2011} are specifically designed to reduce the risk of introducing noisy synthetic observations inside the majority class region. Other SMOTE alternatives aim to focus the generation process on challenging regions of the dataspace. Borderline-SMOTE \cite{Han:2005}, for example, focuses the process of synthetic observation generation on the instances close to the class boundary, and ADASYN \cite{he2008adasyn} prioritizes the difficult instances. The SWIM \cite{sharma2018synthetic} method uses the Mahalanobis distance to determine the best position for synthetic samples, taking into account the existing samples from both classes. \emph{Radial-Based Oversampling} (RBO) \cite{koziarski2019radial} is a method that employs potential estimation to generate new minority objects using radial basis functions. The \emph{Combined Cleaning and Resampling} (CCR) \cite{koziarski2017ccr} method combines two techniques -- cleaning the decision border around minority objects and guided synthetic oversampling. 

RUS preprocesses the data by randomly removing majority class samples. It is conceptually simple and risks removing important objects from the majority class. This can cause the induced classifier to underfit less dense majority class clusters. Guided undersampling approaches aim to avoid this by analyzing the minority and majority class instances in the local neighborhood. Edited Nearest Neighbor, for example, removes majority examples if their set of three nearest neighbors does not include at least one other majority object. Radial-Based Undersampling, on the other hand, employs the concept of mutual class potential to direct undersampling \cite{koziarski2020radial}. Koziarski introduced \emph{Synthetic Minority Undersampling Technique} (SMUTE), which leverages the concept of interpolation of nearby instances, previously introduced in the oversampling setting in SMOTE \cite{koziarski2020csmoute}.

\textbf{Hybrid methods.} Data preprocessing methods can be combined with in-built classification methods for imbalanced learning. Galar \textit{et al.} proposed to hybridize \emph{under-} and \emph{oversampling} with an ensemble of classifiers \cite{Galar:2012}. This approach allows the data to be independently processed for each of the base models. It is worth also mentioning SMOTEBoost, which is based on a combination of the SMOTE algorithm and the boosting procedure \cite{Chawla:2003}. In addition, the \emph{Combined Synthetic Oversampling and Undersampling Technique} (CSMOUTE) integrates SMOTE oversampling with SMUTE undersampling \cite{koziarski2020csmoute}.

\section{Radial-Based Combined Cleaning and Resampling} \label{sec:algorithm}

In this paper, we propose an extension to the original CCR \cite{koziarski2017ccr} algorithm that refines its sampling procedure. In short, CCR is an energy-based oversampling algorithm that relies on spherical regions, centered around the minority class observations, to designate areas in which synthetic minority observations should be generated. These spherical regions expand iteratively, with the rate of expansion inversely proportional to the number of neighboring observations belonging to the majority class, while computationally efficient and conceptually simple, using spherical regions to model the areas designed for oversampling has two limitations. First of all, it enforces a constant rate of expansion of the sphere in every direction, regardless of the majority neighbors' exact position. Secondly, it does not utilize the information about the neighboring minority class observations. We propose a novel sampling procedure to address these issues, which is refining the original spherical regions. In the remainder of this section, we describe the proposed sampling procedure and its integration with the CCR algorithm.

\subsection{Guided sampling procedure}

We base the proposed sampling procedure on the notion of class potential, previously used in the imbalanced data setting by Krawczyk et al. \cite{Krawczyk:2019}. The potential function is a real-valued function that, in a given point in space $x$, measures the cumulative closeness to a given collection of observations $\mathcal{X}$. More formally, using a Gaussian radial basis function with a spread $\gamma$, a potential function can be defined as
\begin{equation}
\Phi(x, \mathcal{X}, \gamma) = \sum_{i=1}^{\mid \mathcal{X} \mid}{e^{-\left(\frac{\lVert \mathcal{X}_i - x \rVert_2}{\gamma}\right)^{2}}}\label{eq:potential}.
\end{equation}

Of particular interest in the imbalanced data oversampling task will be the potential computed concerning either the collection of majority class observations $\mathcal{X}_{maj}$ (\textit{majority class potential}), or minority class observations $\mathcal{X}_{min}$ (\textit{minority class potential}). Such class potential can be regarded as a measure reflecting the degree of certainty we assign to $x$ being a member of either the majority or the minority class. It can also be used to model the regions of interest in which oversampling is to be conducted, which was previously demonstrated in Radial-Based Oversampling (RBO) \cite{Krawczyk:2019} and Sampling With the Majority (SWIM) \cite{bellinger2019framework} algorithms. SMOTE and its derivatives 
define the regions of interest as the lines connecting nearby minority observations. Also, the probability of sampling within any given region of interest is typically uniform. Alternatively, using class potential, as proposed here, offers an informationally richer framework. 
First of all, by using the majority class potential, we can leverage the information about the position of majority observations, which is not used by SMOTE. Secondly, when using potential, we are not constrained to sampling from within a set a lines. Rather, we can sample smoothly from the space around the minority observations. Moreover, the sampling region is non-linear, which enables it to better adapt to the underlying data distribution. 

To reiterate, the drawbacks of the original CCR algorithm are that the sphere expansion procedure progresses at a constant rate in every direction, regardless of the exact position of the majority neighbors, and it does not utilize the information about the position of neighboring minority class observations. Intuitively, neither of these is the desired behavior since it can lead to a lower than expected expansion in the direction of minority observation clusters and higher than expected expansion in the direction of majority observation clusters. While in theory, an obvious modification that could address these issues would be to exchange the spheres used by CCR to more robust shapes, such as ellipsoids, and adjust the expansion step accordingly, in practice, it is not clear how the latter could be achieved. Alternatively, we propose to exploit the efficiency of first defining the sphere around the minority observation and then partitioning it into sub-regions based on the class potential to more effectively guide sample generation. 

\begin{figure*}[!htb]
\centering
\includegraphics[width=0.3\textwidth]{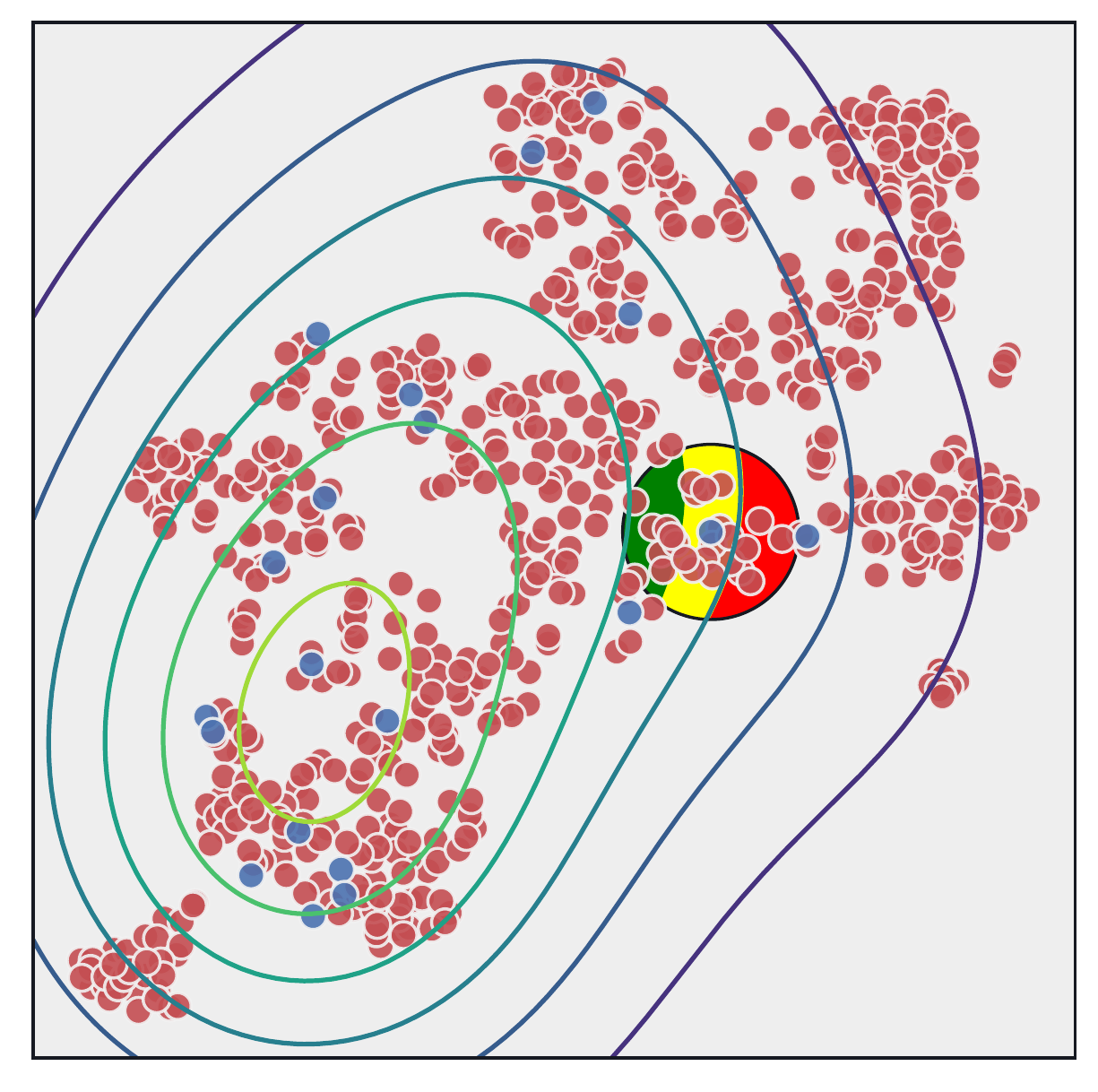}
\caption{An example of a sphere generated around a specific minority observation, partitioned into three regions: high potential (H), indicated with a green color, equal potential (E), indicated with a yellow color, and low potential (L), indicated with a red color. Note that the shape of the regions aligns with that of the produced potential field, indicated with a contour plot.}
\label{fig:example-regions-local}
\end{figure*}

The proposed strategy partitions a given sphere into three target regions, low (L), equal (E), and high (H), based on the class potential. Synthetic samples are generated in a user-specificity target region by randomly generating candidates with uniform probability throughout the sphere. A random subset of these is selected from the target region and added to the training set. The target region and number of samples are specified as parameters of the algorithm. A more detailed formulation of the proposed strategy is presented in Algorithm~\ref{algorithm:sampling}, and an illustration of the sphere partitioning procedure is presented in Figure~\ref{fig:example-regions-local}.


\begin{algorithm}[!htb]
	\caption{Guided sampling procedure}
	\textbf{Input:} sampling seed $x$, sampling radius $r$, collection of minority observations $\mathcal{X}_{min}$ \\
	\textbf{Parameters:} radial basis function spread $\gamma$, sampling $region$ from which returned samples will be drawn, number of candidates $c$ used for potential range estimation, number of returned candidate samples $n$ \\
    \textbf{Output:} collection of synthetic minority observations $S$ located in the sampling $region$ around $x$
		
	\label{algorithm:sampling}	
	\vspace{-0.5\baselineskip}
	
	\hrulefill
	\begin{algorithmic}[1]
    	\STATE \textbf{function} sample($x$, $r$, $\mathcal{X}_{min}$, $\gamma$, $region$, $c$, $n$):
    	\STATE \rev{$\Phi^0 \gets \Phi(x, \mathcal{X}_{min}, \gamma)$ \COMMENT{seed score}}
    	\STATE $C \gets \emptyset$ \COMMENT{collection of candidate samples}
    	\STATE $Z \gets \emptyset$ \COMMENT{collection of candidate potentials}
    	\FOR{\textit{i} \textbf{in} 1 \textbf{to} $c$}
		\STATE $C_i \gets$ random sample inside a $x$-centered sphere with radius $r$
		\STATE $Z_i \gets \Phi(C_i, \mathcal{X}_{min}, \gamma)$
        \ENDFOR
	    \STATE \rev{$bound_L \gets \Phi^0 - \frac{1}{3}(\Phi^0 - \min Z)$ \COMMENT{estimated low potential bound within the sphere}}
	    \STATE \rev{$bound_H \gets \Phi^0 + \frac{1}{3}(\max Z - \Phi^0)$ \COMMENT{estimated high potential bound within the sphere}}
	    \STATE $S \gets \{x\}$ \COMMENT{collection of suitable candidates}
	    \FOR{\textit{i} \textbf{in} 1 \textbf{to} $c$}
		\IF{$Z_i \leq bound_L$}
		\STATE $reg_i \gets L$ \COMMENT{$i$-th candidate in the low potential region}
		\ELSIF{$Z_i \geq bound_H$}
		\STATE $reg_i \gets H$ \COMMENT{$i$-th candidate in the high potential region}
		\ELSE
		\STATE $reg_i \gets E$ \COMMENT{$i$-th candidate in the equal potential region}
		\ENDIF
		\IF{$reg_i = region$}
		\STATE $S \gets S \cup \{C_i\}$
		\ENDIF
        \ENDFOR
	    \STATE $S \gets n$ samples randomly selected with replacement from $S$ 
	\STATE \textbf{return} $S$
	\end{algorithmic}
\end{algorithm}

The CCR algorithm generates samples with uniform probability from within entire sphere. Alternatively, Figure \ref{fig:example-regions-local} illustrates that RB-CCR divides the original sphere into three regions (L, E, H). The regions are defined according to the shape of the globally calculated minority class potential. Subsequent to the partitioning, sample generation can be restricted to a specific region. 
Intuitively, samples in the high potential regions can be regarded as having a higher probability of coming from the underlying minority class distribution than samples in the low potential regions.
This, to some extents, parallels different variants of SMOTE, such as Borderline-SMOTE \cite{han2005borderline} or Safe-Level-SMOTE \cite{bunkhumpornpat2009safe}, which focus on different types of observations to guide the sampling process. However, contrary to SMOTE variants, RB-CCR provides a flexibility to chose an appropriate sampling region for the target data within a single framework. 

\subsection{Integrating guided sampling with the CCR algorithm}


\rev{We begin with a brief description of the original CCR algorithm, as described in \cite{KOZIARSKI2020106223}, where more in-depth discussion of the design choices can be found. The algorithm itself consists of two main steps: cleaning the neighborhood of the minority observations, and second of all, selectively oversampling in the produced, cleaned regions. After describing the original algorithm, we discuss how it can be integrated with the proposed guided sampling procedure.}

\begin{figure}
\centering
\includegraphics[width=0.3\linewidth]{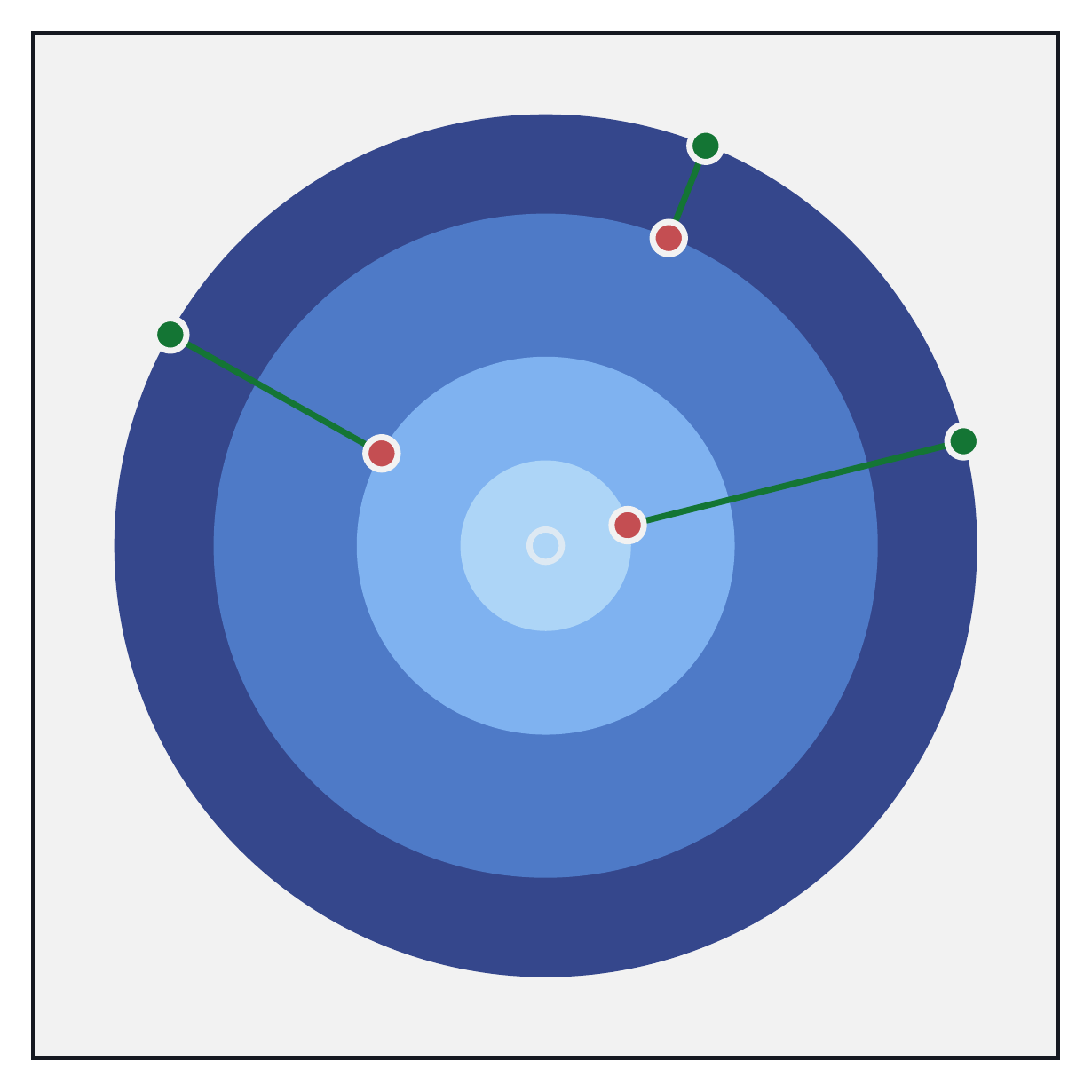}
\caption{\rev{An illustration of the sphere creation for an individual minority observation (in the center) surrounded by majority observations (in red). Sphere expends at a normal cost until it reaches a majority observation, at which point the further expansion cost increases (depicted by blue orbits with an increasingly darker color). Finally, after the expansions, the majority observations within the sphere are being pushed outside (in green). Source: \cite{KOZIARSKI2020106223}.}}
\label{fig:cleaning}
\end{figure}

\rev{\noindent\textbf{Cleaning the minority neighborhoods.} First step of the proposed approach is cleaning the minority class neighborhoods from the majority observations. This is achieved via an energy-based approach, in which spherical regions are being designated for cleaning. The size of the regions is constrained by the presence of majority neighbors and is determined in an iterative procedure, during which spheres expand up to the point of depleting the allocated energy budget. More formally, for a given minority observation denoted by $x_i$, current radius of an associated sphere denoted by $r_i$, a function returning the number of majority observations inside a sphere centered around $x_i$ with radius $r$ denoted by $f_n(r)$, a target radius denoted by $r_i'$, and $f_n(r_i') = f_n(r_i) + 1$, we define the energy change caused by the expansion from $r_i$ to $r_i'$ as
\begin{equation}
    \Delta e = - (r_i' - r_i) \cdot f_n(r_i').
\end{equation}
During the sphere expansion procedure, the radius of a given sphere increases up to the point of completely depleting the energy, with the cost increasing after each encountered majority observation. Finally, the majority observations inside the sphere are being pushed out to its outskirts. The whole process was illustrated in Figure~\ref{fig:cleaning}.}

\rev{\noindent\textbf{Selectively oversampling the minority class.} After the cleaning stage is completed, new synthetic minority observations are being generated in the produced spherical regions. The ratio of the synthetic observations generated around a given minority observation is proportional to the sphere's radius, calculated in the previous step. More formally, for a given minority observation denoted by $x_i$, the radius of an associated sphere denoted by $r_i$, the vector of all calculated radii denoted by $r$, collection of majority observations denoted by $\mathcal{X}_{maj}$, collection of minority observations denoted by $\mathcal{X}_{min}$, and assuming that the oversampling is performed up to the point of achieving balanced class distribution, we define the number of synthetic observations to be generated around $x_i$ as
\begin{equation}
    \label{eq:prop}
    g_i = \lfloor\dfrac{r_i^{-1}}{\sum_{k = 1}^{|\mathcal{X}_{min}|}{r_k^{-1}}} \cdot (|\mathcal{X}_{maj}| - |\mathcal{X}_{min}|)\rfloor.
\end{equation}
This procedure can be interpreted as weighing the difficult observations more heavily, similar to the technique used in ADASYN \cite{he2008adasyn}. The difficulty of observation is determined based on the proximity of nearest majority observations: minority observations with nearby majority neighbors will have a constrained sphere radius, which will result in a higher allocation of produced synthetic observations.}


\rev{\textbf{Combining guided sampling with CCR.}} The proposed sampling strategy can easily be integrated into the original CCR algorithm. Instead of the original sampling within the whole sphere, RB-CCR uses the guided sampling strategy described in the previous section. In initial steps of RB-CCR are the same as CCR. Specifically, they are sphere radius calculation, translation of majority observations, and calculation of the number of synthetic observations generated for each minority observations. We present pseudocode of the proposed RB-CCR algorithm in Algorithm~\ref{algorithm:rb-ccr}. \rev{It should be noted that, except for the addition of a guided sampling procedure, the algorithm is presented as it was previously proposed in \cite{KOZIARSKI2020106223}.}

\begin{algorithm}[!htb]
	\caption{Radial-Based Combined Cleaning and Resampling}
	\textbf{Input:} collections of majority observations $\mathcal{X}_{maj}$ and minority observations $\mathcal{X}_{min}$ \\
	\textbf{Parameters:} $energy$ budget for expansion of each sphere, radial basis function spread $\gamma$, sampling $region$ from which returned samples will be drawn, number of candidates $c$ used for potential range estimation \\
    \textbf{Output:} collections of translated majority observations $\mathcal{X}_{maj}'$ and synthetic minority observations $S$
		
	\label{algorithm:rb-ccr}	
	\vspace{-0.5\baselineskip}
	
	\hrulefill
	\begin{algorithmic}[1]
		\STATE \textbf{function} RB-CCR($\mathcal{X}_{maj}$, $\mathcal{X}_{min}$, $energy$, $\gamma$, $region$, $c$):
		\STATE $S \gets \emptyset$ \COMMENT{synthetic minority observations}
		\STATE $t \gets $ zero matrix of size $|\mathcal{X}_{maj}| \times m$, with $m$ denoting the number of features \COMMENT{translations of majority observations}
		\STATE $r \gets $ zero vector of size $|\mathcal{X}_{min}|$ \COMMENT{radii of spheres associated with the minority observations}
        \FORALL{minority observations $x_i$ in $\mathcal{X}_{min}$}
        \STATE $e$ $\gets$ $energy$ \COMMENT{remaining energy budget}
        \STATE $n_r \gets 0$ \COMMENT{number of majority observations inside the sphere generated around $x_i$}
        \FORALL{majority observations $x_j$ in $\mathcal{X}_{maj}$}
        \STATE $d_j \gets \lVert x_i - x_j \rVert_2$
        \ENDFOR
        \STATE sort $\mathcal{X}_{maj}$ with respect to $d$
        \FORALL{majority observations $x_j$ in $\mathcal{X}_{maj}$}
        \STATE $n_r \gets n_r + 1$
        \STATE $\Delta e \gets - (d_j - r_i) \cdot n_r$
        \IF{$e + \Delta e > 0$}
        \STATE $r_i \gets d_j$
        \STATE $e \gets e + \Delta e$
        \ELSE
        \STATE $r_i \gets r_i + \frac{e}{n_r}$
        \STATE \textbf{break}
        \ENDIF
        \ENDFOR
        \FORALL{majority observations $x_j$ in $\mathcal{X}_{maj}$}
        \IF{$d_j < r_i$}
        \STATE $t_j \gets t_j + \dfrac{r_i - d_j}{d_j} \cdot (x_j - x_i)$
        \ENDIF
        \ENDFOR
        \ENDFOR
        \STATE $\mathcal{X}_{maj}' \gets \mathcal{X}_{maj} + t$
        \FORALL{minority observations $x_i$ in $\mathcal{X}_{min}$}
        \STATE $g_i \gets \lfloor\dfrac{r_i^{-1}}{\sum_{k = 1}^{|\mathcal{X}_{min}|}{r_k^{-1}}} \cdot (|\mathcal{X}_{maj}| - |\mathcal{X}_{min}|)\rfloor$ \label{op:prop} 
		\STATE add sample($x_i$, $r_i$, $\mathcal{X}_{min}$, $\gamma$, $region$, $c$, $g_i$) to $S$
        \ENDFOR
		\STATE \textbf{return} $\mathcal{X}_{maj}'$, $S$
	\end{algorithmic}
\end{algorithm}

\begin{figure*}[!htb]
\centering
\begin{subfigure}[b]{0.23\textwidth}
  \includegraphics[width=\textwidth]{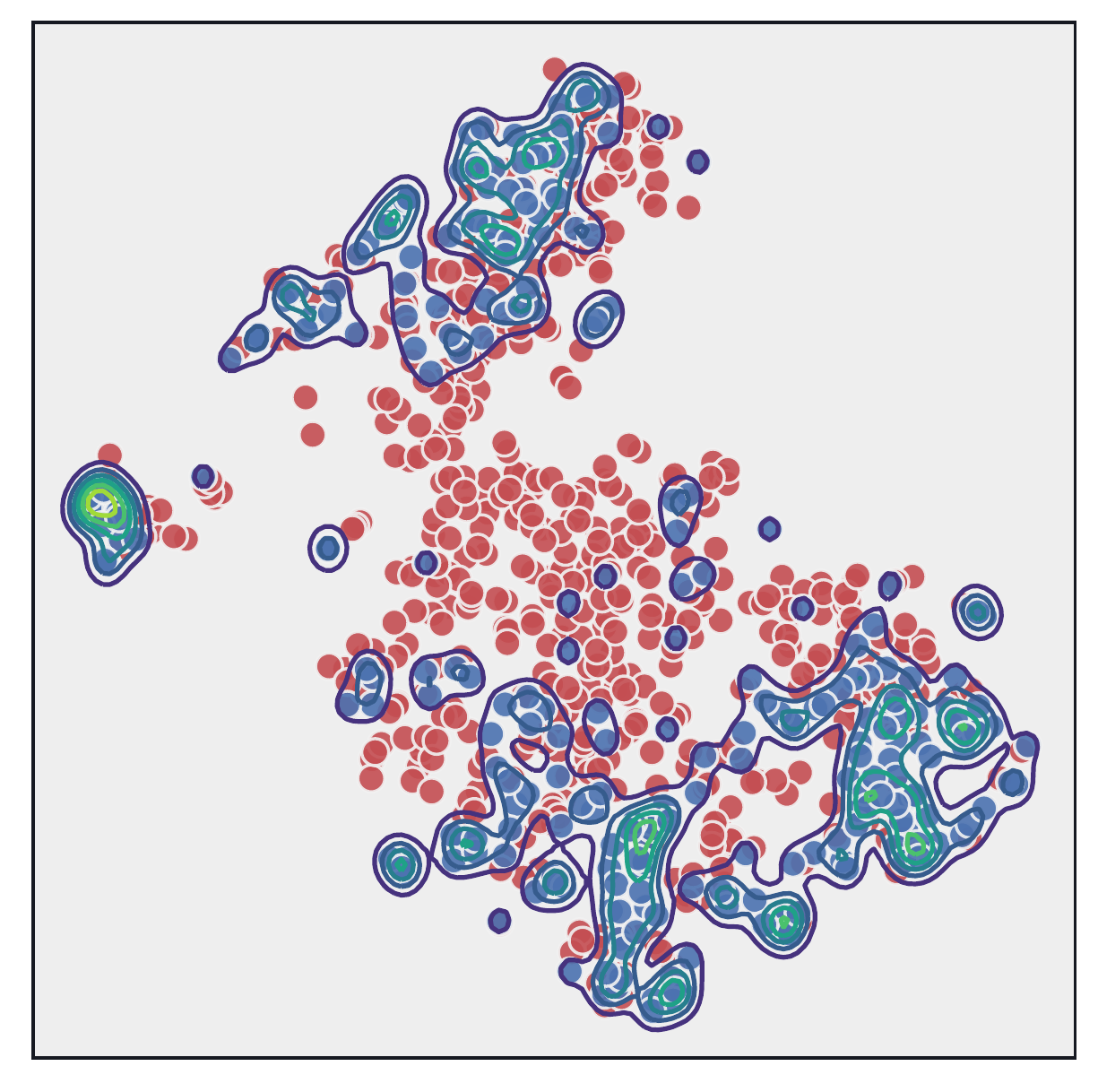}
  \caption{$\gamma = 0.1$}
\end{subfigure}
~
\begin{subfigure}[b]{0.23\textwidth}
  \includegraphics[width=\textwidth]{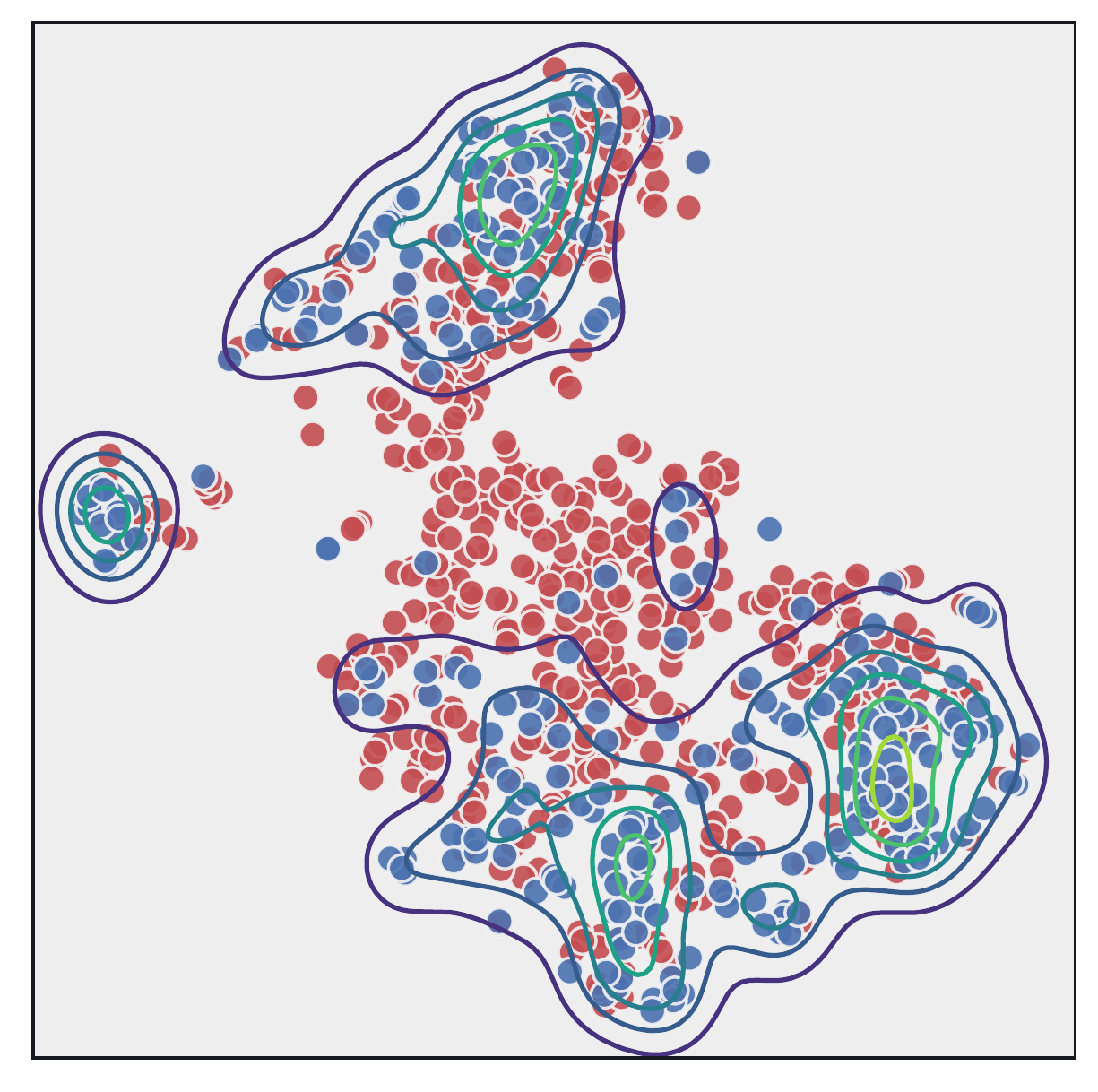}
  \caption{$\gamma = 0.25$}
\end{subfigure}
~
\begin{subfigure}[b]{0.23\textwidth}
  \includegraphics[width=\textwidth]{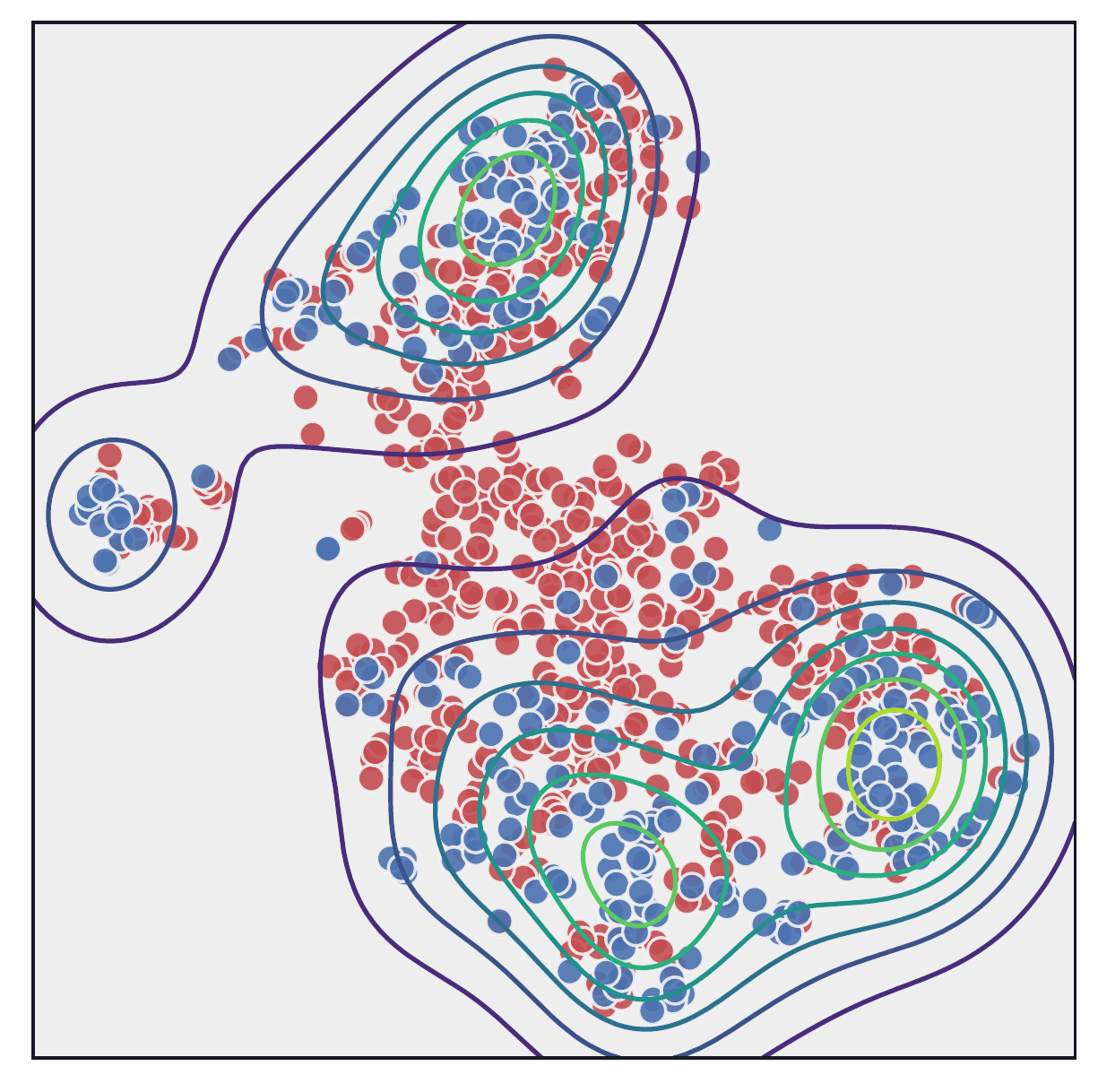}
  \caption{$\gamma = 0.5$}
\end{subfigure}
~
\begin{subfigure}[b]{0.23\textwidth}
  \includegraphics[width=\textwidth]{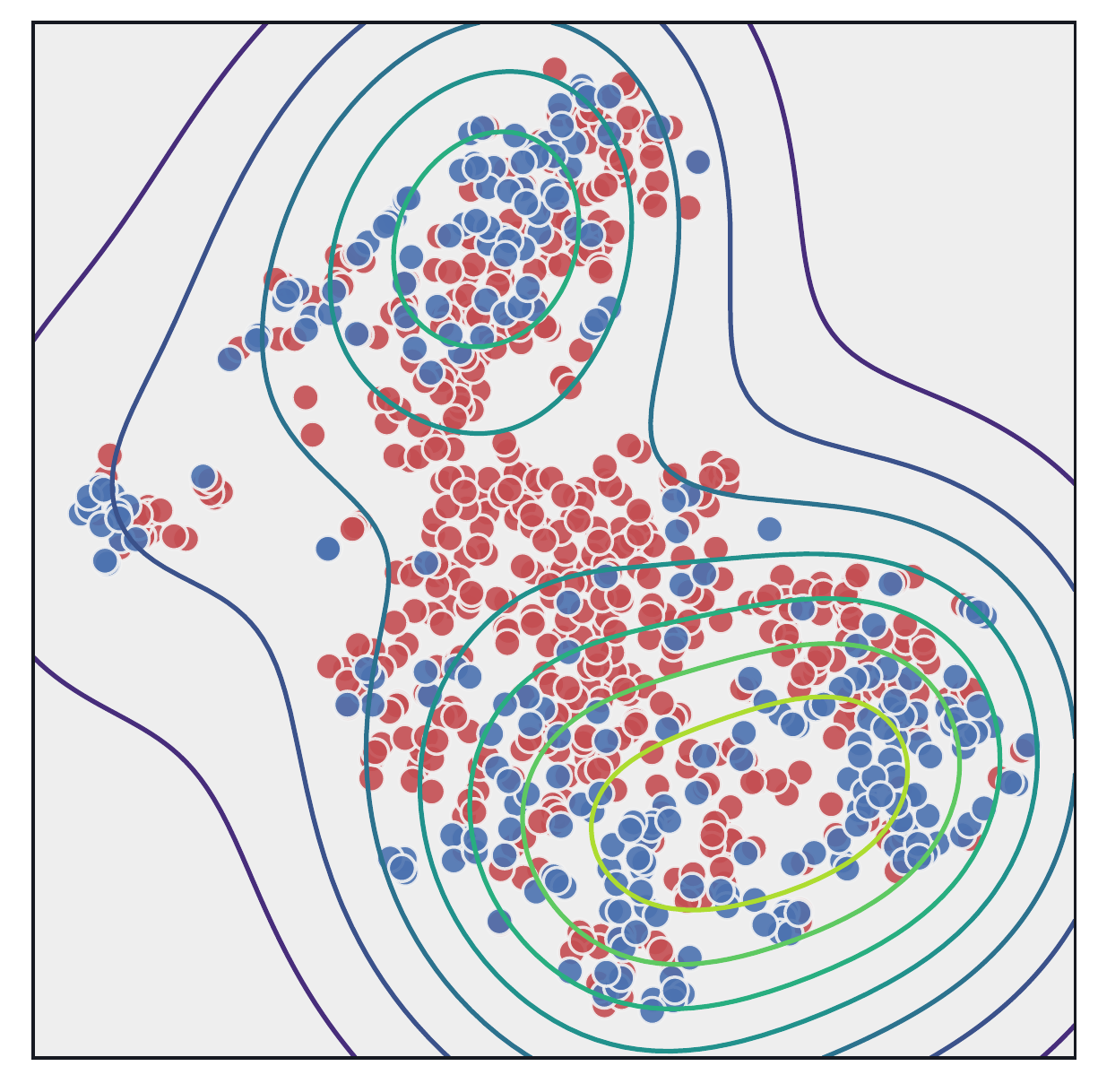}
  \caption{$\gamma = 1.0$}
\end{subfigure}
\caption{Visualization of the impact of $\gamma$ parameter on the shape of minority class potential.}
\label{fig:example-potential}
\end{figure*}

\begin{figure*}[!htb]
\centering
\begin{subfigure}[b]{0.23\textwidth}
  \includegraphics[width=\textwidth]{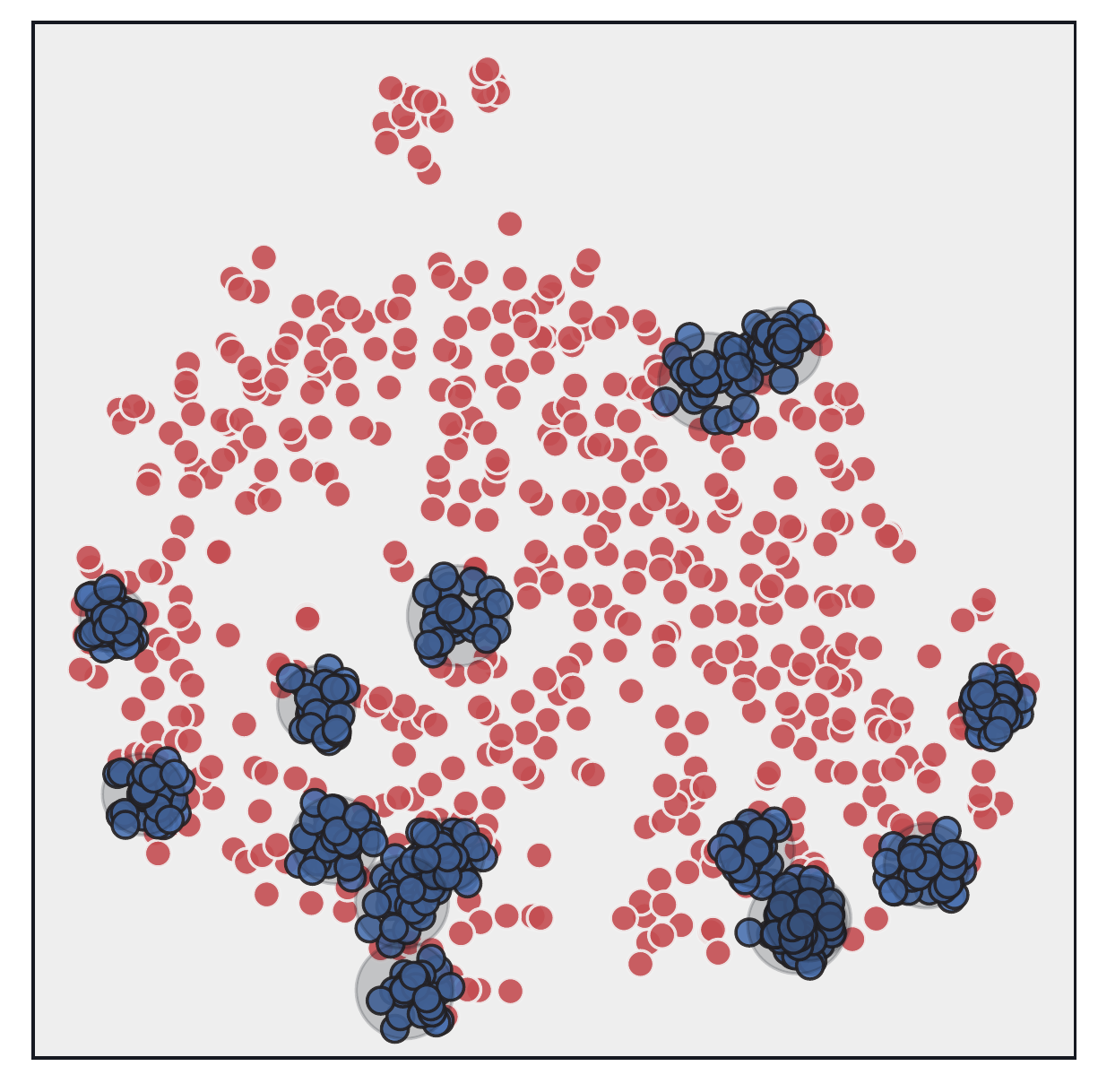}
  \caption{energy $= 0.1$}
\end{subfigure}
~
\begin{subfigure}[b]{0.23\textwidth}
  \includegraphics[width=\textwidth]{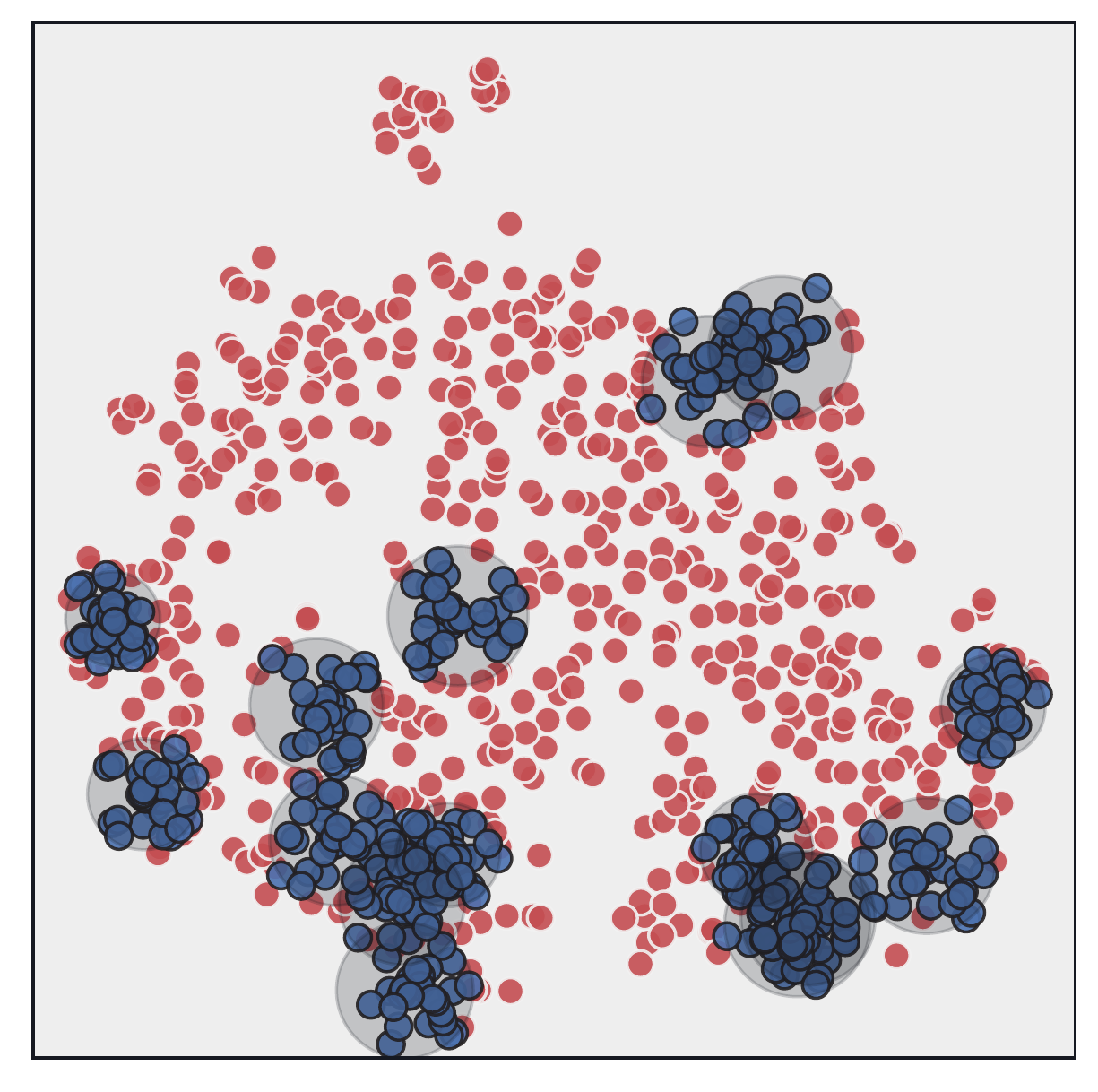}
  \caption{energy $= 0.25$}
\end{subfigure}
~
\begin{subfigure}[b]{0.23\textwidth}
  \includegraphics[width=\textwidth]{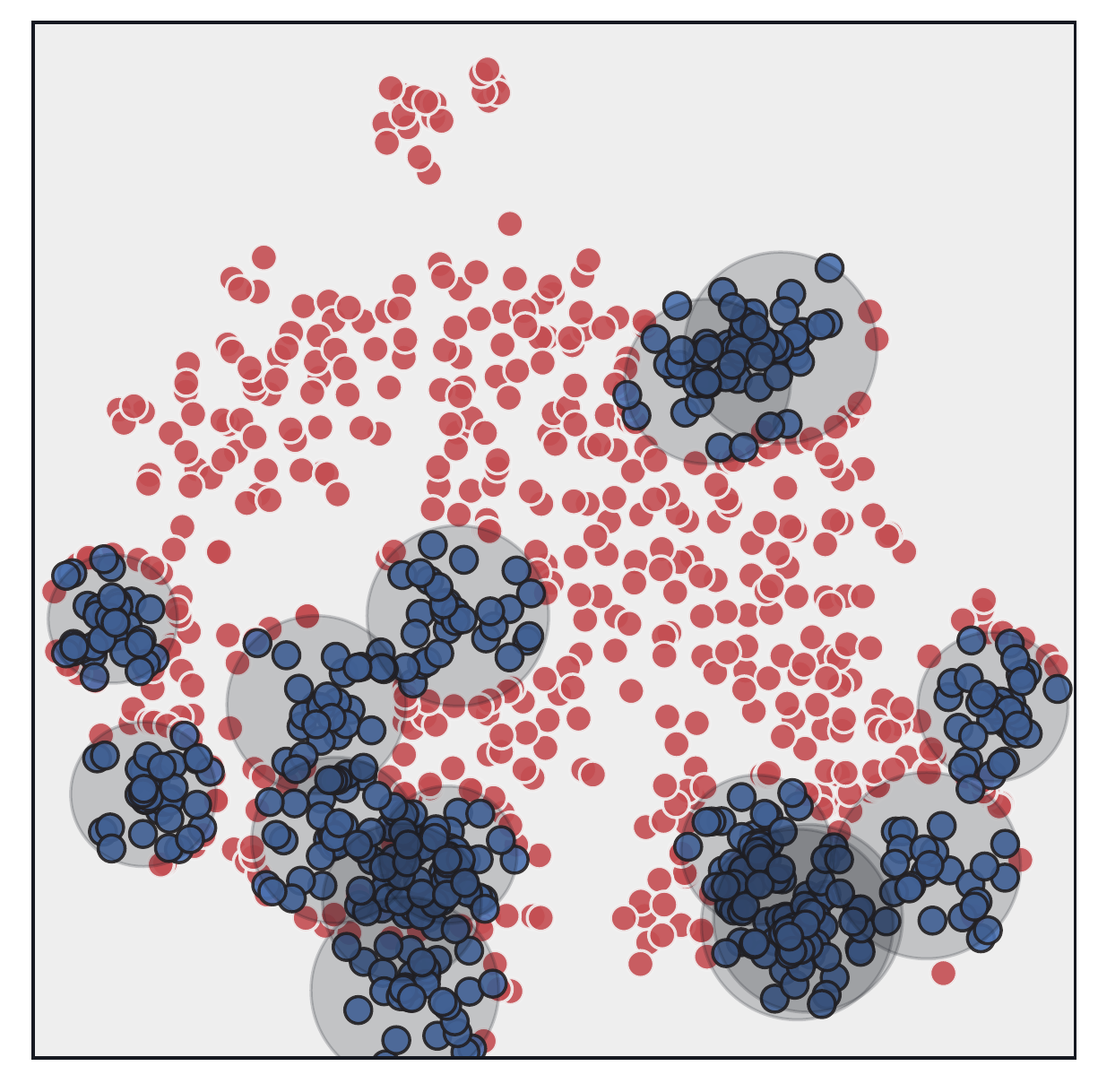}
  \caption{energy $= 0.5$}
\end{subfigure}
~
\begin{subfigure}[b]{0.23\textwidth}
  \includegraphics[width=\textwidth]{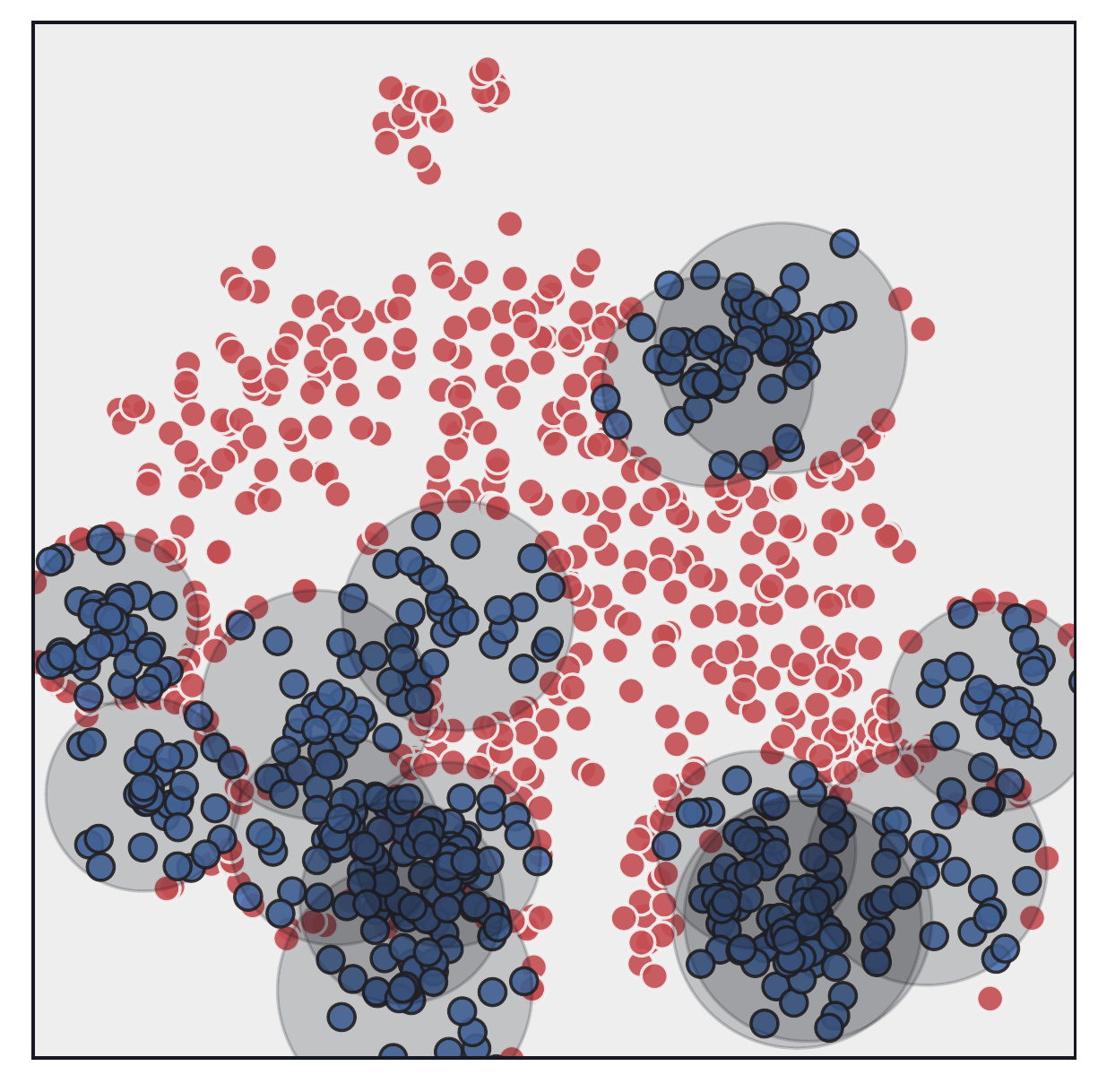}
  \caption{energy $= 1.0$}
\end{subfigure}
\caption{Visualization of the impact of $energy$ parameter on the sphere radius and corresponding region in which synthetic minority observations (indicated by dark outline) are being generated. Note that the majority observations within the sphere are being pushed outside during the cleaning step.}
\label{fig:example-sphere-radius}
\end{figure*}

\begin{figure*}[!htb]
\centering
\begin{subfigure}[b]{0.23\textwidth}
  \includegraphics[width=\textwidth]{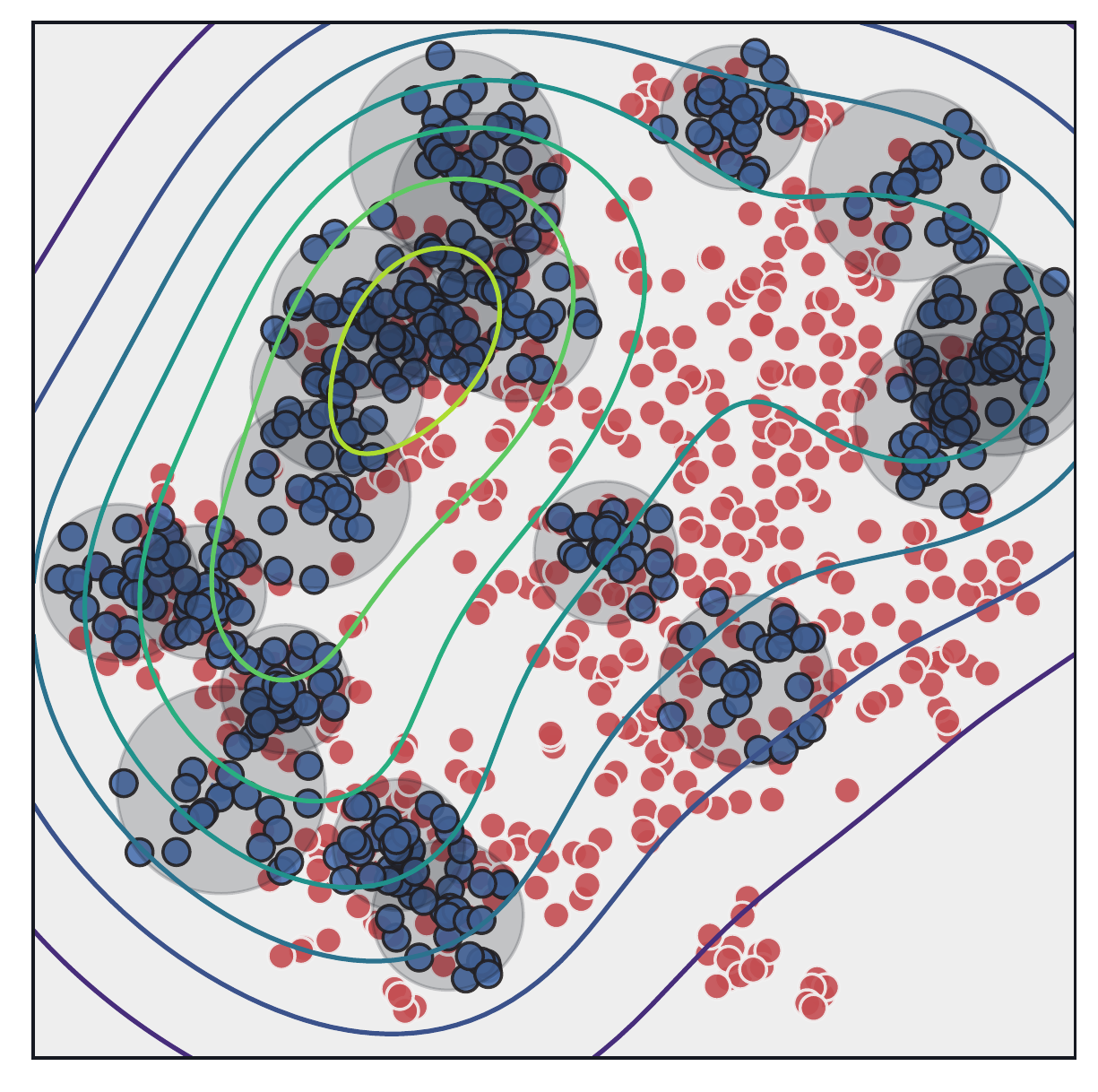}
  \caption{LEH}
\end{subfigure}
~
\begin{subfigure}[b]{0.23\textwidth}
  \includegraphics[width=\textwidth]{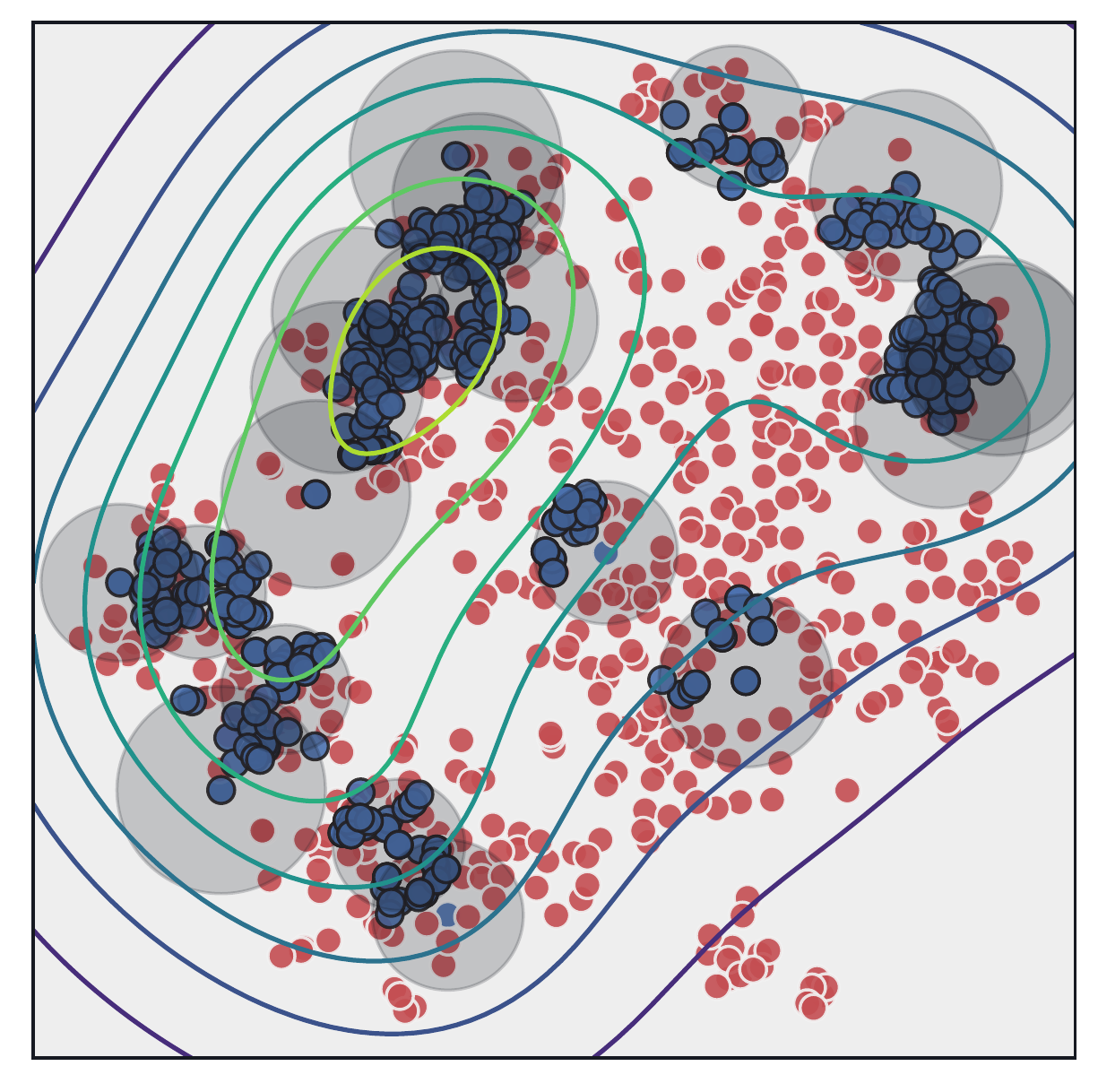}
  \caption{H}
\end{subfigure}
~
\begin{subfigure}[b]{0.23\textwidth}
  \includegraphics[width=\textwidth]{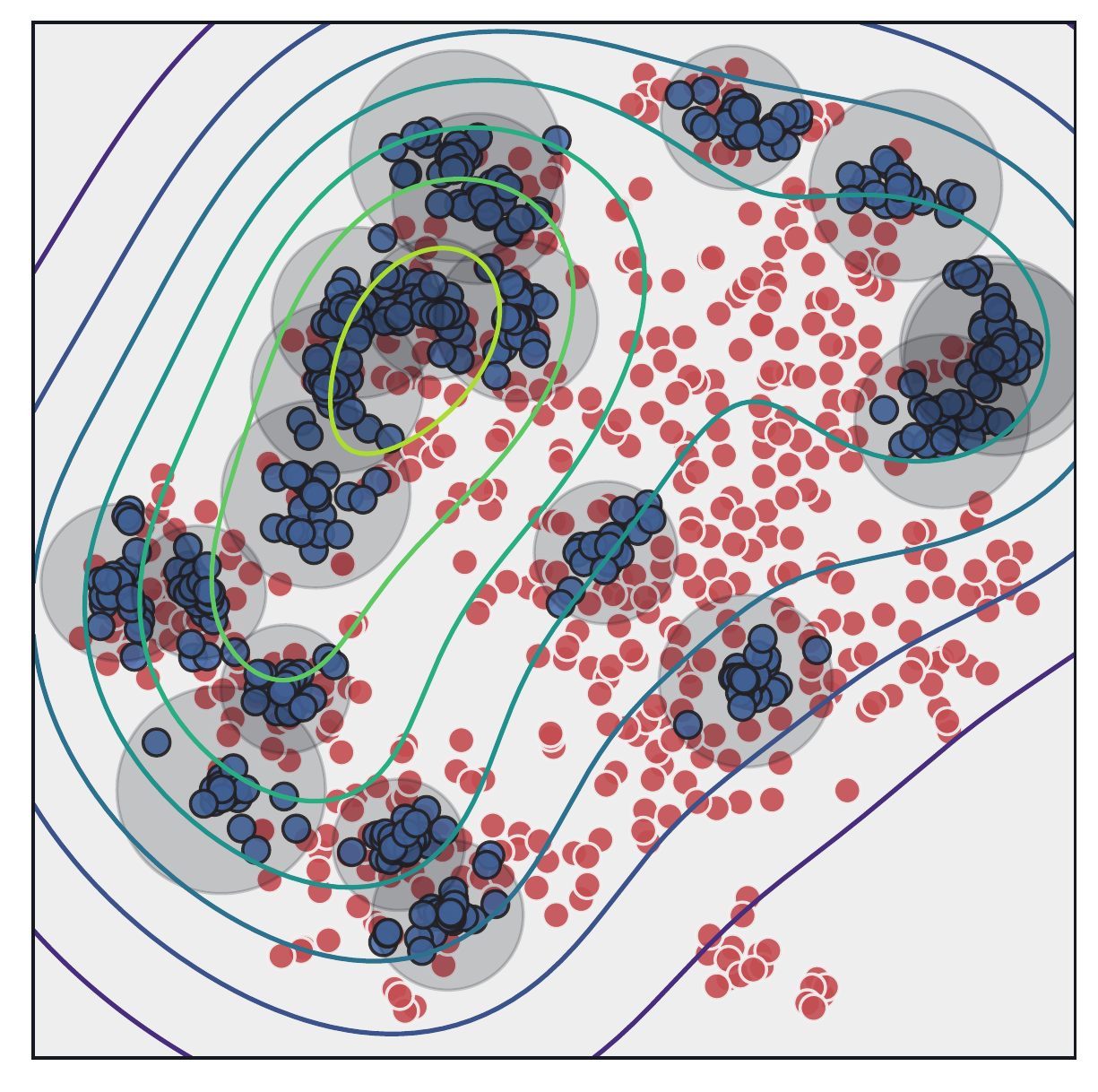}
  \caption{E}
\end{subfigure}
~
\begin{subfigure}[b]{0.23\textwidth}
  \includegraphics[width=\textwidth]{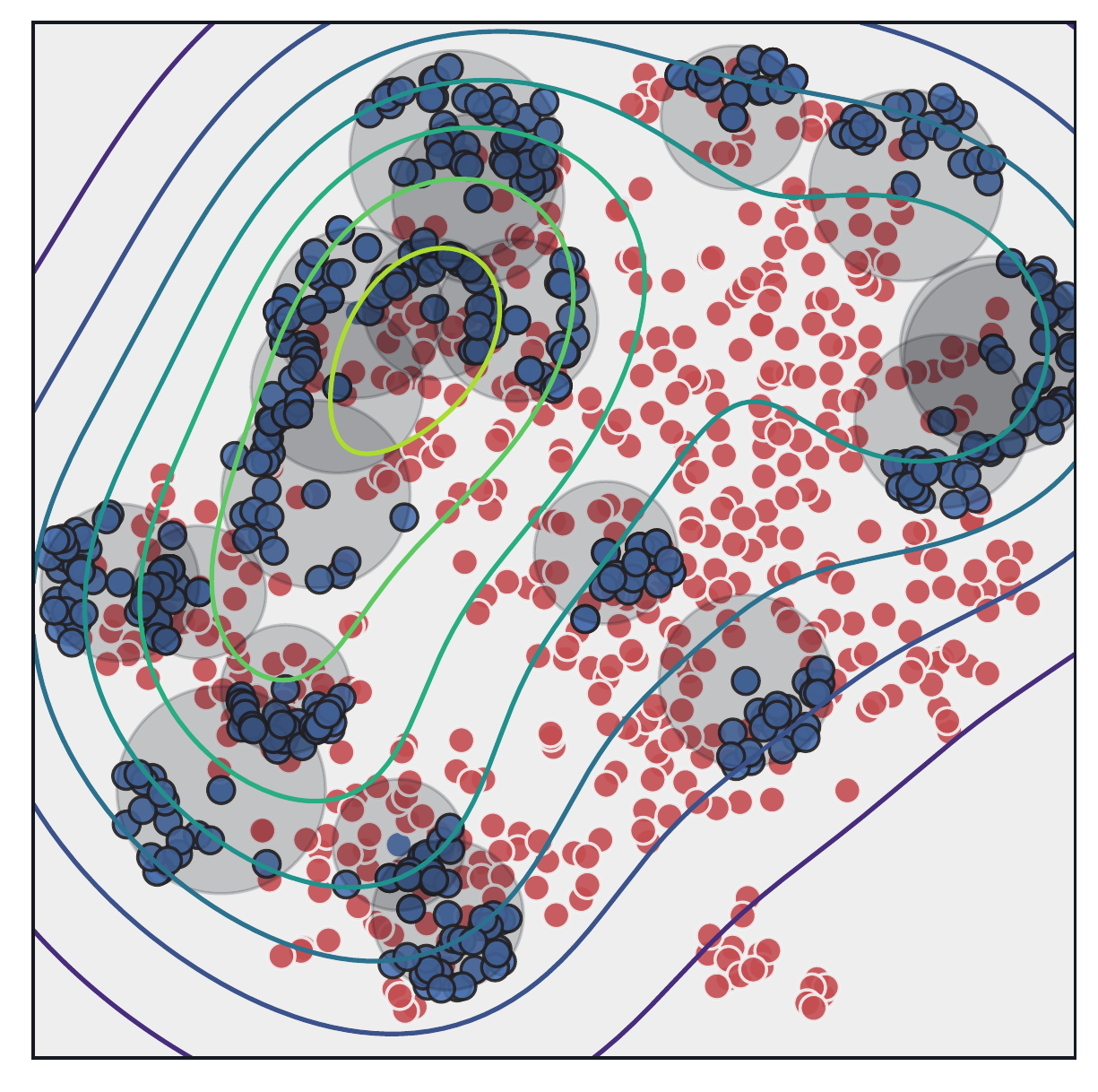}
  \caption{L}
\end{subfigure}
\caption{An example of the choice of sampling region on the distribution of generated minority observations. Baseline case, equivalent to sampling in all of the possible regions (LEH), was compared with sampling in the high (H), equal (E) and low (L) potential regions. Note that the distribution of generated observations aligns with the shape of the potential field.}
\label{fig:example-regions-global} 
\end{figure*}

\begin{figure*}[!htb]
\centering
\begin{subfigure}[b]{0.18\textwidth}
  \includegraphics[width=\textwidth]{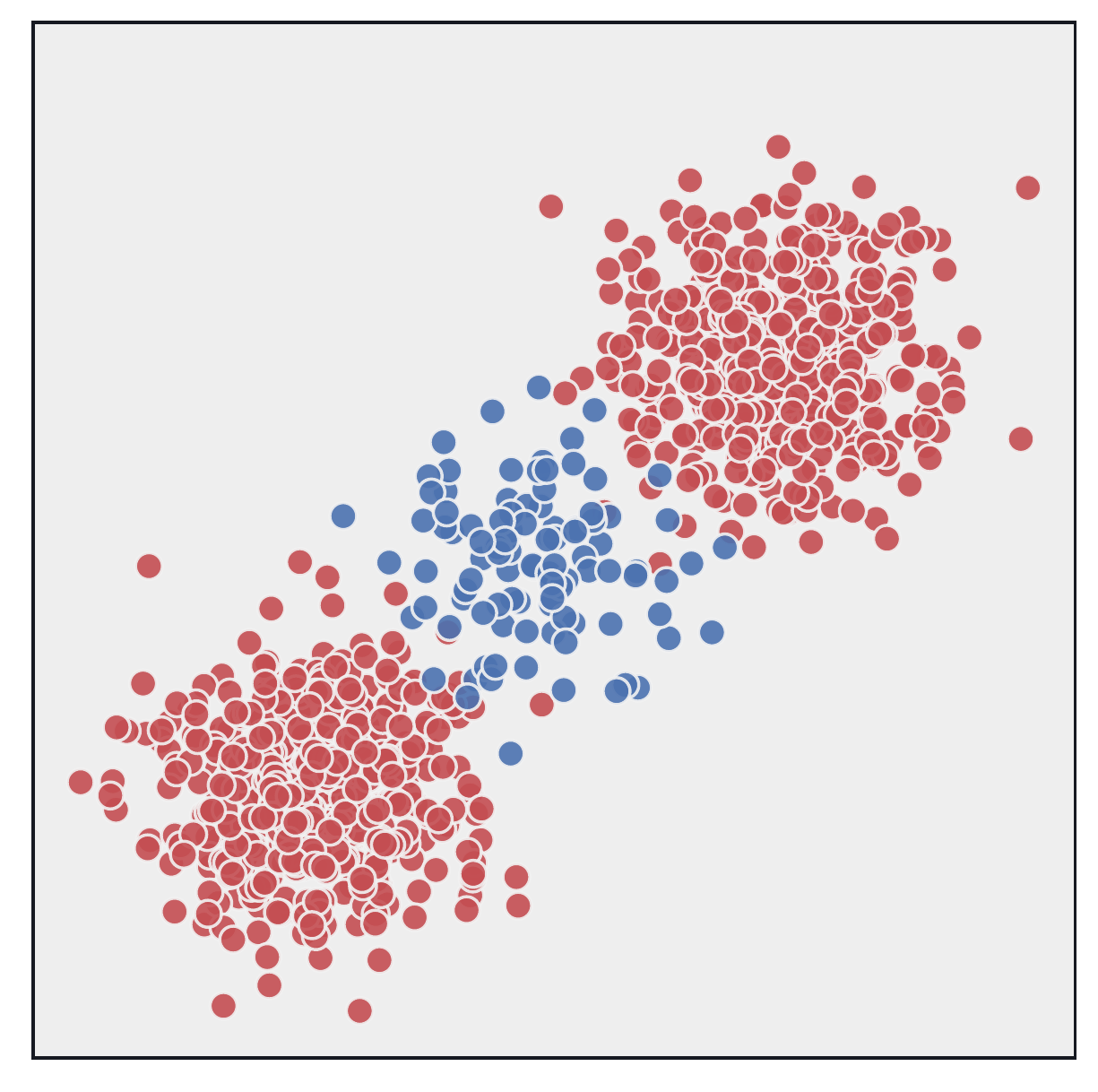}
  \caption{original data}
\end{subfigure}
~
\begin{subfigure}[b]{0.18\textwidth}
  \includegraphics[width=\textwidth]{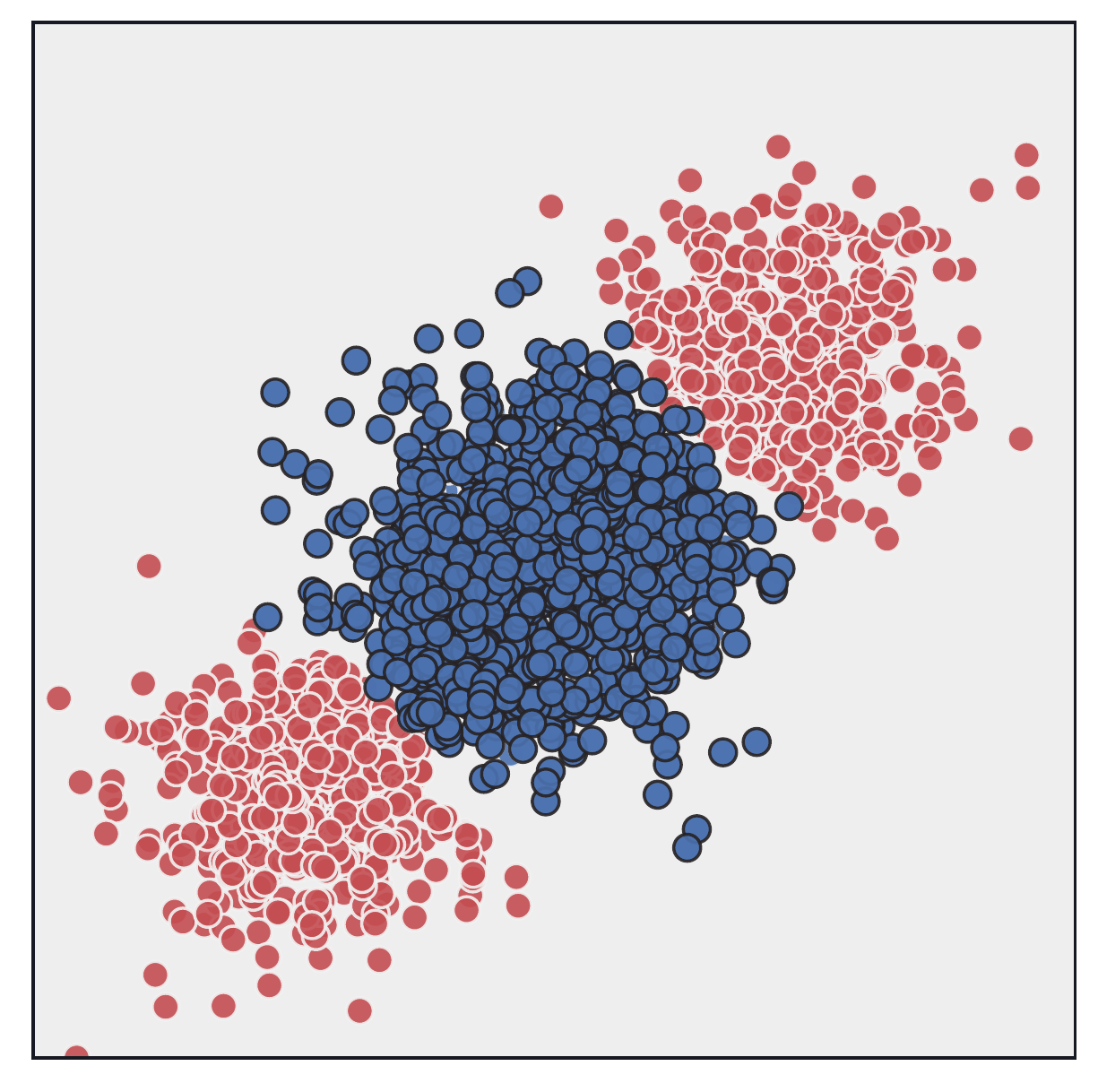}
  \caption{CCR}
\end{subfigure}
~
\begin{subfigure}[b]{0.18\textwidth}
  \includegraphics[width=\textwidth]{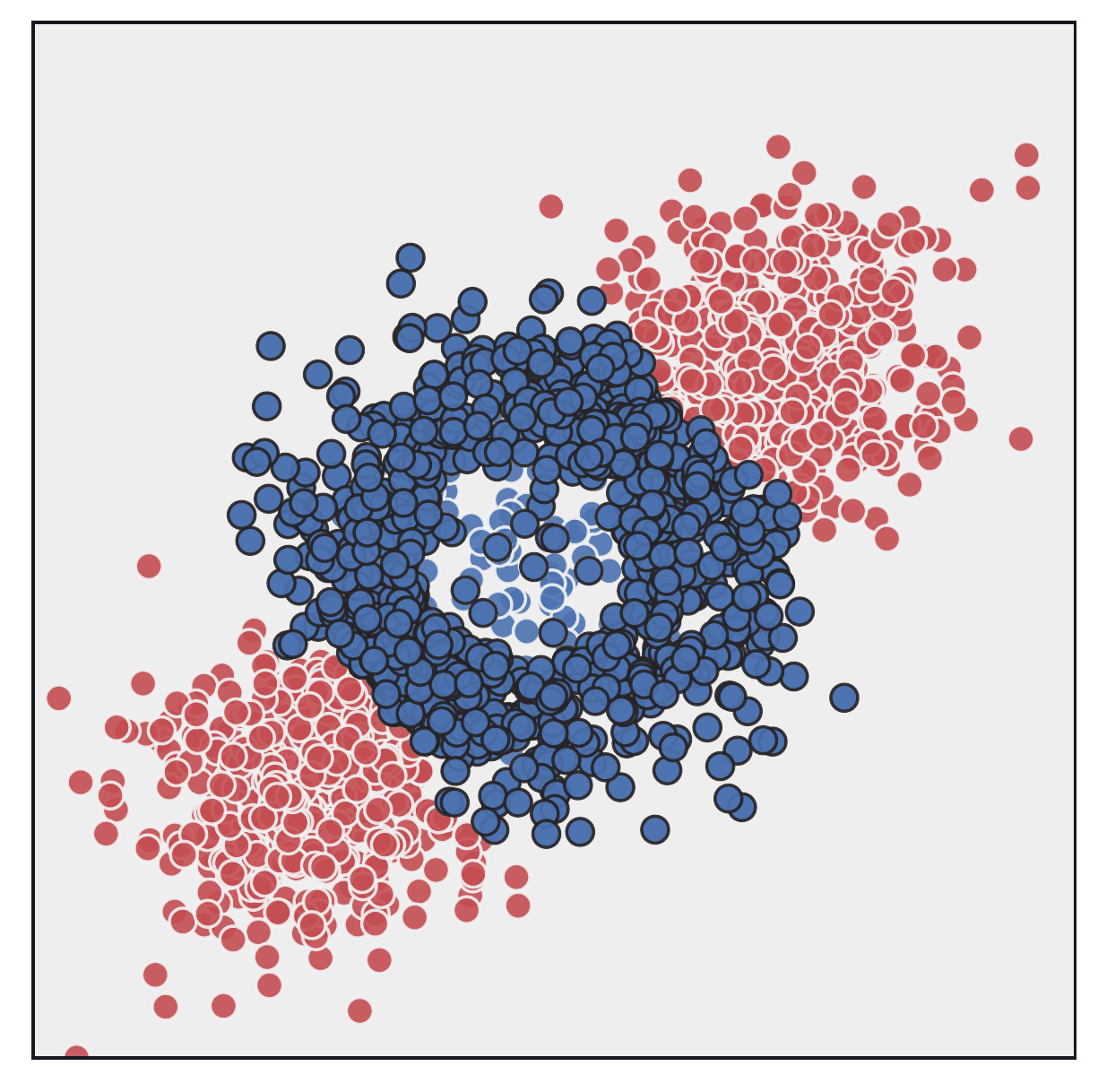}
  \caption{RB-CCR (L)}
\end{subfigure}
~
\begin{subfigure}[b]{0.18\textwidth}
  \includegraphics[width=\textwidth]{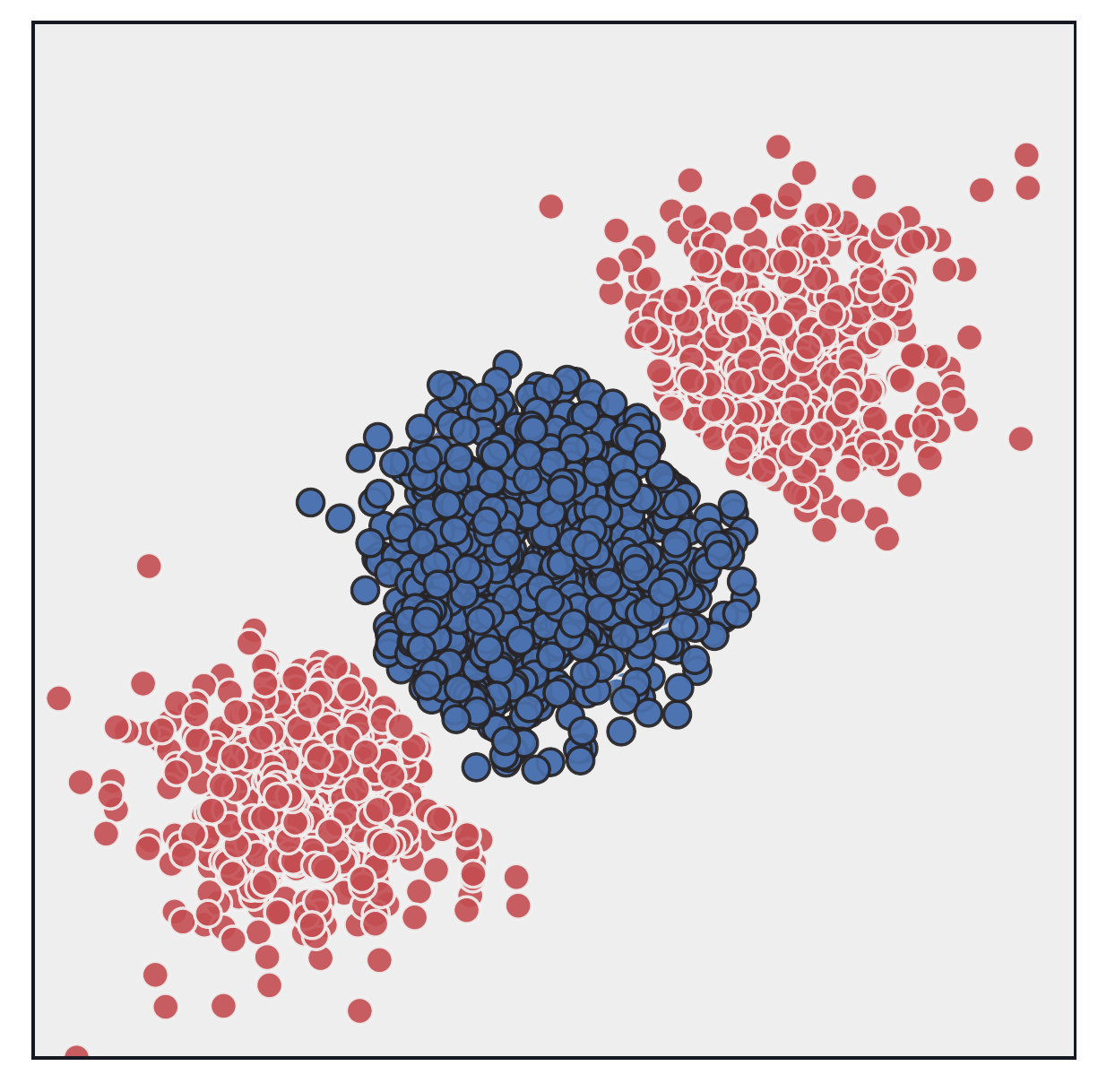}
  \caption{RB-CCR (E)}
\end{subfigure}
~
\begin{subfigure}[b]{0.18\textwidth}
  \includegraphics[width=\textwidth]{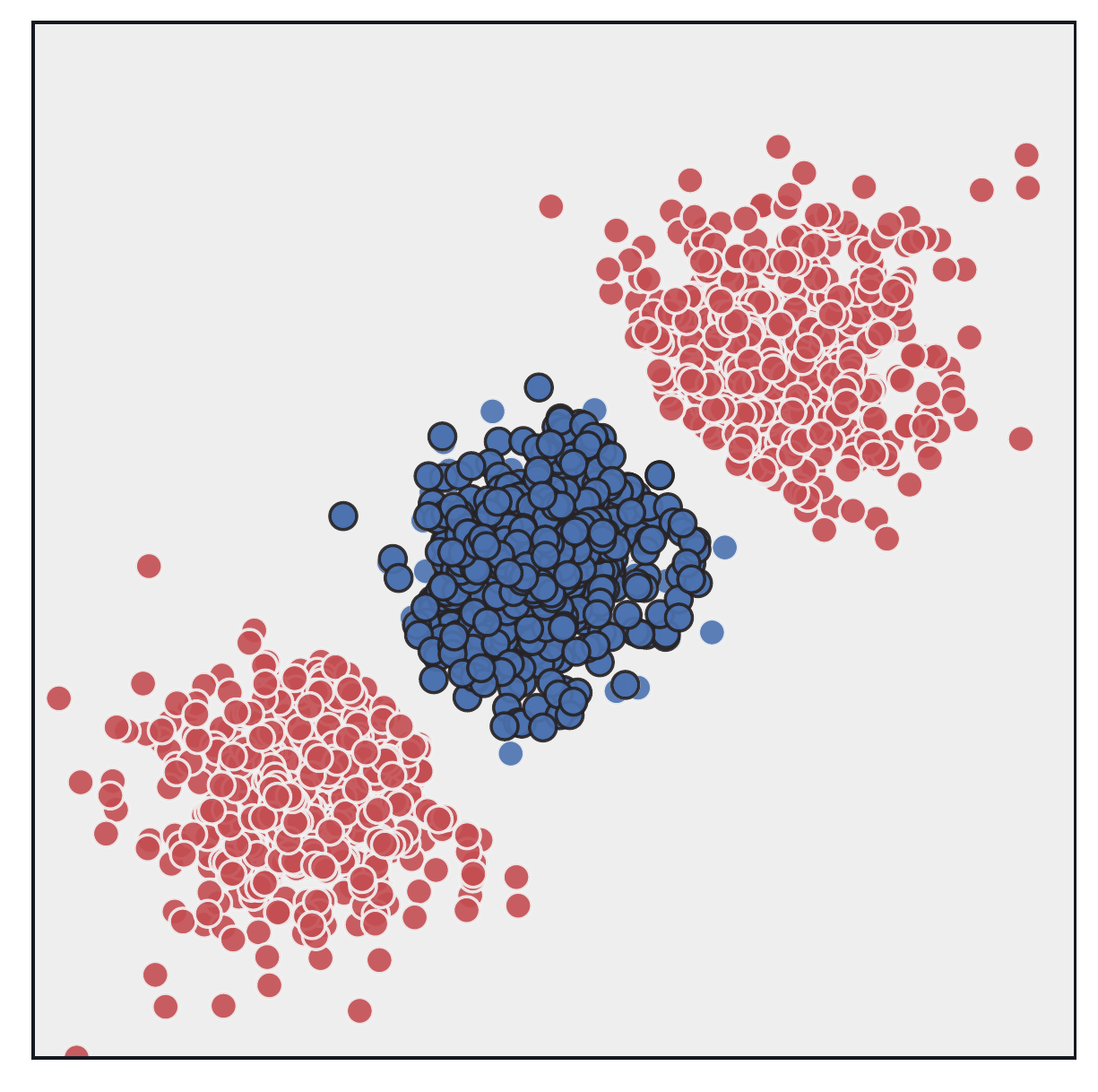}
  \caption{RB-CCR (H)}
\end{subfigure}
\caption{\rev{Comparison of CCR and RB-CCR with different sampling regions on a simplified dataset.}}
\label{fig:example-ccr-rb-ccr}
\end{figure*}

The behavior of the proposed algorithm changes depending on the choice of its three major hyperparameters: RBF spread $\gamma$, energy used for sphere expansion, and sampling region. The impact of $\gamma$ was illustrated in Figure~\ref{fig:example-potential}. As can be seen, $\gamma$ regulates the smoothness of the potential shape, with low values of $\gamma$ producing a less regular contour, conditioned mainly on the position of minority neighbors located in close proximity. On the contrary, higher $\gamma$ values produce a smoother, less prone to overfitting potential, with a smaller number of distinguishable clusters. Secondly, the value of energy affects the radius of the produced spheres, which controls the size of sampling regions and the range of translations, as illustrated in Figure~\ref{fig:example-sphere-radius}. It is worth noting that as the energy approaches zero, the algorithm degenerates to random oversampling. The choice of the energy is also highly dependent on the dimensionality of the data. It has to be scaled to the number of features a given dataset contains, with higher dimensional datasets requiring higher energy to achieve a similar sphere expansion. Finally, the choice of the sampling region determines how the generated samples align with the minority class potential. This is demonstrated in Figure~\ref{fig:example-regions-global}. Sampling in all of the available regions (LEH) is equivalent to the original CCR algorithm. This completely ignores the potential and uses whole spheres as a region of interest. Sampling in region E constrains samples to areas with class potential that is approximately equal class potential of real minority observation. Sampling in region H pushes the generated observations towards areas of the data space estimated to have a higher minority class potential. 
This can be interpreted as focusing the sampling process on generating samples that are safer, and better resemble the original minority observations. The opposite is true for sampling in the region L. \rev{This was further illustrated on a simplified dataset in Figure~\ref{fig:example-ccr-rb-ccr}.}

\begin{figure*}[!htb]
\centering
\begin{subfigure}[b]{0.23\textwidth}
  \includegraphics[width=\textwidth]{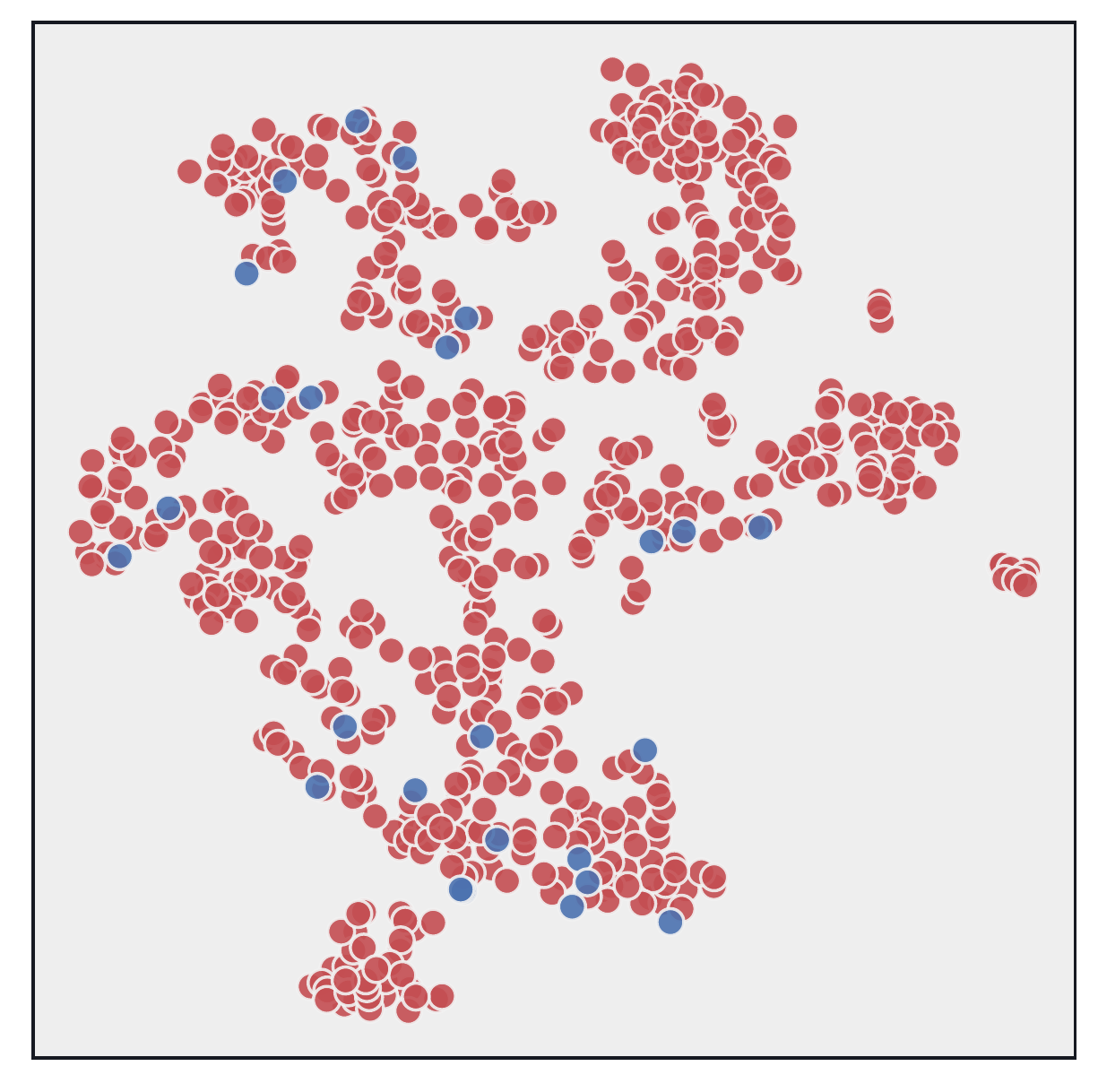}
  \caption{original data}
\end{subfigure}
~
\begin{subfigure}[b]{0.23\textwidth}
  \includegraphics[width=\textwidth]{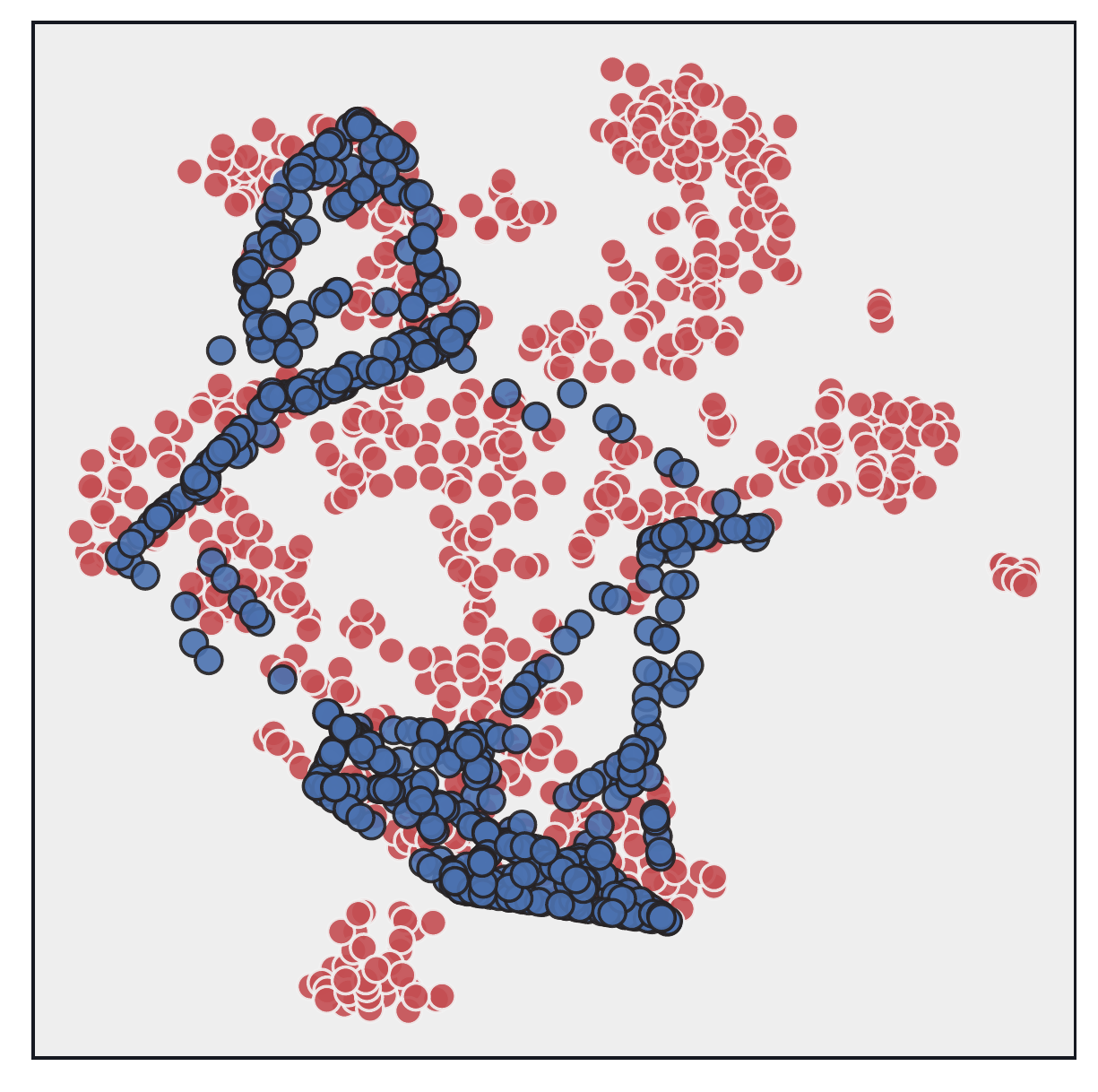}
  \caption{SMOTE}
\end{subfigure}
~
\begin{subfigure}[b]{0.23\textwidth}
  \includegraphics[width=\textwidth]{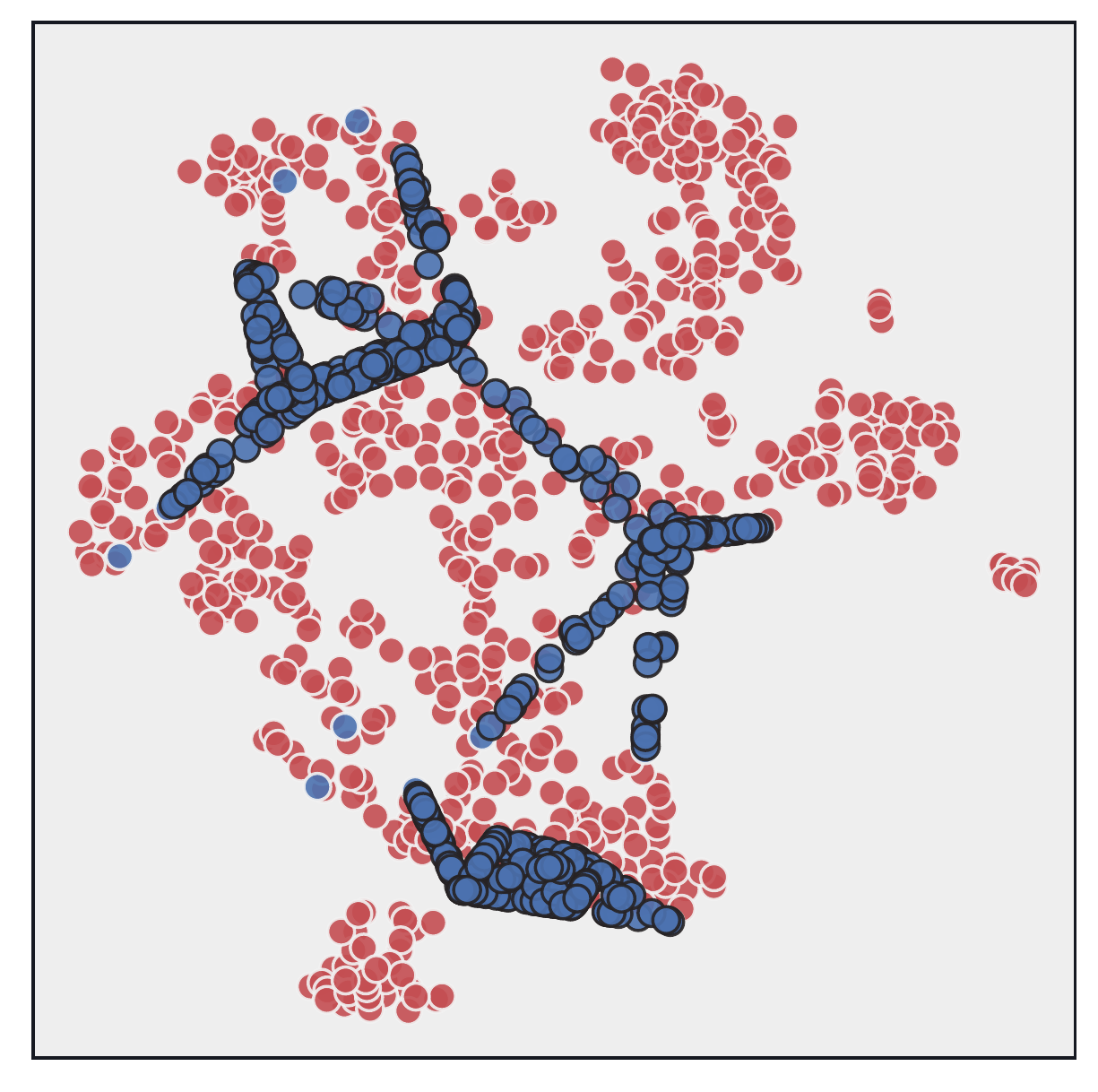}
  \caption{Bord}
\end{subfigure}

\begin{subfigure}[b]{0.23\textwidth}
  \includegraphics[width=\textwidth]{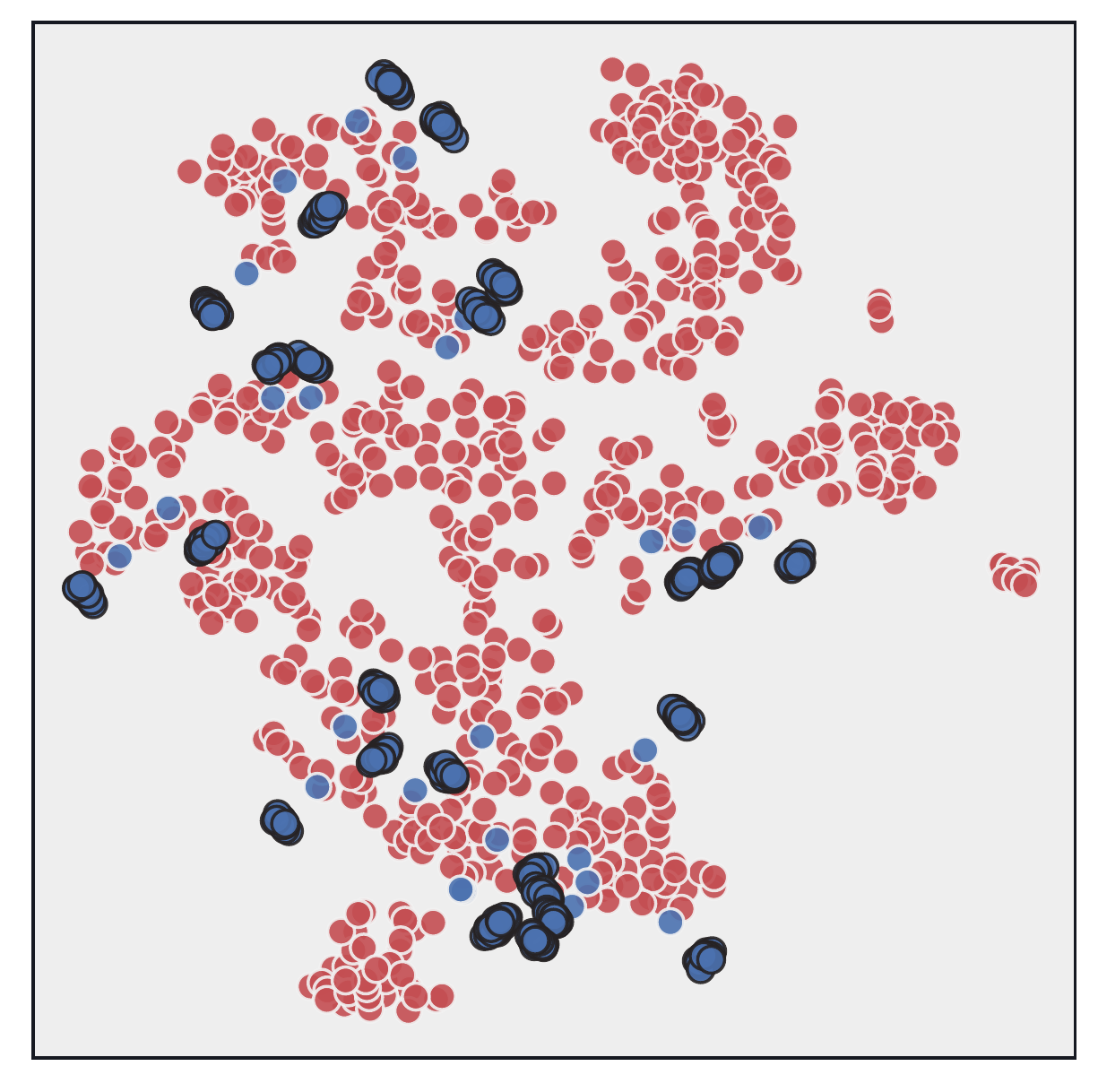}
  \caption{RBO}
\end{subfigure}
~
\begin{subfigure}[b]{0.23\textwidth}
  \includegraphics[width=\textwidth]{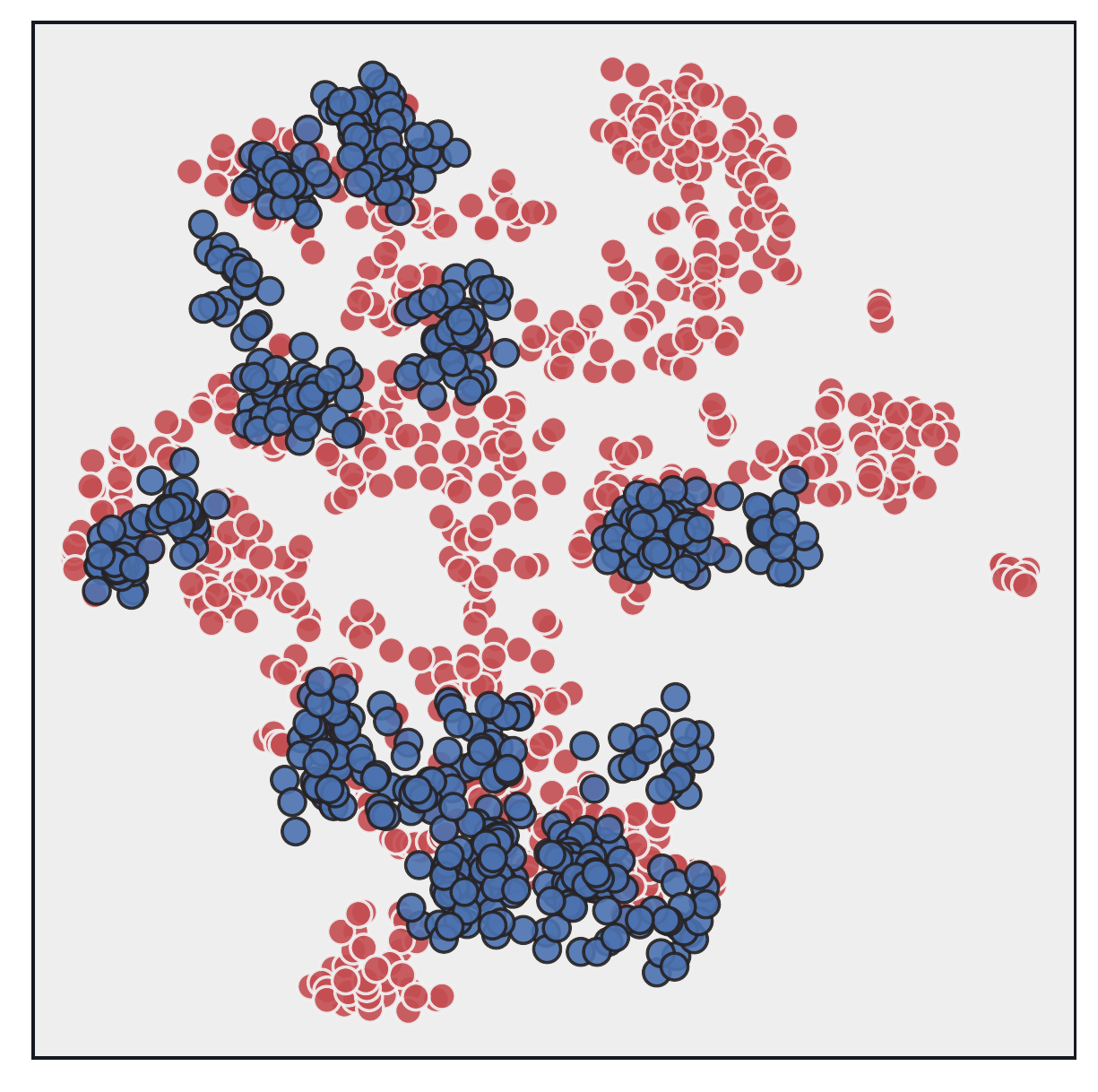}
  \caption{CCR}
\end{subfigure}
~
\begin{subfigure}[b]{0.23\textwidth}
  \includegraphics[width=\textwidth]{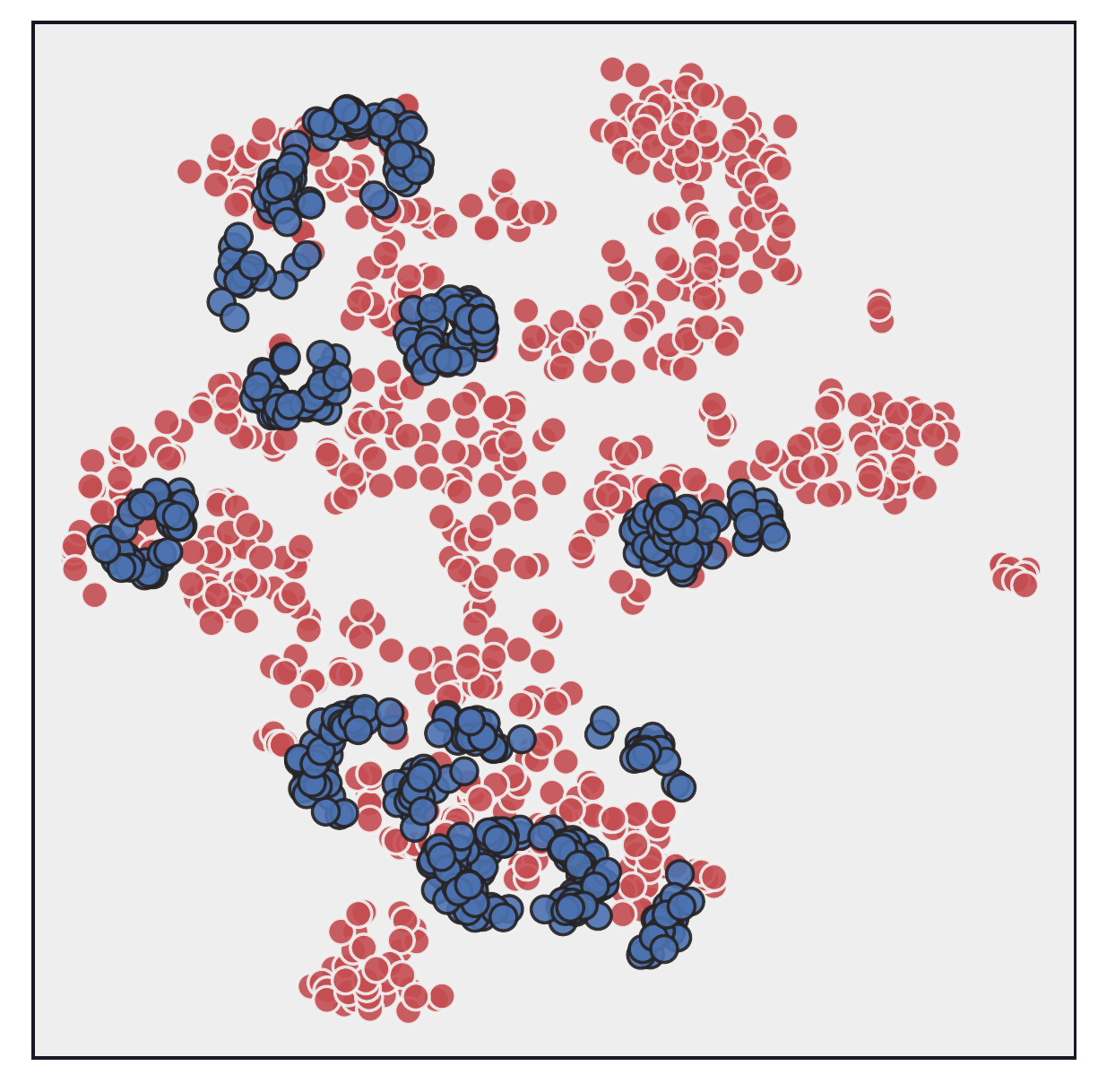}
  \caption{RB-CCR}
\end{subfigure}
\caption{A comparison of data distribution after oversampling with different algorithms on a highly imbalanced dataset with disjoint minority class distributions. \rev{SMOTE introduces a high degree of class overlap; Borderline-SMOTE (Bord) solves the problem only partially, still introducing some overlap, at the same time completely omitting to oversample around selected observations.} RBO does not produce artificial overlap, but at the same time, it is very conservative during sampling, in particular within originally overlapping regions. CCR and RB-CCR produce a distribution that leads to a higher bias towards the minority class, both due to synthesizing observations around all of the instances and the conducted translation of majority observations while minimizing class overlap. Compared to CCR, RB-CCR produces more constrained samples based on the underlying potential.}
\label{fig:example-method-comparison}
\end{figure*}

Finally, it is worth discussing how RB-CCR compares to the other oversampling algorithms. An illustration of differences between several popular methods was presented in Figure~\ref{fig:example-method-comparison}, with a highly imbalanced dataset characterized by a disjoint minority class distribution used as a benchmark. As can be seen, when compared to the SMOTE-based approaches, RB-CCR tends to introduce lower class overlap, which can occur for SMOTE when dealing with disjoint distributions, the presence of noise or outliers. RBO avoids sampling in the majority class regions. However, it produces very conservative and highly clustered samples. These can cause the classifier to overfit in a manner similar to random oversampling. RB-CCR avoids the risk of overfitting with larger regions of interest.
Moreover, the larger regions enable a greater reduction in the classifier's bias towards the majority class. The energy parameter facilitates the control of this behavior, with higher values of energy leading to less conservative sampling. Information provided by the class potential is used to fine-tune the shape of regions of interest within the sphere. It enables better control of the sampling.

\subsection{Computational complexity analysis}

Let us define the total number of observations by $n$, the number of majority and minority observations by, respectively, $n_{maj}$ and $n_{min}$, the number of features by $m$, and the number of candidate samples used in a single sampling step of Algorithm~\ref{algorithm:sampling} by $c$. As previously described in \cite{KOZIARSKI2020106223}, the original CCR algorithm can be divided into three steps: calculating the sphere radii, cleaning the majority observations inside the spheres, and synthesizing new observations, with each of the steps done iteratively for every minority observation. The same applies to the RB-CCR, for which only the complexity of the third step will differ from that of CCR.

\begin{itemize}
\item 
As described in \cite{KOZIARSKI2020106223}, the first step consists of a) calculating a distance vector, b) sorting said vector, and c) calculating the resulting radius. Combined, these operations have complexity equal to $\mathcal{O}((m + \log{n})n^2)$.
\item 
As described in \cite{KOZIARSKI2020106223}, the second step, cleaning the majority observations inside the spheres, has complexity equal to $\mathcal{O}(mn)$.
\item
Finally, the third step, synthesizing new observations, consists of a) calculating the proportion of samples generated for a given observation $g_i$, with the complexity equal to $\mathcal{O}(n_{min})$ \cite{KOZIARSKI2020106223}, and b) sampling the synthetic observations. In the case of the original CCR algorithm, as discussed in \cite{KOZIARSKI2020106223}, this sub-step consists of $n_{maj} - n_{min}$ operations of sampling a random observation inside the sphere, each with complexity equal to $\mathcal{O}(m)$, leading to a total complexity of the third step of CCR that can be simplified to $\mathcal{O}(mn)$. On the other hand, the sampling used by RB-CCR has a higher complexity due to the chosen guided strategy. In particular, when considering the procedure described in Algorithm~\ref{algorithm:sampling}, its complexity is dominated by the potential calculation for all of the candidate samples. Potential calculation, defined in Equation~\ref{eq:potential}, when computed with respect to the collection of minority class observations $\mathcal{X}_{min}$, consists of $n_{min}$ summations and $n_{min}$ RBF function computations, with the later having complexity equal to $\mathcal{O}(m)$. As a result, a single computation of the minority class potential has a complexity that can be simplified to $\mathcal{O}(mn)$. Whole sampling step, which requires $c$ potential function computations per minority observations, has therefore a total complexity equal to $\mathcal{O}(cmnn_{min})$, which can be simplified to $\mathcal{O}(cmn^2)$.
\end{itemize}
As can be seen, for the original CCR algorithm, the complexity is dominated by the first step and is equal to $\mathcal{O}((m + \log{n})n^2)$. On the other hand, in the case of RB-CCR, both the first and the third step influence the total complexity of the algorithm, which is equal to $\mathcal{O}((cm + \log{n})n^2)$.

\section{Experimental study} \label{sec:experiments}

To empirically evaluate the usefulness of the proposed RB-CCR algorithm, we conducted a series of experiment, the aim of which was to answer the following research questions:

\begin{itemize}
    \item[RQ1]  Is it possible to improve the original CCR algorithm's performance by focusing resampling in the specific regions?
    \item[RQ2] Are the trends displayed by the RB-CCR consistent across different classification algorithms and performance metrics? Is it possible to control the behavior of the algorithm by a proper choice of parameters?
    \item[RQ3] How does RB-CCR compare with state-of-the-art reference methods, and how does the choice of classification algorithm affect that comparison?
\end{itemize}

\subsection{Set-up}

\textbf{Datasets.} We based our experiments on 57 binary datasets taken from the KEEL repository \cite{alcala2011keel}, the details of which, namely their names, imbalance ratios (IR), number of contained samples and features, were presented in Table~\ref{table:datasets}. We employed a dataset selection procedure previously used in \cite{koziarski2020radial}, that is we excluded datasets for which AUC greater than 0.85 was achieved with a linear SVM without any resampling. Prior to the resampling and classification, all datasets were preprocessed: categorical features were converted to integers first, and afterward, all of the features were normalized to zero mean and unit variance.

\begin{table}
\caption{Summary of the characteristics of datasets used throughout the experimental study.}
\label{table:datasets}
\centering
\begin{tabular}{lrrrlrrr}
\toprule
Name & IR & Samples & Features & Name & IR & Samples & Features \\
\cmidrule(l){1-4} \cmidrule(l){5-8} 
glass1 & 1.82 & 214 & 9 & glass016vs5 & 19.44 & 184 & 9 \\
pima & 1.87 & 768 & 8 & yeast1458vs7 & 22.10 & 693 & 8 \\
glass0 & 2.06 & 214 & 9 & glass5 & 22.78 & 214 & 9 \\
yeast1 & 2.46 & 1484 & 8 & yeast2vs8 & 23.10 & 482 & 8 \\
haberman & 2.78 & 306 & 3 & flareF & 23.79 & 1066 & 11 \\
vehicle1 & 2.90 & 846 & 18 & yeast4 & 28.10 & 1484 & 8 \\
vehicle3 & 2.99 & 846 & 18 & winequalityred4 & 29.17 & 1599 & 11 \\
ecoli1 & 3.36 & 336 & 7 & poker9vs7 & 29.50 & 244 & 10 \\
ecoli2 & 5.46 & 336 & 7 & yeast1289vs7 & 30.57 & 947 & 8 \\
yeast3 & 8.10 & 1484 & 8 & winequalitywhite9vs4 & 32.60 & 168 & 11 \\
ecoli3 & 8.60 & 336 & 7 & yeast5 & 32.73 & 1484 & 8 \\
pageblocks0 & 8.79 & 5472 & 10 & winequalityred8vs6 & 35.44 & 656 & 11 \\
yeast2vs4 & 9.08 & 514 & 8 & ecoli0137vs26 & 39.14 & 281 & 7 \\
ecoli067vs35 & 9.09 & 222 & 7 & abalone17vs78910 & 39.31 & 2338 & 8 \\
yeast0359vs78 & 9.12 & 506 & 8 & abalone21vs8 & 40.50 & 581 & 8 \\
glass015vs2 & 9.12 & 172 & 9 & yeast6 & 41.40 & 1484 & 8 \\
yeast0256vs3789 & 9.14 & 1004 & 8 & winequalitywhite3vs7 & 44.00 & 900 & 11 \\
ecoli01vs235 & 9.17 & 244 & 7 & winequalityred8vs67 & 46.50 & 855 & 11 \\
ecoli0267vs35 & 9.18 & 224 & 7 & abalone19vs10111213 & 49.69 & 1622 & 8 \\
yeast05679vs4 & 9.35 & 528 & 8 & krvskzerovseight & 53.07 & 1460 & 6 \\
glass016vs2 & 10.29 & 192 & 9 & winequalitywhite39vs5 & 58.28 & 1482 & 11 \\
ecoli0147vs2356 & 10.59 & 336 & 7 & poker89vs6 & 58.40 & 1485 & 10 \\
glass0146vs2 & 11.06 & 205 & 9 & winequalityred3vs5 & 68.10 & 691 & 11 \\
glass2 & 11.59 & 214 & 9 & abalone20vs8910 & 72.69 & 1916 & 8 \\
cleveland0vs4 & 12.31 & 173 & 13 & poker89vs5 & 82.00 & 2075 & 10 \\
yeast1vs7 & 14.30 & 459 & 7 & poker8vs6 & 85.88 & 1477 & 10 \\
glass4 & 15.46 & 214 & 9 & abalone19 & 129.44 & 4174 & 8 \\
pageblocks13vs4 & 15.86 & 472 & 10 &  &  &  &  \\
abalone918 & 16.40 & 731 & 8 &  &  &  &  \\
zoo3 & 19.20 & 101 & 16 &  &  &  &  \\
\bottomrule
\end{tabular}
\end{table}

\textbf{Classification algorithms.} During the conducted experiments, we considered classification with a total of 9 different algorithms: CART decision tree, k-nearest neighbors classifier (KNN), support vector machine with linear (L-SVM), RBF (R-SVM) and polynomial (P-SVM) kernels, logistic regression (LR), Naive Bayes (NB), and multi-layer perceptron with ReLU (R-MLP) and linear (L-MLP) activation functions in the hidden layer. We considered a relatively high number of classification algorithms to examine how the choice of base learner affects the usefulness of RB-CCR. Implementations of all of the classification algorithms were taken from the scikit-learn library \cite{pedregosa2011scikit}, and their default parameters were used.

\textbf{Reference methods.} In addition to the original CCR algorithm, we compared the performance of RB-CCR with several over- and undersampling strategies, namely: SMOTE \cite{chawla2002smote}, Borderline-SMOTE (Bord) \cite{han2005borderline}, Neighborhood Cleaning Rule (NCL) \cite{laurikkala2001improving}, SMOTE combined with Tomek links (SMOTE+TL) \cite{tomek1976two} and Edited Nearest Neighbor rule (SMOTE+EN) \cite{wilson1972asymptotic}. The hyperparameters for each resampling method were tuned individually for each dataset. The SMOTE variants considered the values of $k$ neighborhood in \{1, 3, 5, 7, 9\}. In addition to $K$, the Bord method considered the values of $m$ neighborhood in \{5, 10, 15\}. For NCL, we considered the value $k$ of its neighborhood in \{1, 3, 5, 7\}. \rev{Finally, for all methods in which the resampling ratio was an inherent parameter, resampling was performed up to the point of achieving a balanced class distributions.} Implementation of all of the reference methods was taken from the imbalanced-learn library \cite{lemaitre2017imbalanced}.

\textbf{Performance metrics.} \rev{We utilize 6 performance metrics for classifier evaluation. This includes precision, recall, and specificity of the predictions, and the combined metrics AUC, F-measure, and G-mean. This set of metrics is standard in the imbalanced classification literature and provides a diverse perspective on model performance. Precision, recall, and specificity provide insight into the class specific errors that are to be expected from each algorithm. The combine metrics, AUC, F-measure, and G-mean, provide a more wholesome perspective on performance by taking into account the trade-off between the performance on majority and minority class. As mentioned in the related work, AUC and F-measure have previously been criticized in the context of imbalance learning. Nonetheless, we include them as they remain standard benchmarks in the literature and provide orthogonal perspective on performance.} 


\textbf{Evaluation procedure.} To ensure the stability of the results, we used $5\times2$ cross-validation \cite{alpaydin1999combined} during all of the experiments. Furthermore, during the parameter selection for resampling algorithms we used additional 3-fold cross-validation on the training partition of the data, with AUC used as the optimization criterion.

\textbf{Statistical analysis.} To assess the statistical significance of the results, we used two types of statistical tests. We used a one-sided Wilcoxon signed-rank test in a direct comparison between the original CCR algorithm and the proposed RB-CCR algorithm. Secondly, when simultaneously comparing multiple methods, we used the Friedman test combined with Shaffer's posthoc. In all cases, unless $p$-values were specified, the results were reported at the significance level $\alpha = 0.10$.

\textbf{Implementation and reproducibility.} To ensure the reproducibility of the results, we made publicly available the following: the implementation of the algorithm, code sufficient to reproduce all of the described experiments, statistical tests, and all of the figures presented in this paper, as well as the partitioning of the data into folds and raw results. All of the above can be accessed at\footnote{\url{https://github.com/michalkoziarski/RB-CCR}}.

\subsection{Evaluation of the choice of sampling region on the algorithms performance}
\label{sec:region-comp}

In the first stage of the conducted experimental analysis, we examined the suitability of sampling in specific regions. We compared the performance of four variants of RB-CCR algorithm, in which sampling was performed only in the low potential region (L), only in the approximately equal potential region (E), and only in the high potential region (H), as well as the variant in which sampling was performed in all of the regions (LEH), which is equivalent to the original CCR algorithm. In all of the cases, we selected the energy parameter from \{0.5, 1.0, 2.5, 5.0, ..., 100.0\}. Furthermore, except LEH sampling we also selected the value of $\gamma$ from \{0.5, 1.0, 2.5, 5.0, 10.0\}. We present a summary of regions achieving the highest average rank 
for every classifier and metric combination in Table~\ref{table:regions}. Detailed $p$-values for the conducted experiments can also be found in Appendix~\ref{apx:regions}.

\begin{table}
\caption{A summary of sampling strategies that achieved highest average rank for a given classifier and metric combination. Cases in which the best strategy achieved a statistically significantly better results than at least one of the other strategies were denoted with a + sign, and cases in which the best strategy achieved statistically significantly better results than sampling in all of the regions (LEH) were denoted with a ++ sign.}
\label{table:regions}
\centering
\begin{tabular}{lllllll}
\toprule
& Precision & Recall & Specificity & AUC & F-measure & G-mean \\
\midrule
CART & H \textsubscript{++} & L \textsubscript{+} & H \textsubscript{++} & L \textsubscript{+} & H & L\\
KNN & H \textsubscript{+} & L & H \textsubscript{++} & H & H & H\\
L-SVM & H \textsubscript{+} & E \textsubscript{+} & H \textsubscript{++} & E \textsubscript{+} & H \textsubscript{+} & E \textsubscript{+}\\
R-SVM & H \textsubscript{++} & L \textsubscript{+} & H \textsubscript{++} & L \textsubscript{+} & H & L \textsubscript{+}\\
P-SVM & LEH & H \textsubscript{+} & LEH & H & LEH & H \textsubscript{++}\\
LR & H \textsubscript{++} & E & H \textsubscript{++} & H \textsubscript{++} & H \textsubscript{++} & H \textsubscript{++}\\
NB & L \textsubscript{++} & E \textsubscript{+} & H \textsubscript{++} & L \textsubscript{+} & L \textsubscript{+} & L\\
R-MLP & H & L \textsubscript{+} & H \textsubscript{++} & L \textsubscript{+} & L & L\\
L-MLP & H \textsubscript{++} & LEH \textsubscript{+} & H \textsubscript{++} & E & H \textsubscript{++} & E\\
\bottomrule
\end{tabular}
\end{table}

Several observations can be made based on the presented results. First of all, the observed performance was consistent across the classification algorithms 
concerning the precision, recall, and specificity, at least when comparing sampling in H region with the remaining variants: in the case of precision sampling exclusively in the H region produced, on average, the best performance when combined with 7 out of 9 considered classifiers, with the remaining two being NB and P-SVM. Furthermore, in the case of specificity, this behavior was observed for 8 out of 9 classifiers, once again except P-SVM. Finally, the reverse was true in the case of recall, were sampling in the H region gave the worst average rank for 8 out of 9 considered classifiers. All of the trends mentioned above were also statistically significant in the majority of cases. This indicates that using the guided sampling approach has a non-random influence on the algorithm's performance and its bias towards the majority class, particularly when comparing sampling in the H region with the other variants, which is desirable behavior. Furthermore, from a general resampling perspective, this suggests that if the problem domain requires high precision or specificity, it is beneficial to focus sampling in the H region. On the other hand, if a high recall is required, sampling in L or E region is usually preferred.

However, the sampling region's impact on the combined metrics is less clear in the general case. Although the baseline variant of CCR that is LEH sampling, achieved the best average rank only in 1 out of 27 cases (for the combination of P-SVM and F-measure), there was usually either a complete lack of significance, meaning that there were no statistically significant differences between any of the sampling strategies, or partial significance, meaning that only some of the variants displayed statistically significant differences. Importantly, when comparing with the LEH sampling, there was a statistically significant improvement concerning all of the combined metrics for a single classifier, LR; and for a single metric for the combination of L-MLP and F-measure, as well as the combination of G-mean and P-SVM. In all of the above cases, the best performance strategy was sampling in the H region. Nevertheless, for the remaining combinations of classification algorithms and performance metrics there was no clearly dominant strategy, even when at least partial significance was observed. All of the above leads to the conclusion that while sampling in the specific regions has a non-random impact that is consistent across the classification algorithms with respect to direction (focusing sampling in the H region leading to a statistically significantly better precision and specificity, and worse recall), the trade-off between them, which can be observed using the combined metrics, varies depending on both the classifier and the dataset.

\subsection{Comparison of CCR and RB-CCR}

We have empirically demonstrated that no single sample region is optimal for all datasets, classification algorithms, and performance metrics. It is consistent with a current state of knowledge, particularly the "no free lunch" theorem, according to which the choice of sampling strategy strongly depends on the dataset characteristics. Instead, 
we considered the approach in which we treat the sampling region as a parameter of the algorithm and adjust it on a per-dataset and per-classifier basis. To this end, we conducted two comparisons. 

First of all, considered an idealized variant of RB-CCR. The region is giving the best performance, chosen only from \{L, E, \}, was selected individually for each dataset based on the test set results. Importantly, sampling in the LEH region was not included in the selection of available regions. This approach can be treated as an upper bound of performance that could be achieved by restricting sampling to a specific region. Once again, this variant of RB-CCR was compared with the original CCR algorithm, with the results presented in Table~\ref{table:rb-ccr-best-comparison}. As can be seen, by constraining sampling to a specific region, we were able to achieve improved performance for almost every considered dataset, regardless of the choice of classifier or performance metric.

\begin{table}
\caption{Comparison of the original CCR algorithm with an idealized variant of RB-CCR, for which the sampling region giving the best performance was chosen individually for each dataset. The number of datasets for which either CCR or RB-CCR achieved better average performance, as well as $p$-value, were presented.}
\label{table:rb-ccr-best-comparison}
\centering
\begin{tabular}{llllllllll}
\toprule
& \multicolumn{3}{l}{AUC} & \multicolumn{3}{l}{F-measure} & \multicolumn{3}{l}{G-mean} \\
\cmidrule(l){2-4} \cmidrule(l){5-7} \cmidrule(l){8-10}
Clf. & CCR & RB-CCR & $p$-value & CCR & RB-CCR & $p$-value & CCR & RB-CCR & $p$-value \\
\midrule
CART & 3 & 54 & \textbf{0.0000} & 0 & 57 & \textbf{0.0000} & 3 & 54 & \textbf{0.0000} \\
KNN & 3 & 54 & \textbf{0.0000} & 1 & 56 & \textbf{0.0000} & 3 & 54 & \textbf{0.0000} \\
L-SVM & 2 & 55 & \textbf{0.0000} & 0 & 57 & \textbf{0.0000} & 2 & 55 & \textbf{0.0000} \\
R-SVM & 2 & 55 & \textbf{0.0000} & 1 & 56 & \textbf{0.0000} & 2 & 55 & \textbf{0.0000} \\
P-SVM & 1 & 56 & \textbf{0.0000} & 3 & 54 & \textbf{0.0000} & 1 & 56 & \textbf{0.0000} \\
LR & 0 & 57 & \textbf{0.0000} & 2 & 55 & \textbf{0.0000} & 0 & 57 & \textbf{0.0000} \\
NB & 3 & 54 & \textbf{0.0000} & 0 & 57 & \textbf{0.0000} & 3 & 54 & \textbf{0.0000} \\
R-MLP & 1 & 56 & \textbf{0.0000} & 3 & 54 & \textbf{0.0000} & 1 & 56 & \textbf{0.0000} \\
L-MLP & 3 & 54 & \textbf{0.0000} & 3 & 54 & \textbf{0.0000} & 4 & 53 & \textbf{0.0000} \\
\bottomrule
\end{tabular}
\end{table}

\rev{Secondly, we conducted a comparison between the original CCR algorithm and RB-CCR with the sampling region chosen from \{L, E, H, LEH\} using cross-validation.} The results of this comparison were presented in Table~\ref{table:rb-ccr-cv-comparison}. As can be seen, when adjusting the sampling region individually for each dataset we were able to achieve a statistically significant improvement in performance for at least one of the combined metrics for 7 out of 9 classifiers. This improvement was observed more often in the case of G-mean and AUC, and only in two cases for F-measure, which can be explained by the fact that AUC was used as the optimization criterion during cross-validation, and AUC and G-mean tend to be more correlated than F-measure. \rev{We hypothesize that the flexibility in RB-CCR offered by class potential regions enables the samples to be generated in areas that have the greatest positive impact on the metric being optimized. The results presented in Table~\ref{table:regions}, where focusing on the high potential regions produces a significant improvement in precision and specificity, seem to support this hypothesis. Thus, using F-measure as an optimization criterion for models trained with RB-CCR would have the opposite effect as AUC (\textit{i.e.} it would produce better precision, specificity and F-measure, since these are related, at the expanse of recall, AUC and G-mean.) }

\begin{table}
\caption{Comparison of the original CCR algorithm with RB-CCR using cross-validation to select resampling regions. The number of datasets for which either CCR or RB-CCR achieved better average performance, as well as $p$-value, were presented.}
\label{table:rb-ccr-cv-comparison}
\centering
\begin{tabular}{llllllllll}
\toprule
& \multicolumn{3}{l}{AUC} & \multicolumn{3}{l}{F-measure} & \multicolumn{3}{l}{G-mean} \\
\cmidrule(l){2-4} \cmidrule(l){5-7} \cmidrule(l){8-10}
Clf. & CCR & RB-CCR & $p$-value & CCR & RB-CCR & $p$-value & CCR & RB-CCR & $p$-value \\
\midrule
CART & 22 & 35 & 0.2010 & 24 & 33 & 0.2664 & 21 & 36 & \textbf{0.0189} \\
KNN & 25 & 32 & \textbf{0.0547} & 18 & 39 & \textbf{0.0021} & 21 & 36 & \textbf{0.0168} \\
L-SVM & 25 & 32 & \textbf{0.0651} & 30 & 27 & 0.4478 & 27 & 30 & \textbf{0.0956} \\
R-SVM & 23 & 34 & 0.1672 & 23 & 34 & 0.1288 & 25 & 32 & 0.1857 \\
P-SVM & 21 & 36 & \textbf{0.0149} & 24 & 33 & 0.2986 & 17 & 40 & \textbf{0.0011} \\
LR & 26 & 31 & 0.1288 & 27 & 30 & 0.3829 & 25 & 32 & \textbf{0.0641} \\
NB & 23 & 34 & \textbf{0.0224} & 19 & 38 & \textbf{0.0006} & 25 & 32 & \textbf{0.0192} \\
R-MLP & 22 & 35 & 0.1391 & 24 & 33 & 0.1159 & 23 & 34 & 0.2193 \\
L-MLP & 23 & 34 & \textbf{0.0530} & 27 & 30 & 0.3859 & 23 & 34 & \textbf{0.0260} \\
\bottomrule
\end{tabular}
\end{table}

Results of both of the above experiments indicate that, in principle, constraining sampling to a specific region can yield a clear performance improvement compared to the baseline approach. \rev{Using cross-validation to choose the optimal region for every case is a suitable strategy for picking region}, resulting in a statistically significant performance improvement in most cases. Still, it falls short of the performance of the idealized variant. It indicates that either a better parameter selection strategy, more suited for the imbalanced datasets, or a specific heuristic for choosing the sampling region, could improve the proposed method's overall performance.

\subsection{Comparison of RB-CCR with the reference methods}

\begin{figure*}[!htb]
\centering
\includegraphics[width=\textwidth]{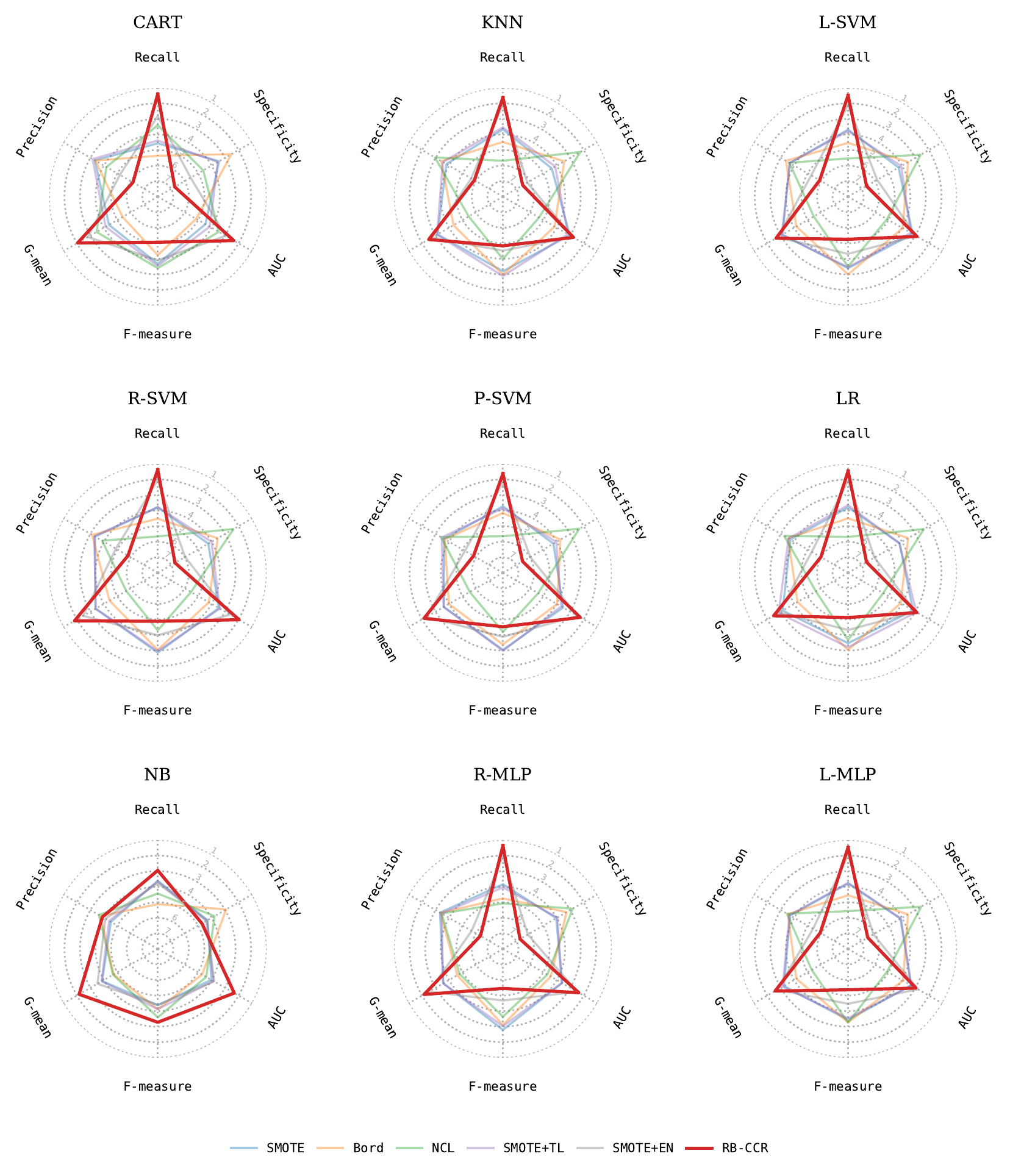}
\caption{\rev{A visualization of the average ranks achieved by the individual methods with respect to different performance metrics.}}
\label{fig:radar-plots}
\end{figure*}

We compared RB-CCR with several over- and undersampling reference methods in the next stage of the conducted experiments. We presented average ranks achieved by all of the methods, as well as the statistical significance of the comparison, in Table~\ref{table:results-final}. \rev{Furthermore, we presented a visualization of the average ranks achieved by the specific methods concerning different performance metrics in Figure~\ref{fig:radar-plots}.}  First of all, as can be seen, the general trend was that RB-CCR achieved the best recall at the expense of precision and specificity, which held true for all of the classification algorithms. As in the previous experiments, this had a varying impact on the combined metrics depending on their exact choice when F-measure was considered, which led to statistically significantly worse performance than the reference methods.
However, at the same time, it improved the performance concerning AUC and G-mean: RB-CCR achieved the highest average rank in 17 out of 18 cases, with the only exception of AUC observed for L-MLP classifier, for which it achieved the second-best rank. The results of this comparison were also statistically significant in the majority of cases: for all of the classifiers when compared to the baseline case with no resampling, Bord and NCL; for 5 out of 9 classifiers when compared to SMOTE and SMOTE+TL; and in a single case of NB when compared to SMOTE+EN, which was the second-best performer. \rev{The differences between the results measured using F-measure, AUC and G-mean can be attributed to the previously discussed bias of F-measure towards the majority class performance: since RB-CCR is heavily skewed towards the recall at the cost of precision, it is natural that using metric weighted more heavily towards precision produces worse performance.} \rev{Still, the observed results indicate high usefulness of the proposed RB-CCR algorithm when compared to the reference methods if a higher cost of misclassification of minority observations is assigned, as is the case with AUC and G-mean.}

\begin{table}
\small
\caption{A comparison of RB-CCR with the reference methods, with average ranks presented, and the methods for which RB-CCR achieved a statistically significantly better performance indicated with a + sign, and statistically significantly worse performance with a -- sign.}
\label{table:results-final}
\centering
\begin{tabularx}{\textwidth}{llYYYYYYY}
\toprule
Clf. & Metric & None & SMOTE & Bord & NCL & SMOTE+TL & SMOTE+EN & RB-CCR \\
\midrule
\multirow{6}{*}{CART} & Precision & \textbf{2.61} \textsubscript{--} & 3.33 \textsubscript{--} & 3.36 \textsubscript{--} & 4.19 \textsubscript{--} & 3.16 \textsubscript{--} & 5.19 & 6.16 \\
 & Recall & 6.04 \textsubscript{+} & 4.57 \textsubscript{+} & 5.37 \textsubscript{+} & 3.40 \textsubscript{+} & 4.39 \textsubscript{+} & 2.82 \textsubscript{+} & \textbf{1.40} \\
 & Specificity & \textbf{1.56} \textsubscript{--} & 3.45 \textsubscript{--} & 2.53 \textsubscript{--} & 4.62 \textsubscript{--} & 3.54 \textsubscript{--} & 5.56 \textsubscript{--} & 6.74 \\
 & AUC & 5.46 \textsubscript{+} & 4.46 \textsubscript{+} & 5.15 \textsubscript{+} & 3.47 \textsubscript{+} & 4.14 \textsubscript{+} & 2.96 & \textbf{2.35} \\
 & F-measure & 4.15 & 3.68 \textsubscript{--} & 4.24 & \textbf{3.40} \textsubscript{--} & 3.54 \textsubscript{--} & 3.91 \textsubscript{--} & 5.07 \\
 & G-mean & 5.78 \textsubscript{+} & 4.38 \textsubscript{+} & 5.39 \textsubscript{+} & 3.46 \textsubscript{+} & 4.13 \textsubscript{+} & 2.81 & \textbf{2.05} \\
\midrule
\multirow{6}{*}{KNN} & Precision & \textbf{2.77} \textsubscript{--} & 3.80 \textsubscript{--} & 3.42 \textsubscript{--} & 2.95 \textsubscript{--} & 3.54 \textsubscript{--} & 5.65 & 5.88 \\
 & Recall & 6.90 \textsubscript{+} & 3.67 \textsubscript{+} & 4.48 \textsubscript{+} & 5.69 \textsubscript{+} & 3.55 \textsubscript{+} & 2.07 & \textbf{1.63} \\
 & Specificity & \textbf{1.10} \textsubscript{--} & 4.37 \textsubscript{--} & 3.45 \textsubscript{--} & 2.25 \textsubscript{--} & 4.12 \textsubscript{--} & 6.19 & 6.53 \\
 & AUC & 6.74 \textsubscript{+} & 3.04 & 4.16 \textsubscript{+} & 5.32 \textsubscript{+} & 3.17 & 2.82 & \textbf{2.76} \\
 & F-measure & 5.57 & 3.20 \textsubscript{--} & 2.98 \textsubscript{--} & 3.99 & \textbf{2.89} \textsubscript{--} & 4.53 & 4.84 \\
 & G-mean & 6.73 \textsubscript{+} & 3.15 & 4.33 \textsubscript{+} & 5.46 \textsubscript{+} & 3.11 & 2.72 & \textbf{2.49} \\
\midrule
\multirow{6}{*}{L-SVM} & Precision & \textbf{2.48} \textsubscript{--} & 3.67 \textsubscript{--} & 3.39 \textsubscript{--} & 3.64 \textsubscript{--} & 3.67 \textsubscript{--} & 5.26 & 5.89 \\
 & Recall & 6.86 \textsubscript{+} & 3.69 \textsubscript{+} & 4.54 \textsubscript{+} & 5.54 \textsubscript{+} & 3.78 \textsubscript{+} & 2.11 & \textbf{1.48} \\
 & Specificity & \textbf{1.08} \textsubscript{--} & 4.25 \textsubscript{--} & 3.54 \textsubscript{--} & 2.61 \textsubscript{--} & 4.06 \textsubscript{--} & 5.84 & 6.63 \\
 & AUC & 6.55 \textsubscript{+} & 3.21 & 3.95 \textsubscript{+} & 5.11 \textsubscript{+} & 3.34 & 2.96 & \textbf{2.88} \\
 & F-measure & 5.06 & 3.40 \textsubscript{--} & \textbf{3.00} \textsubscript{--} & 3.50 \textsubscript{--} & 3.46 \textsubscript{--} & 4.32 & 5.26 \\
 & G-mean & 6.76 \textsubscript{+} & 3.12 & 4.16 \textsubscript{+} & 5.45 \textsubscript{+} & 3.09 & 2.77 & \textbf{2.65} \\
\midrule
\multirow{6}{*}{R-SVM} & Precision & 3.69 \textsubscript{--} & 3.34 \textsubscript{--} & \textbf{3.12} \textsubscript{--} & 3.87 \textsubscript{--} & 3.29 \textsubscript{--} & 4.89 & 5.79 \\
 & Recall & 6.87 \textsubscript{+} & 3.79 \textsubscript{+} & 4.52 \textsubscript{+} & 5.66 \textsubscript{+} & 3.77 \textsubscript{+} & 2.03 & \textbf{1.37} \\
 & Specificity & \textbf{1.13} \textsubscript{--} & 4.25 \textsubscript{--} & 3.55 \textsubscript{--} & 2.37 \textsubscript{--} & 4.01 \textsubscript{--} & 5.97 & 6.72 \\
 & AUC & 6.82 \textsubscript{+} & 3.46 \textsubscript{+} & 4.21 \textsubscript{+} & 5.53 \textsubscript{+} & 3.39 \textsubscript{+} & 2.63 & \textbf{1.96} \\
 & F-measure & 5.97 \textsubscript{+} & \textbf{2.87} \textsubscript{--} & 3.04 \textsubscript{--} & 4.29 & 2.97 \textsubscript{--} & 3.98 & 4.88 \\
 & G-mean & 6.87 \textsubscript{+} & 3.38 \textsubscript{+} & 4.42 \textsubscript{+} & 5.66 \textsubscript{+} & 3.39 \textsubscript{+} & 2.46 & \textbf{1.82} \\
\midrule
\multirow{6}{*}{P-SVM} & Precision & \textbf{2.80} \textsubscript{--} & 3.64 \textsubscript{--} & 3.68 \textsubscript{--} & 3.41 \textsubscript{--} & 3.46 \textsubscript{--} & 5.18 & 5.82 \\
 & Recall & 6.82 \textsubscript{+} & 3.72 \textsubscript{+} & 4.15 \textsubscript{+} & 5.65 \textsubscript{+} & 3.84 \textsubscript{+} & 2.19 & \textbf{1.62} \\
 & Specificity & \textbf{1.13} \textsubscript{--} & 4.24 \textsubscript{--} & 3.74 \textsubscript{--} & 2.33 \textsubscript{--} & 4.04 \textsubscript{--} & 5.97 & 6.54 \\
 & AUC & 6.71 \textsubscript{+} & 3.53 \textsubscript{+} & 3.98 \textsubscript{+} & 5.38 \textsubscript{+} & 3.67 \textsubscript{+} & 2.47 & \textbf{2.26} \\
 & F-measure & 5.97 \textsubscript{+} & 3.04 \textsubscript{--} & 3.39 \textsubscript{--} & 4.20 & \textbf{2.98} \textsubscript{--} & 3.88 & 4.53 \\
 & G-mean & 6.80 \textsubscript{+} & 3.61 \textsubscript{+} & 3.99 \textsubscript{+} & 5.54 \textsubscript{+} & 3.60 \textsubscript{+} & 2.33 & \textbf{2.14} \\
\midrule
\multirow{6}{*}{LR} & Precision & \textbf{2.63} \textsubscript{--} & 3.72 \textsubscript{--} & 3.54 \textsubscript{--} & 3.25 \textsubscript{--} & 3.58 \textsubscript{--} & 5.30 & 5.98 \\
 & Recall & 6.85 \textsubscript{+} & 3.79 \textsubscript{+} & 4.50 \textsubscript{+} & 5.69 \textsubscript{+} & 3.65 \textsubscript{+} & 2.07 & \textbf{1.45} \\
 & Specificity & \textbf{1.04} \textsubscript{--} & 4.18 \textsubscript{--} & 3.54 \textsubscript{--} & 2.37 \textsubscript{--} & 4.20 \textsubscript{--} & 6.04 & 6.63 \\
 & AUC & 6.53 \textsubscript{+} & 3.18 & 4.10 \textsubscript{+} & 5.39 \textsubscript{+} & 2.93 & 2.98 & \textbf{2.89} \\
 & F-measure & 5.16 & 3.47 \textsubscript{--} & \textbf{3.04} \textsubscript{--} & 3.68 \textsubscript{--} & 3.21 \textsubscript{--} & 4.32 & 5.12 \\
 & G-mean & 6.79 \textsubscript{+} & 3.14 & 4.25 \textsubscript{+} & 5.61 \textsubscript{+} & 2.84 & 2.89 & \textbf{2.47} \\
\midrule
\multirow{6}{*}{NB} & Precision & 3.68 & 4.52 & 3.77 & \textbf{3.63} & 4.39 & 4.11 & 3.89 \\
 & Recall & 4.44 \textsubscript{+} & 3.62 & 5.12 \textsubscript{+} & 4.44 \textsubscript{+} & 3.66 & 3.77 & \textbf{2.95} \\
 & Specificity & 3.55 \textsubscript{--} & 4.30 & \textbf{2.92} \textsubscript{--} & 3.78 & 4.39 & 4.34 & 4.72 \\
 & AUC & 4.88 \textsubscript{+} & 4.02 \textsubscript{+} & 4.71 \textsubscript{+} & 4.40 \textsubscript{+} & 3.86 \textsubscript{+} & 3.82 \textsubscript{+} & \textbf{2.32} \\
 & F-measure & 4.11 & 4.38 & 4.11 & 3.60 & 4.13 & 4.39 & \textbf{3.28} \\
 & G-mean & 5.22 \textsubscript{+} & 3.87 \textsubscript{+} & 4.72 \textsubscript{+} & 4.68 \textsubscript{+} & 3.83 \textsubscript{+} & 3.52 \textsubscript{+} & \textbf{2.16} \\
\midrule
\multirow{6}{*}{R-MLP} & Precision & \textbf{2.39} \textsubscript{--} & 3.33 \textsubscript{--} & 3.36 \textsubscript{--} & 3.45 \textsubscript{--} & 3.45 \textsubscript{--} & 5.68 & 6.33 \\
 & Recall & 6.90 \textsubscript{+} & 3.84 \textsubscript{+} & 4.75 \textsubscript{+} & 5.07 \textsubscript{+} & 4.04 \textsubscript{+} & 2.04 & \textbf{1.35} \\
 & Specificity & \textbf{1.04} \textsubscript{--} & 4.04 \textsubscript{--} & 3.27 \textsubscript{--} & 2.82 \textsubscript{--} & 3.96 \textsubscript{--} & 6.15 & 6.73 \\
 & AUC & 6.75 \textsubscript{+} & 3.62 \textsubscript{+} & 4.45 \textsubscript{+} & 4.70 \textsubscript{+} & 3.61 \textsubscript{+} & 2.49 & \textbf{2.37} \\
 & F-measure & 5.29 & \textbf{2.79} \textsubscript{--} & 3.11 \textsubscript{--} & 3.64 \textsubscript{--} & 3.03 \textsubscript{--} & 4.68 & 5.46 \\
 & G-mean & 6.83 \textsubscript{+} & 3.58 \textsubscript{+} & 4.61 \textsubscript{+} & 4.78 \textsubscript{+} & 3.55 \textsubscript{+} & 2.51 & \textbf{2.14} \\
\midrule
\multirow{6}{*}{L-MLP} & Precision & \textbf{2.33} \textsubscript{--} & 3.70 \textsubscript{--} & 3.53 \textsubscript{--} & 3.46 \textsubscript{--} & 3.65 \textsubscript{--} & 5.40 & 5.93 \\
 & Recall & 6.89 \textsubscript{+} & 3.79 \textsubscript{+} & 4.55 \textsubscript{+} & 5.57 \textsubscript{+} & 3.75 \textsubscript{+} & 1.97 & \textbf{1.47} \\
 & Specificity & \textbf{1.03} \textsubscript{--} & 4.11 \textsubscript{--} & 3.56 \textsubscript{--} & 2.56 \textsubscript{--} & 4.08 \textsubscript{--} & 6.12 & 6.54 \\
 & AUC & 6.60 \textsubscript{+} & 3.32 & 3.92 & 5.10 \textsubscript{+} & 3.30 & \textbf{2.81} & 2.96 \\
 & F-measure & 4.65 & 3.44 \textsubscript{--} & 3.28 \textsubscript{--} & \textbf{3.25} \textsubscript{--} & 3.54 \textsubscript{--} & 4.47 & 5.37 \\
 & G-mean & 6.84 \textsubscript{+} & 3.23 & 4.16 \textsubscript{+} & 5.30 \textsubscript{+} & 3.14 & 2.75 & \textbf{2.58} \\
\bottomrule
\end{tabularx}
\end{table}

Finally, in the last stage of the conducted experiments, we compared different combinations of classification and resampling algorithms to establish their relative usefulness. We presented the average ranks observed for different combined metrics in Tables~\ref{table:ranks-auc} through~\ref{table:ranks-g-mean}, separately for the individual metrics. As can be seen, when F-measure was considered, RB-CCR was outperformed by the reference methods, who achieved the best performance when combined with either R-MLP or R-SVM, which was also the case for RB-CCR. However, when AUC and G-mean were considered, the combination of algorithms that achieved the highest average rank was RB-CCR and L-MLP, for both of those metrics. Besides L-MLP, the top-performing classifiers were R-MLP, R-SVM, and LR, in that order, all achieving the best performance when combined with RB-CCR. Overall, presented rankings indicate the importance of improving performance due to the resampling method for any given classification algorithm. From that point of view, out of the statistically significant improvements presented previously in Table~\ref{table:results-final}, of most importance were those achieved for R-MLP and R-SVM, for which RB-CCR achieved a statistically significantly better performance than all of the resamplers except SMOTE+EN. On the other hand, it is worth noting that linear methods, that is L-MLP, LR and L-SVM, achieved relatively high performance, populating 3 out of 5 spots for highest performing classification algorithms, at the same time achieving less statistically significant improvement due to using RB-CCR when compared to the reference methods. This may suggest the importance of further work aimed particularly at improving the performance of RB-CCR for linear methods, which seem to be particularly predisposed to the classification of imbalanced datasets.

\begin{table}
\caption{Average ranks achieved by the specific combinations of classification and resampling algorithms, with \textbf{AUC} used as the performance metric.}
\label{table:ranks-auc}
\centering
\begin{tabular}{llllll}
\toprule
Method & Rank & Method & Rank & Method & Rank \\
\midrule
(RB-CCR, L-MLP) & 16.30 & (SMOTE+TL, KNN) & 24.46 & (NCL, KNN) & 37.93 \\
(SMOTE+TL, L-MLP) & 18.79 & (Bord, R-MLP) & 24.52 & (NCL, R-SVM) & 39.84 \\
(RB-CCR, R-MLP) & 18.80 & (Bord, L-SVM) & 24.59 & (SMOTE+TL, CART) & 40.09 \\
(SMOTE+EN, R-MLP) & 19.20 & (Bord, KNN) & 25.14 & (SMOTE+TL, NB) & 40.43 \\
(SMOTE+EN, L-MLP) & 19.79 & (SMOTE, LR) & 25.26 & (SMOTE+EN, NB) & 40.73 \\
(SMOTE, L-MLP) & 20.08 & (SMOTE, R-SVM) & 25.50 & (SMOTE, NB) & 41.17 \\
(Bord, L-MLP) & 20.75 & (SMOTE+TL, R-SVM) & 25.83 & (Bord, CART) & 41.18 \\
(RB-CCR, R-SVM) & 21.00 & (RB-CCR, CART) & 26.66 & (NCL, LR) & 41.59 \\
(RB-CCR, LR) & 21.36 & (Bord, LR) & 27.54 & (SMOTE, CART) & 41.82 \\
(SMOTE+EN, LR) & 21.98 & (RB-CCR, P-SVM) & 27.54 & (Bord, NB) & 42.33 \\
(SMOTE+TL, L-SVM) & 22.05 & (Bord, R-SVM) & 27.69 & (None, R-MLP) & 42.66 \\
(SMOTE+EN, KNN) & 22.81 & (SMOTE+EN, P-SVM) & 31.93 & (NCL, NB) & 43.02 \\
(SMOTE, L-SVM) & 22.99 & (NCL, R-MLP) & 31.99 & (None, NB) & 43.14 \\
(SMOTE+EN, L-SVM) & 23.05 & (RB-CCR, NB) & 34.52 & (None, CART) & 43.92 \\
(RB-CCR, L-SVM) & 23.32 & (SMOTE+EN, CART) & 35.17 & (NCL, P-SVM) & 46.54 \\
(SMOTE+TL, LR) & 23.55 & (SMOTE+TL, P-SVM) & 35.32 & (None, L-SVM) & 46.61 \\
(SMOTE, R-MLP) & 24.06 & (NCL, L-MLP) & 35.46 & (None, KNN) & 48.35 \\
(RB-CCR, KNN) & 24.11 & (NCL, L-SVM) & 35.54 & (None, L-MLP) & 49.09 \\
(SMOTE+TL, R-MLP) & 24.16 & (SMOTE, P-SVM) & 36.28 & (None, LR) & 50.47 \\
(SMOTE+EN, R-SVM) & 24.28 & (NCL, CART) & 36.96 & (None, R-SVM) & 51.25 \\
(SMOTE, KNN) & 24.33 & (Bord, P-SVM) & 37.43 & (None, P-SVM) & 51.75 \\
\bottomrule
\end{tabular}
\end{table}

\begin{table}
\caption{Average ranks achieved by the specific combinations of classification and resampling algorithms, with \textbf{F-measure} used as the performance metric.}
\label{table:ranks-f-measure}
\centering
\begin{tabular}{llllll}
\toprule
Method & Rank & Method & Rank & Method & Rank \\
\midrule
(Bord, R-MLP) & 15.98 & (SMOTE+TL, P-SVM) & 29.84 & (NCL, R-SVM) & 34.18 \\
(SMOTE+TL, R-MLP) & 16.25 & (NCL, CART) & 29.97 & (RB-CCR, KNN) & 34.60 \\
(SMOTE, R-MLP) & 16.87 & (SMOTE+EN, P-SVM) & 30.28 & (SMOTE, CART) & 34.94 \\
(SMOTE+EN, R-MLP) & 20.08 & (NCL, KNN) & 30.32 & (RB-CCR, L-SVM) & 35.37 \\
(Bord, R-SVM) & 21.23 & (SMOTE, P-SVM) & 30.39 & (RB-CCR, CART) & 36.17 \\
(SMOTE, R-SVM) & 21.39 & (RB-CCR, R-SVM) & 30.46 & (RB-CCR, LR) & 36.39 \\
(SMOTE+TL, R-SVM) & 21.41 & (NCL, L-SVM) & 30.65 & (None, CART) & 36.49 \\
(Bord, KNN) & 22.68 & (RB-CCR, R-MLP) & 30.81 & (NCL, P-SVM) & 38.03 \\
(NCL, R-MLP) & 22.84 & (SMOTE, LR) & 31.04 & (None, L-MLP) & 38.72 \\
(Bord, L-SVM) & 24.08 & (Bord, P-SVM) & 31.09 & (None, KNN) & 40.38 \\
(Bord, L-MLP) & 24.19 & (RB-CCR, P-SVM) & 31.35 & (None, L-SVM) & 40.45 \\
(SMOTE+EN, R-SVM) & 25.18 & (None, R-MLP) & 31.37 & (RB-CCR, NB) & 42.10 \\
(SMOTE+TL, L-MLP) & 26.24 & (SMOTE+EN, L-MLP) & 31.89 & (NCL, NB) & 43.42 \\
(SMOTE+TL, KNN) & 27.02 & (SMOTE+EN, KNN) & 32.25 & (None, LR) & 43.77 \\
(SMOTE, KNN) & 27.10 & (SMOTE+EN, LR) & 32.38 & (None, P-SVM) & 43.99 \\
(SMOTE+TL, L-SVM) & 27.40 & (RB-CCR, L-MLP) & 32.39 & (None, NB) & 44.05 \\
(SMOTE, L-SVM) & 27.52 & (SMOTE+EN, L-SVM) & 33.07 & (SMOTE+TL, NB) & 44.80 \\
(SMOTE, L-MLP) & 27.66 & (SMOTE+TL, CART) & 33.25 & (SMOTE+EN, NB) & 45.11 \\
(Bord, LR) & 28.31 & (SMOTE+EN, CART) & 33.33 & (Bord, NB) & 45.24 \\
(SMOTE+TL, LR) & 29.09 & (Bord, CART) & 33.44 & (SMOTE, NB) & 45.90 \\
(NCL, L-MLP) & 29.28 & (NCL, LR) & 34.11 & (None, R-SVM) & 46.46 \\
\bottomrule
\end{tabular}
\end{table}

\begin{table}
\caption{Average ranks achieved by the specific combinations of classification and resampling algorithms, with \textbf{G-mean} used as the performance metric.}
\label{table:ranks-g-mean}
\centering
\begin{tabular}{llllll}
\toprule
Method & Rank & Method & Rank & Method & Rank \\
\midrule
(RB-CCR, L-MLP) & 14.68 & (SMOTE+TL, KNN) & 24.19 & (SMOTE+TL, NB) & 39.21 \\
(SMOTE+TL, L-MLP) & 17.57 & (SMOTE+TL, R-MLP) & 24.79 & (SMOTE+EN, NB) & 39.42 \\
(RB-CCR, R-MLP) & 18.41 & (SMOTE, R-MLP) & 24.81 & (NCL, KNN) & 39.72 \\
(SMOTE+EN, L-MLP) & 18.97 & (SMOTE, R-SVM) & 25.08 & (SMOTE, NB) & 39.87 \\
(SMOTE, L-MLP) & 19.21 & (RB-CCR, CART) & 25.14 & (SMOTE+TL, CART) & 40.89 \\
(SMOTE+EN, R-MLP) & 19.30 & (Bord, KNN) & 25.32 & (Bord, NB) & 41.23 \\
(RB-CCR, R-SVM) & 19.50 & (Bord, R-MLP) & 25.55 & (NCL, R-SVM) & 41.97 \\
(RB-CCR, LR) & 19.84 & (SMOTE+TL, R-SVM) & 25.57 & (Bord, CART) & 42.04 \\
(Bord, L-MLP) & 20.25 & (Bord, LR) & 26.69 & (SMOTE, CART) & 42.08 \\
(SMOTE+EN, LR) & 20.34 & (RB-CCR, P-SVM) & 27.06 & (NCL, LR) & 42.50 \\
(SMOTE+TL, L-SVM) & 20.75 & (Bord, R-SVM) & 28.39 & (NCL, NB) & 43.79 \\
(RB-CCR, L-SVM) & 20.76 & (RB-CCR, NB) & 30.68 & (None, CART) & 44.56 \\
(SMOTE, L-SVM) & 21.32 & (SMOTE+EN, P-SVM) & 32.00 & (None, NB) & 44.57 \\
(SMOTE+EN, L-SVM) & 21.42 & (NCL, R-MLP) & 33.48 & (None, R-MLP) & 45.43 \\
(SMOTE+TL, LR) & 21.96 & (SMOTE+EN, CART) & 35.16 & (NCL, P-SVM) & 47.93 \\
(SMOTE+EN, KNN) & 22.39 & (SMOTE+TL, P-SVM) & 35.94 & (None, L-SVM) & 48.61 \\
(RB-CCR, KNN) & 22.47 & (SMOTE, P-SVM) & 36.75 & (None, KNN) & 50.18 \\
(SMOTE+EN, R-SVM) & 23.58 & (NCL, L-MLP) & 37.03 & (None, L-MLP) & 51.12 \\
(SMOTE, LR) & 23.59 & (NCL, L-SVM) & 37.14 & (None, LR) & 52.29 \\
(SMOTE, KNN) & 24.04 & (NCL, CART) & 37.60 & (None, P-SVM) & 53.54 \\
(Bord, L-SVM) & 24.10 & (Bord, P-SVM) & 38.59 & (None, R-SVM) & 53.61 \\
\bottomrule
\end{tabular}
\end{table}

\subsection{Lessons learned}

Based on the described results of the conducted experiments, we will now attempt to answer the research questions raised at the beginning of this section.

\textit{RQ1: Is it possible to improve the original CCR algorithm's performance by focusing resampling in the specific regions?}

We demonstrated that using RB-CCR leads to a statistically significantly better performance than CCR for most considered classification algorithms when the sampling region is determined using cross-validation. However, selecting the sampling region individually for each dataset and treating it as another hyperparameter was crucial in achieving that performance improvement in most cases. Finally, we also demonstrated that in almost every case sampling in a specific region leads to a better performance than unguided sampling within the whole sphere, indicating that choosing the optimal sampling region remains a major challenge that cross-validation solves only partially.

\textit{RQ2: Are the trends displayed by the RB-CCR consistent across different classification algorithms and performance metrics? Is it possible to control the behavior of the algorithm by a proper choice of parameters? }

The behavior of RB-CCR was consistent concerning precision, specificity and recall, with sampling solely within the H region improving precision and specificity at the expense of recall, and sampling within either L or E region having the opposite effect. As a result, it is possible to control the algorithm's bias towards the specific classes by properly choosing the sampling region. However, the performance concerning AUC, F-measure, and G-mean was less consistent, indicating that the choice of sampling region yielding the optimal trade-off between precision and recall is both dataset- and classifier-specific.

\textit{RQ3: How does RB-CCR compare with state-of-the-art reference methods, and how does the choice of classification algorithm affect that comparison?}

RB-CCR, on average, outperforms all of the considered reference methods concerning recall, AUC, and G-mean, and underperforms concerning precision, specificity and F-measure, with statistically significant differences between the majority of methods. It indicates that RB-CCR is a suitable choice whenever the performance of the minority class is the main consideration, which is usually the case in the imbalanced data classification task. Finally, a more significant improvement in performance due to using RB-CCR was observed for non-linear classification algorithms. Compared with the fact that linear methods, in general, achieved a favorable performance on the considered imbalanced datasets, this might indicate the need for further work focused specifically on improving the results for this type of classifiers.

\section{Conclusions} \label{sec:conclusion}

In this work, we proposed the Radial-Based Combined Cleaning and Resampling algorithm (RB-CCR). We hypothesized that the refining resampling procedure employed by CCR could garner additional performance gains. RB-CCR uses the concept of class potential to divide the dataspace around each minority instance into \emph{sampling regions} characterized by high, equal, or low class potential. Resampling is then restricted to the sub-regions with the specified characteristics, determined by cross-validation or user specification. Our results show that this is superior in the precision-recall trade-off to uniformly resampling around the minority class instances.

Our empirical assessment utilized 57 benchmark binary datasets, 9 classification algorithms and 5 state-of-the-art sampling techniques. The results measured as over 5-times 2-fold cross-validation show that sampling the high potential region with RB-CCR generally produces significantly better precision and specificity, with less impact on recall than CCR. Thus, RB-CCR achieves a better balance in the precision-recall trade-off. Moreover, on average RB-CCR outperforms the considered reference methods concerning recall, AUC and G-mean.

Future work may focus on designing a better region selection method than cross-validation, including a strategy for picking regions individually for each observation, which could not be done using cross-validation. Another potential direction is adjusting the RB-CCR algorithm to linear classifiers, which generally achieve good performance but are least affected by resampler choice and likely require a more drastic shift in the synthetic observation distribution to display a significant change classifier behavior.

\bibliographystyle{unsrt}  
\bibliography{references}

\appendix

\newpage

\section{Detailed $p$-values observed during sampling region comparison}
\label{apx:regions}


\begin{table}[!htb]
\small
\caption{A comparison of sampling in a specific regions, with \textbf{precision} used as the performance metric.}
\label{table:regions-Precision}
\centering
\scalebox{0.82}{
\subfloat[CART]{
\begin{tabularx}{0.32\textwidth}{lYYYY}
\toprule
& L & E & H & LEH \\
\midrule
Rank & 2.58 & 2.79 & \textbf{1.91} & 2.72 \\
\midrule
L & - & 1.000 & 0.018 & 1.000 \\
E & 1.000 & - & 0.002 & 1.000 \\
H & 0.018 & 0.002 & - & 0.003 \\
LEH & 1.000 & 1.000 & 0.003 & - \\
\bottomrule
\end{tabularx}
}
\subfloat[KNN]{
\begin{tabularx}{0.32\textwidth}{lYYYY}
\toprule
& L & E & H & LEH \\
\midrule
Rank & 2.79 & 2.61 & \textbf{2.16} & 2.44 \\
\midrule
L & - & 0.936 & 0.054 & 0.440 \\
E & 0.936 & - & 0.178 & 0.936 \\
H & 0.054 & 0.178 & - & 0.737 \\
LEH & 0.440 & 0.936 & 0.737 & - \\
\bottomrule
\end{tabularx}
}
\subfloat[L-SVM]{
\begin{tabularx}{0.32\textwidth}{lYYYY}
\toprule
& L & E & H & LEH \\
\midrule
Rank & 2.89 & 2.51 & \textbf{2.05} & 2.54 \\
\midrule
L & - & 0.331 & 0.003 & 0.331 \\
E & 0.331 & - & 0.178 & 0.885 \\
H & 0.003 & 0.178 & - & 0.127 \\
LEH & 0.331 & 0.885 & 0.127 & - \\
\bottomrule
\end{tabularx}
}
}

\scalebox{0.82}{
\subfloat[R-SVM]{
\begin{tabularx}{0.32\textwidth}{lYYYY}
\toprule
& L & E & H & LEH \\
\midrule
Rank & 2.70 & 2.53 & \textbf{2.02} & 2.75 \\
\midrule
L & - & 1.000 & 0.014 & 1.000 \\
E & 1.000 & - & 0.106 & 1.000 \\
H & 0.014 & 0.106 & - & 0.014 \\
LEH & 1.000 & 1.000 & 0.014 & - \\
\bottomrule
\end{tabularx}
}
\subfloat[P-SVM]{
\begin{tabularx}{0.32\textwidth}{lYYYY}
\toprule
& L & E & H & LEH \\
\midrule
Rank & 2.37 & 2.74 & 2.53 & \textbf{2.37} \\
\midrule
L & - & 0.766 & 1.000 & 1.000 \\
E & 0.766 & - & 1.000 & 0.766 \\
H & 1.000 & 1.000 & - & 1.000 \\
LEH & 1.000 & 0.766 & 1.000 & - \\
\bottomrule
\end{tabularx}
}
\subfloat[LR]{
\begin{tabularx}{0.32\textwidth}{lYYYY}
\toprule
& L & E & H & LEH \\
\midrule
Rank & 3.11 & 2.72 & \textbf{1.67} & 2.51 \\
\midrule
L & - & 0.221 & 0.000 & 0.041 \\
E & 0.221 & - & 0.000 & 0.384 \\
H & 0.000 & 0.000 & - & 0.001 \\
LEH & 0.041 & 0.384 & 0.001 & - \\
\bottomrule
\end{tabularx}
}
}

\scalebox{0.82}{
\subfloat[NB]{
\begin{tabularx}{0.32\textwidth}{lYYYY}
\toprule
& L & E & H & LEH \\
\midrule
Rank & \textbf{1.98} & 3.04 & 2.46 & 2.53 \\
\midrule
L & - & 0.000 & 0.106 & 0.074 \\
E & 0.000 & - & 0.050 & 0.106 \\
H & 0.106 & 0.050 & - & 0.772 \\
LEH & 0.074 & 0.106 & 0.772 & - \\
\bottomrule
\end{tabularx}
}
\subfloat[R-MLP]{
\begin{tabularx}{0.32\textwidth}{lYYYY}
\toprule
& L & E & H & LEH \\
\midrule
Rank & 2.44 & 2.54 & \textbf{2.28} & 2.74 \\
\midrule
L & - & 1.000 & 1.000 & 0.652 \\
E & 1.000 & - & 0.830 & 1.000 \\
H & 1.000 & 0.830 & - & 0.356 \\
LEH & 0.652 & 1.000 & 0.356 & - \\
\bottomrule
\end{tabularx}
}
\subfloat[L-MLP]{
\begin{tabularx}{0.32\textwidth}{lYYYY}
\toprule
& L & E & H & LEH \\
\midrule
Rank & 3.09 & 2.50 & \textbf{1.85} & 2.56 \\
\midrule
L & - & 0.045 & 0.000 & 0.059 \\
E & 0.045 & - & 0.022 & 0.800 \\
H & 0.000 & 0.022 & - & 0.010 \\
LEH & 0.059 & 0.800 & 0.010 & - \\
\bottomrule
\end{tabularx}
}
}
\end{table}

\begin{table}[!htb]
\small
\caption{A comparison of sampling in a specific regions, with \textbf{recall} used as the performance metric.}
\label{table:regions-Recall}
\centering
\scalebox{0.82}{
\subfloat[CART]{
\begin{tabularx}{0.32\textwidth}{lYYYY}
\toprule
& L & E & H & LEH \\
\midrule
Rank & \textbf{2.06} & 2.65 & 3.12 & 2.17 \\
\midrule
L & - & 0.045 & 0.000 & 0.663 \\
E & 0.045 & - & 0.138 & 0.138 \\
H & 0.000 & 0.138 & - & 0.000 \\
LEH & 0.663 & 0.138 & 0.000 & - \\
\bottomrule
\end{tabularx}
}
\subfloat[KNN]{
\begin{tabularx}{0.32\textwidth}{lYYYY}
\toprule
& L & E & H & LEH \\
\midrule
Rank & \textbf{2.35} & 2.50 & 2.73 & 2.42 \\
\midrule
L & - & 1.000 & 0.713 & 1.000 \\
E & 1.000 & - & 1.000 & 1.000 \\
H & 0.713 & 1.000 & - & 0.713 \\
LEH & 1.000 & 1.000 & 0.713 & - \\
\bottomrule
\end{tabularx}
}
\subfloat[L-SVM]{
\begin{tabularx}{0.32\textwidth}{lYYYY}
\toprule
& L & E & H & LEH \\
\midrule
Rank & 2.51 & \textbf{2.17} & 3.03 & 2.30 \\
\midrule
L & - & 0.471 & 0.097 & 0.768 \\
E & 0.471 & - & 0.002 & 0.768 \\
H & 0.097 & 0.002 & - & 0.008 \\
LEH & 0.768 & 0.768 & 0.008 & - \\
\bottomrule
\end{tabularx}
}
}

\scalebox{0.82}{
\subfloat[R-SVM]{
\begin{tabularx}{0.32\textwidth}{lYYYY}
\toprule
& L & E & H & LEH \\
\midrule
Rank & \textbf{1.86} & 2.61 & 3.63 & 1.90 \\
\midrule
L & - & 0.006 & 0.000 & 0.856 \\
E & 0.006 & - & 0.000 & 0.007 \\
H & 0.000 & 0.000 & - & 0.000 \\
LEH & 0.856 & 0.007 & 0.000 & - \\
\bottomrule
\end{tabularx}
}
\subfloat[P-SVM]{
\begin{tabularx}{0.32\textwidth}{lYYYY}
\toprule
& L & E & H & LEH \\
\midrule
Rank & 2.90 & 2.30 & \textbf{2.18} & 2.61 \\
\midrule
L & - & 0.037 & 0.018 & 0.575 \\
E & 0.037 & - & 0.637 & 0.575 \\
H & 0.018 & 0.637 & - & 0.226 \\
LEH & 0.575 & 0.575 & 0.226 & - \\
\bottomrule
\end{tabularx}
}
\subfloat[LR]{
\begin{tabularx}{0.32\textwidth}{lYYYY}
\toprule
& L & E & H & LEH \\
\midrule
Rank & 2.51 & \textbf{2.32} & 2.65 & 2.52 \\
\midrule
L & - & 1.000 & 1.000 & 1.000 \\
E & 1.000 & - & 1.000 & 1.000 \\
H & 1.000 & 1.000 & - & 1.000 \\
LEH & 1.000 & 1.000 & 1.000 & - \\
\bottomrule
\end{tabularx}
}
}

\scalebox{0.82}{
\subfloat[NB]{
\begin{tabularx}{0.32\textwidth}{lYYYY}
\toprule
& L & E & H & LEH \\
\midrule
Rank & 2.52 & \textbf{2.14} & 3.11 & 2.24 \\
\midrule
L & - & 0.356 & 0.045 & 0.491 \\
E & 0.356 & - & 0.000 & 0.690 \\
H & 0.045 & 0.000 & - & 0.001 \\
LEH & 0.491 & 0.690 & 0.001 & - \\
\bottomrule
\end{tabularx}
}
\subfloat[R-MLP]{
\begin{tabularx}{0.32\textwidth}{lYYYY}
\toprule
& L & E & H & LEH \\
\midrule
Rank & \textbf{2.11} & 2.40 & 3.12 & 2.36 \\
\midrule
L & - & 0.694 & 0.000 & 0.694 \\
E & 0.694 & - & 0.009 & 0.856 \\
H & 0.000 & 0.009 & - & 0.005 \\
LEH & 0.694 & 0.856 & 0.005 & - \\
\bottomrule
\end{tabularx}
}
\subfloat[L-MLP]{
\begin{tabularx}{0.32\textwidth}{lYYYY}
\toprule
& L & E & H & LEH \\
\midrule
Rank & 2.46 & 2.32 & 2.93 & \textbf{2.29} \\
\midrule
L & - & 1.000 & 0.150 & 1.000 \\
E & 1.000 & - & 0.049 & 1.000 \\
H & 0.150 & 0.049 & - & 0.049 \\
LEH & 1.000 & 1.000 & 0.049 & - \\
\bottomrule
\end{tabularx}
}
}
\end{table}

\begin{table}[!htb]
\small
\caption{A comparison of sampling in a specific regions, with \textbf{specificity} used as the performance metric.}
\label{table:regions-Specificity}
\centering
\scalebox{0.82}{
\subfloat[CART]{
\begin{tabularx}{0.32\textwidth}{lYYYY}
\toprule
& L & E & H & LEH \\
\midrule
Rank & 2.64 & 2.52 & \textbf{2.00} & 2.84 \\
\midrule
L & - & 0.808 & 0.024 & 0.808 \\
E & 0.808 & - & 0.097 & 0.539 \\
H & 0.024 & 0.097 & - & 0.003 \\
LEH & 0.808 & 0.539 & 0.003 & - \\
\bottomrule
\end{tabularx}
}
\subfloat[KNN]{
\begin{tabularx}{0.32\textwidth}{lYYYY}
\toprule
& L & E & H & LEH \\
\midrule
Rank & 2.70 & 2.66 & \textbf{1.99} & 2.65 \\
\midrule
L & - & 1.000 & 0.020 & 1.000 \\
E & 1.000 & - & 0.020 & 1.000 \\
H & 0.020 & 0.020 & - & 0.020 \\
LEH & 1.000 & 1.000 & 0.020 & - \\
\bottomrule
\end{tabularx}
}
\subfloat[L-SVM]{
\begin{tabularx}{0.32\textwidth}{lYYYY}
\toprule
& L & E & H & LEH \\
\midrule
Rank & 2.82 & 2.65 & \textbf{1.96} & 2.57 \\
\midrule
L & - & 0.981 & 0.003 & 0.929 \\
E & 0.981 & - & 0.014 & 0.981 \\
H & 0.003 & 0.014 & - & 0.037 \\
LEH & 0.929 & 0.981 & 0.037 & - \\
\bottomrule
\end{tabularx}
}
}

\scalebox{0.82}{
\subfloat[R-SVM]{
\begin{tabularx}{0.32\textwidth}{lYYYY}
\toprule
& L & E & H & LEH \\
\midrule
Rank & 2.77 & 2.56 & \textbf{1.75} & 2.92 \\
\midrule
L & - & 0.768 & 0.000 & 0.768 \\
E & 0.768 & - & 0.002 & 0.411 \\
H & 0.000 & 0.002 & - & 0.000 \\
LEH & 0.768 & 0.411 & 0.000 & - \\
\bottomrule
\end{tabularx}
}
\subfloat[P-SVM]{
\begin{tabularx}{0.32\textwidth}{lYYYY}
\toprule
& L & E & H & LEH \\
\midrule
Rank & 2.30 & 2.72 & 2.77 & \textbf{2.21} \\
\midrule
L & - & 0.245 & 0.150 & 1.000 \\
E & 0.245 & - & 1.000 & 0.122 \\
H & 0.150 & 1.000 & - & 0.122 \\
LEH & 1.000 & 0.122 & 0.122 & - \\
\bottomrule
\end{tabularx}
}
\subfloat[LR]{
\begin{tabularx}{0.32\textwidth}{lYYYY}
\toprule
& L & E & H & LEH \\
\midrule
Rank & 2.89 & 2.75 & \textbf{1.80} & 2.56 \\
\midrule
L & - & 0.850 & 0.000 & 0.539 \\
E & 0.850 & - & 0.000 & 0.850 \\
H & 0.000 & 0.000 & - & 0.005 \\
LEH & 0.539 & 0.850 & 0.005 & - \\
\bottomrule
\end{tabularx}
}
}

\scalebox{0.82}{
\subfloat[NB]{
\begin{tabularx}{0.32\textwidth}{lYYYY}
\toprule
& L & E & H & LEH \\
\midrule
Rank & 2.18 & 3.18 & \textbf{2.02} & 2.62 \\
\midrule
L & - & 0.000 & 0.491 & 0.139 \\
E & 0.000 & - & 0.000 & 0.067 \\
H & 0.491 & 0.000 & - & 0.037 \\
LEH & 0.139 & 0.067 & 0.037 & - \\
\bottomrule
\end{tabularx}
}
\subfloat[R-MLP]{
\begin{tabularx}{0.32\textwidth}{lYYYY}
\toprule
& L & E & H & LEH \\
\midrule
Rank & 2.48 & 2.48 & \textbf{2.16} & 2.88 \\
\midrule
L & - & 1.000 & 0.539 & 0.308 \\
E & 1.000 & - & 0.539 & 0.308 \\
H & 0.539 & 0.539 & - & 0.018 \\
LEH & 0.308 & 0.308 & 0.018 & - \\
\bottomrule
\end{tabularx}
}
\subfloat[L-MLP]{
\begin{tabularx}{0.32\textwidth}{lYYYY}
\toprule
& L & E & H & LEH \\
\midrule
Rank & 2.99 & 2.52 & \textbf{1.80} & 2.69 \\
\midrule
L & - & 0.150 & 0.000 & 0.435 \\
E & 0.150 & - & 0.009 & 0.468 \\
H & 0.000 & 0.009 & - & 0.001 \\
LEH & 0.435 & 0.468 & 0.001 & - \\
\bottomrule
\end{tabularx}
}
}
\end{table}

\begin{table}[!htb]
\small
\caption{A comparison of sampling in a specific regions, with \textbf{AUC} used as the performance metric.}
\label{table:regions-AUC}
\centering
\scalebox{0.82}{
\subfloat[CART]{
\begin{tabularx}{0.32\textwidth}{lYYYY}
\toprule
& L & E & H & LEH \\
\midrule
Rank & \textbf{2.12} & 2.77 & 2.75 & 2.35 \\
\midrule
L & - & 0.044 & 0.044 & 0.691 \\
E & 0.044 & - & 0.942 & 0.245 \\
H & 0.044 & 0.942 & - & 0.286 \\
LEH & 0.691 & 0.245 & 0.286 & - \\
\bottomrule
\end{tabularx}
}
\subfloat[KNN]{
\begin{tabularx}{0.32\textwidth}{lYYYY}
\toprule
& L & E & H & LEH \\
\midrule
Rank & 2.51 & 2.71 & \textbf{2.22} & 2.56 \\
\midrule
L & - & 1.000 & 0.694 & 1.000 \\
E & 1.000 & - & 0.253 & 1.000 \\
H & 0.694 & 0.253 & - & 0.471 \\
LEH & 1.000 & 1.000 & 0.471 & - \\
\bottomrule
\end{tabularx}
}
\subfloat[L-SVM]{
\begin{tabularx}{0.32\textwidth}{lYYYY}
\toprule
& L & E & H & LEH \\
\midrule
Rank & 2.79 & \textbf{2.14} & 2.56 & 2.51 \\
\midrule
L & - & 0.044 & 0.737 & 0.737 \\
E & 0.044 & - & 0.245 & 0.383 \\
H & 0.737 & 0.245 & - & 0.828 \\
LEH & 0.737 & 0.383 & 0.828 & - \\
\bottomrule
\end{tabularx}
}
}

\scalebox{0.82}{
\subfloat[R-SVM]{
\begin{tabularx}{0.32\textwidth}{lYYYY}
\toprule
& L & E & H & LEH \\
\midrule
Rank & \textbf{1.79} & 2.67 & 3.37 & 2.18 \\
\midrule
L & - & 0.001 & 0.000 & 0.110 \\
E & 0.001 & - & 0.011 & 0.084 \\
H & 0.000 & 0.011 & - & 0.000 \\
LEH & 0.110 & 0.084 & 0.000 & - \\
\bottomrule
\end{tabularx}
}
\subfloat[P-SVM]{
\begin{tabularx}{0.32\textwidth}{lYYYY}
\toprule
& L & E & H & LEH \\
\midrule
Rank & 2.75 & 2.44 & \textbf{2.24} & 2.58 \\
\midrule
L & - & 0.613 & 0.212 & 1.000 \\
E & 0.613 & - & 1.000 & 1.000 \\
H & 0.212 & 1.000 & - & 0.471 \\
LEH & 1.000 & 1.000 & 0.471 & - \\
\bottomrule
\end{tabularx}
}
\subfloat[LR]{
\begin{tabularx}{0.32\textwidth}{lYYYY}
\toprule
& L & E & H & LEH \\
\midrule
Rank & 3.02 & 2.32 & \textbf{1.96} & 2.70 \\
\midrule
L & - & 0.011 & 0.000 & 0.331 \\
E & 0.011 & - & 0.331 & 0.331 \\
H & 0.000 & 0.331 & - & 0.007 \\
LEH & 0.331 & 0.331 & 0.007 & - \\
\bottomrule
\end{tabularx}
}
}

\scalebox{0.82}{
\subfloat[NB]{
\begin{tabularx}{0.32\textwidth}{lYYYY}
\toprule
& L & E & H & LEH \\
\midrule
Rank & \textbf{2.05} & 2.88 & 2.60 & 2.47 \\
\midrule
L & - & 0.004 & 0.074 & 0.245 \\
E & 0.004 & - & 0.491 & 0.286 \\
H & 0.074 & 0.491 & - & 0.612 \\
LEH & 0.245 & 0.286 & 0.612 & - \\
\bottomrule
\end{tabularx}
}
\subfloat[R-MLP]{
\begin{tabularx}{0.32\textwidth}{lYYYY}
\toprule
& L & E & H & LEH \\
\midrule
Rank & \textbf{2.07} & 2.61 & 2.81 & 2.51 \\
\midrule
L & - & 0.074 & 0.014 & 0.209 \\
E & 0.074 & - & 0.850 & 0.850 \\
H & 0.014 & 0.850 & - & 0.652 \\
LEH & 0.209 & 0.850 & 0.652 & - \\
\bottomrule
\end{tabularx}
}
\subfloat[L-MLP]{
\begin{tabularx}{0.32\textwidth}{lYYYY}
\toprule
& L & E & H & LEH \\
\midrule
Rank & 2.82 & \textbf{2.30} & 2.32 & 2.56 \\
\midrule
L & - & 0.177 & 0.177 & 0.830 \\
E & 0.177 & - & 0.942 & 0.830 \\
H & 0.177 & 0.942 & - & 0.830 \\
LEH & 0.830 & 0.830 & 0.830 & - \\
\bottomrule
\end{tabularx}
}
}
\end{table}

\begin{table}[!htb]
\small
\caption{A comparison of sampling in a specific regions, with \textbf{F-measure} used as the performance metric.}
\label{table:regions-F-measure}
\centering
\scalebox{0.82}{
\subfloat[CART]{
\begin{tabularx}{0.32\textwidth}{lYYYY}
\toprule
& L & E & H & LEH \\
\midrule
Rank & 2.49 & 2.68 & \textbf{2.19} & 2.63 \\
\midrule
L & - & 1.000 & 0.652 & 1.000 \\
E & 1.000 & - & 0.253 & 1.000 \\
H & 0.652 & 0.253 & - & 0.253 \\
LEH & 1.000 & 1.000 & 0.253 & - \\
\bottomrule
\end{tabularx}
}
\subfloat[KNN]{
\begin{tabularx}{0.32\textwidth}{lYYYY}
\toprule
& L & E & H & LEH \\
\midrule
Rank & 2.70 & 2.53 & \textbf{2.19} & 2.58 \\
\midrule
L & - & 1.000 & 0.212 & 1.000 \\
E & 1.000 & - & 0.504 & 1.000 \\
H & 0.212 & 0.504 & - & 0.331 \\
LEH & 1.000 & 1.000 & 0.331 & - \\
\bottomrule
\end{tabularx}
}
\subfloat[L-SVM]{
\begin{tabularx}{0.32\textwidth}{lYYYY}
\toprule
& L & E & H & LEH \\
\midrule
Rank & 2.93 & 2.49 & \textbf{2.05} & 2.53 \\
\midrule
L & - & 0.209 & 0.002 & 0.209 \\
E & 0.209 & - & 0.209 & 0.885 \\
H & 0.002 & 0.209 & - & 0.150 \\
LEH & 0.209 & 0.885 & 0.150 & - \\
\bottomrule
\end{tabularx}
}
}

\scalebox{0.82}{
\subfloat[R-SVM]{
\begin{tabularx}{0.32\textwidth}{lYYYY}
\toprule
& L & E & H & LEH \\
\midrule
Rank & 2.37 & 2.61 & \textbf{2.33} & 2.68 \\
\midrule
L & - & 0.929 & 1.000 & 0.881 \\
E & 0.929 & - & 0.881 & 1.000 \\
H & 1.000 & 0.881 & - & 0.881 \\
LEH & 0.881 & 1.000 & 0.881 & - \\
\bottomrule
\end{tabularx}
}
\subfloat[P-SVM]{
\begin{tabularx}{0.32\textwidth}{lYYYY}
\toprule
& L & E & H & LEH \\
\midrule
Rank & 2.46 & 2.61 & 2.51 & \textbf{2.42} \\
\midrule
L & - & 1.000 & 1.000 & 1.000 \\
E & 1.000 & - & 1.000 & 1.000 \\
H & 1.000 & 1.000 & - & 1.000 \\
LEH & 1.000 & 1.000 & 1.000 & - \\
\bottomrule
\end{tabularx}
}
\subfloat[LR]{
\begin{tabularx}{0.32\textwidth}{lYYYY}
\toprule
& L & E & H & LEH \\
\midrule
Rank & 3.11 & 2.60 & \textbf{1.70} & 2.60 \\
\midrule
L & - & 0.106 & 0.000 & 0.106 \\
E & 0.106 & - & 0.001 & 1.000 \\
H & 0.000 & 0.001 & - & 0.001 \\
LEH & 0.106 & 1.000 & 0.001 & - \\
\bottomrule
\end{tabularx}
}
}

\scalebox{0.82}{
\subfloat[NB]{
\begin{tabularx}{0.32\textwidth}{lYYYY}
\toprule
& L & E & H & LEH \\
\midrule
Rank & \textbf{2.12} & 2.95 & 2.39 & 2.54 \\
\midrule
L & - & 0.004 & 0.553 & 0.245 \\
E & 0.004 & - & 0.061 & 0.286 \\
H & 0.553 & 0.061 & - & 0.553 \\
LEH & 0.245 & 0.286 & 0.553 & - \\
\bottomrule
\end{tabularx}
}
\subfloat[R-MLP]{
\begin{tabularx}{0.32\textwidth}{lYYYY}
\toprule
& L & E & H & LEH \\
\midrule
Rank & \textbf{2.30} & 2.60 & 2.44 & 2.67 \\
\midrule
L & - & 0.766 & 1.000 & 0.766 \\
E & 0.766 & - & 1.000 & 1.000 \\
H & 1.000 & 1.000 & - & 1.000 \\
LEH & 0.766 & 1.000 & 1.000 & - \\
\bottomrule
\end{tabularx}
}
\subfloat[L-MLP]{
\begin{tabularx}{0.32\textwidth}{lYYYY}
\toprule
& L & E & H & LEH \\
\midrule
Rank & 3.18 & 2.39 & \textbf{1.89} & 2.54 \\
\midrule
L & - & 0.004 & 0.000 & 0.027 \\
E & 0.004 & - & 0.071 & 0.537 \\
H & 0.000 & 0.071 & - & 0.020 \\
LEH & 0.027 & 0.537 & 0.020 & - \\
\bottomrule
\end{tabularx}
}
}
\end{table}

\begin{table}[!htb]
\small
\caption{A comparison of sampling in a specific regions, with \textbf{G-mean} used as the performance metric.}
\label{table:regions-G-mean}
\centering
\scalebox{0.82}{
\subfloat[CART]{
\begin{tabularx}{0.32\textwidth}{lYYYY}
\toprule
& L & E & H & LEH \\
\midrule
Rank & \textbf{2.18} & 2.68 & 2.72 & 2.42 \\
\midrule
L & - & 0.147 & 0.147 & 0.830 \\
E & 0.147 & - & 0.885 & 0.830 \\
H & 0.147 & 0.885 & - & 0.652 \\
LEH & 0.830 & 0.830 & 0.652 & - \\
\bottomrule
\end{tabularx}
}
\subfloat[KNN]{
\begin{tabularx}{0.32\textwidth}{lYYYY}
\toprule
& L & E & H & LEH \\
\midrule
Rank & 2.46 & 2.70 & \textbf{2.19} & 2.65 \\
\midrule
L & - & 0.929 & 0.830 & 0.929 \\
E & 0.929 & - & 0.212 & 0.929 \\
H & 0.830 & 0.212 & - & 0.212 \\
LEH & 0.929 & 0.929 & 0.212 & - \\
\bottomrule
\end{tabularx}
}
\subfloat[L-SVM]{
\begin{tabularx}{0.32\textwidth}{lYYYY}
\toprule
& L & E & H & LEH \\
\midrule
Rank & 2.75 & \textbf{2.12} & 2.56 & 2.56 \\
\midrule
L & - & 0.054 & 1.000 & 1.000 \\
E & 0.054 & - & 0.209 & 0.209 \\
H & 1.000 & 0.209 & - & 1.000 \\
LEH & 1.000 & 0.209 & 1.000 & - \\
\bottomrule
\end{tabularx}
}
}

\scalebox{0.82}{
\subfloat[R-SVM]{
\begin{tabularx}{0.32\textwidth}{lYYYY}
\toprule
& L & E & H & LEH \\
\midrule
Rank & \textbf{1.82} & 2.68 & 3.37 & 2.12 \\
\midrule
L & - & 0.001 & 0.000 & 0.217 \\
E & 0.001 & - & 0.014 & 0.041 \\
H & 0.000 & 0.014 & - & 0.000 \\
LEH & 0.217 & 0.041 & 0.000 & - \\
\bottomrule
\end{tabularx}
}
\subfloat[P-SVM]{
\begin{tabularx}{0.32\textwidth}{lYYYY}
\toprule
& L & E & H & LEH \\
\midrule
Rank & 2.81 & 2.35 & \textbf{2.12} & 2.72 \\
\midrule
L & - & 0.178 & 0.028 & 0.717 \\
E & 0.178 & - & 0.691 & 0.383 \\
H & 0.028 & 0.691 & - & 0.041 \\
LEH & 0.717 & 0.383 & 0.041 & - \\
\bottomrule
\end{tabularx}
}
\subfloat[LR]{
\begin{tabularx}{0.32\textwidth}{lYYYY}
\toprule
& L & E & H & LEH \\
\midrule
Rank & 2.98 & 2.35 & \textbf{1.93} & 2.74 \\
\midrule
L & - & 0.027 & 0.000 & 0.310 \\
E & 0.027 & - & 0.245 & 0.245 \\
H & 0.000 & 0.245 & - & 0.003 \\
LEH & 0.310 & 0.245 & 0.003 & - \\
\bottomrule
\end{tabularx}
}
}

\scalebox{0.82}{
\subfloat[NB]{
\begin{tabularx}{0.32\textwidth}{lYYYY}
\toprule
& L & E & H & LEH \\
\midrule
Rank & \textbf{2.30} & 2.67 & 2.54 & 2.49 \\
\midrule
L & - & 0.766 & 0.929 & 1.000 \\
E & 0.766 & - & 1.000 & 1.000 \\
H & 0.929 & 1.000 & - & 1.000 \\
LEH & 1.000 & 1.000 & 1.000 & - \\
\bottomrule
\end{tabularx}
}
\subfloat[R-MLP]{
\begin{tabularx}{0.32\textwidth}{lYYYY}
\toprule
& L & E & H & LEH \\
\midrule
Rank & \textbf{2.18} & 2.58 & 2.72 & 2.53 \\
\midrule
L & - & 0.286 & 0.147 & 0.440 \\
E & 0.286 & - & 1.000 & 1.000 \\
H & 0.147 & 1.000 & - & 1.000 \\
LEH & 0.440 & 1.000 & 1.000 & - \\
\bottomrule
\end{tabularx}
}
\subfloat[L-MLP]{
\begin{tabularx}{0.32\textwidth}{lYYYY}
\toprule
& L & E & H & LEH \\
\midrule
Rank & 2.72 & \textbf{2.20} & 2.43 & 2.65 \\
\midrule
L & - & 0.194 & 0.694 & 1.000 \\
E & 0.194 & - & 1.000 & 0.194 \\
H & 0.694 & 1.000 & - & 1.000 \\
LEH & 1.000 & 0.194 & 1.000 & - \\
\bottomrule
\end{tabularx}
}
}
\end{table}

\newpage

\section{Examination of the impact of energy parameter}

\begin{figure*}[!htb]
\centering
\includegraphics[width=\textwidth]{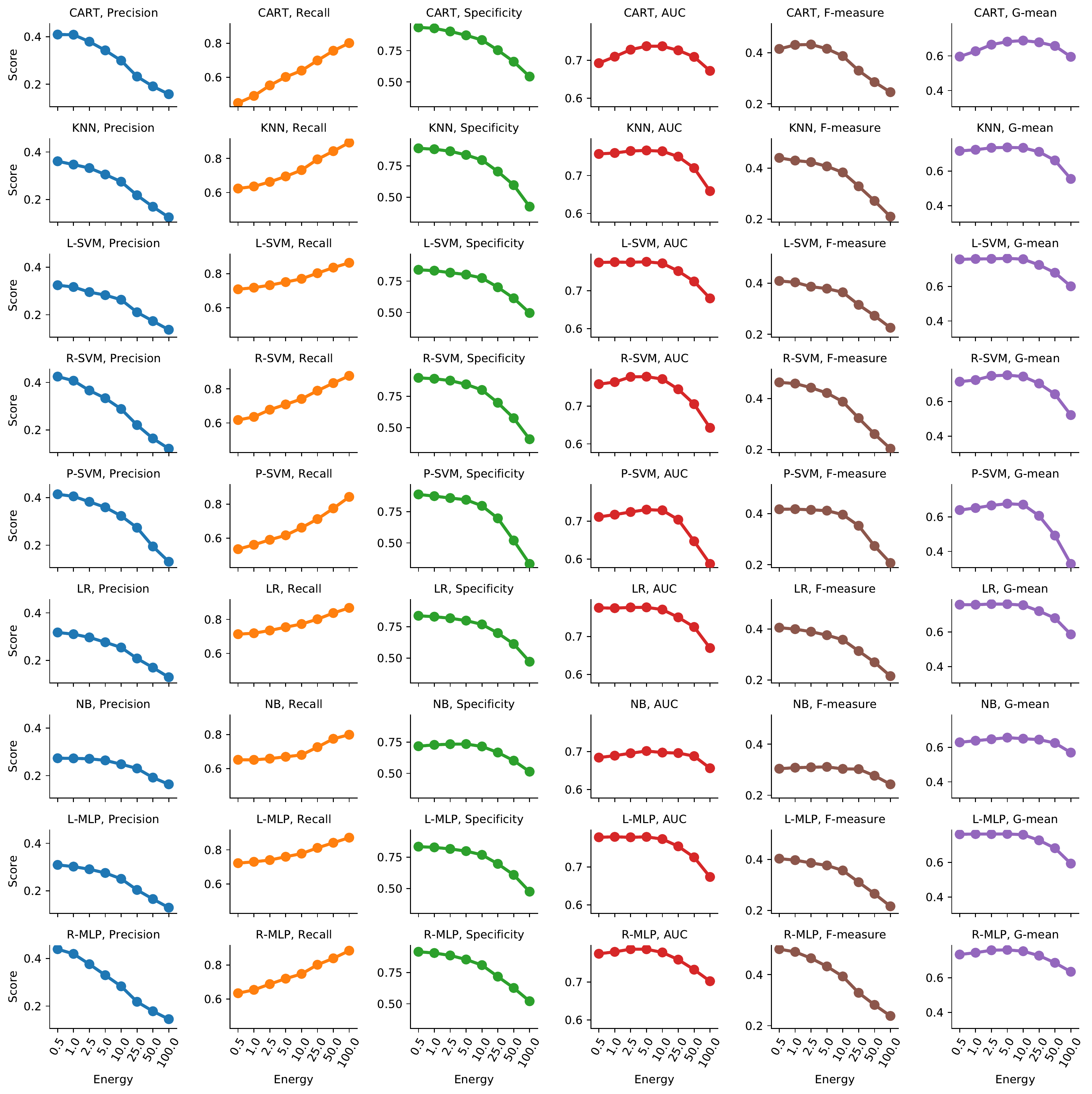}
\caption{Visualization of the impact of $energy$ parameter on the performance with respect to different metrics.}
\label{fig:preliminary-energy}
\end{figure*}

In addition to the sampling region, another hyperparameter that can have a significant influence on the performance of RB-CCR is its energy, which regulates the size of sampling regions and the extent of translation. To assess the exact impact of energy on the algorithms behavior we conducted an experiment, in which we measured the change in performance depending on the choice of energy value. Similar to the previous experiments, we used cross-validation to adjust the values of $\gamma$ parameter in \{0.5, 1.0, 2.5, 5.0, 10.0\}, and the choice of sampling regions in \{L, E, H, LEH\}. In Figure~\ref{fig:preliminary-energy} we presented the impact of energy, with values chosen from \{0.5, 1.0, 2.5, 5.0, ..., 100.0\}, on the performance averaged across all of the datasets. First of all, as can be seen the choice of energy has, on average, a clear impact on precision, specificity and recall, with the first two decreasing monotonically proportional to the energy, and the last one increasing monotonically. This is relevant because it indicates that CCR already has an inbuilt mechanism for controlling the precision-recall trade-off, and as a result the performance improvement displayed by the RB-CCR cannot be explained solely due to providing that, rather it provides a more optimal trade-off (with respect to the combined metrics).

Furthermore, as can be seen, the value of energy for which RB-CCR achieves the best average performance depends on the choice of classifier and metric. In the case of F-measure the best performance is observed for the minimal energy, when the precision-to-recall ratio is the highest. This is another empirical confirmation of the claim made in \cite{brzezinski2019dynamics}, according to which F-measure tends to be more biased towards the majority class performance. More importantly, in the case of AUC and G-mean, both of which tend to be highly correlated, two types of behavior can be observed. First of all, in the case of linear models, that is LR, L-SVM and L-MLP, the best average performance was observed with the energy values in \{0.5, 1.0, 2.5, 5.0\}, with little to no difference between those values. Secondly, in the case of the remaining classifiers the optimal performance was observed around the value of energy equal to 5.0, with both decrease and the increase of energy negatively affecting the performance. Considering the fact that as the energy goes down the methods behavior starts resembling random oversampling more closely, this seems to indicate that the expected performance gain due to using RB-CCR is highest for non-linear methods, capable of producing more complex decision boundaries. Finally, irregardless of the choice of classifier, from the practical standpoint observed results also suggest that using the value of energy equal to 5.0 is a sensible default.

\end{document}